\documentclass[preprint,12pt]{elsarticle}

\usepackage{amssymb}

\usepackage{amsmath,amssymb,amsfonts,bm}
\usepackage{flushend}

\usepackage{graphicx}
\usepackage{float}
\floatstyle{plaintop}
\restylefloat{table}
\usepackage{textcomp}
\usepackage{float}
\usepackage{tablefootnote}
\usepackage{epstopdf}    
\usepackage{flushend}       
\usepackage{array}
\usepackage{mdwtab}
\usepackage{eqparbox}
\usepackage{url}
\usepackage{hyperref}
\usepackage[export]{adjustbox}
\usepackage{wrapfig}
\usepackage{lipsum}
\usepackage{graphicx,subfigure}
\usepackage{adjustbox}
\usepackage[normalem]{ulem}
\useunder{\uline}{\ul}{}
\usepackage{amssymb}
 \usepackage{amsmath,amssymb,amsfonts,mathtools}
\DeclarePairedDelimiter{\norm}{\lVert}{\rVert}
\usepackage{pdflscape}
\usepackage{afterpage}
\newcommand{\RNum}[1]{\lowercase\expandafter{\romannumeral #1\relax}}
\newcommand{\RNumU}[1]{\uppercase\expandafter{\romannumeral #1\relax}}
\usepackage{graphicx}
\usepackage{booktabs,makecell}
\usepackage{multirow}
\usepackage{caption}
\usepackage{mathtools}
\usepackage{subfigure}
\usepackage{comment}
\usepackage{amsmath}
\usepackage{longtable}
\usepackage{algorithm}
\usepackage[noend]{algpseudocode}
\usepackage{siunitx}

\usepackage{blindtext}
\usepackage{float}
\usepackage{geometry}
\floatstyle{plaintop}
\restylefloat{table}

\journal{Elsevier}

\begin{document}

\begin{frontmatter}



\title{Support matrix machine: A review}


\author[inst1]{Anuradha Kumari}

\affiliation[inst1]{organization={Department of Mathematics, Indian Institute of Technology Indore},
            addressline={Simrol}, 
            city={Indore},
            postcode={453552}, 
            state={Madhya Pradesh},
            country={India}}

\author[inst1]{Mushir Akhtar}
\author[inst2]{Rupal Shah}
\affiliation[inst2]{organization={Department of Electrical Engineering, Indian Institute of Technology Indore},
            addressline={Simrol}, 
            city={Indore},
            postcode={453552}, 
            state={Madhya Pradesh},
            country={India}}
\author[inst1]{M. Tanveer\corref{mycorrespondingauthor}}

\cortext[mycorrespondingauthor]{Corresponding author}

\begin{abstract}
Support vector machine (SVM) is one of the most studied paradigms in the realm of machine learning for classification and regression problems. It relies on vectorized input data. However, a significant portion of the real-world data exists in matrix format, which is given as input to SVM by reshaping the matrices into vectors. The process of reshaping disrupts the spatial correlations inherent in the matrix data. Also, converting matrices into vectors results in input data with a high dimensionality, which introduces significant computational complexity.  To overcome these issues in classifying matrix input data, support matrix machine (SMM) is proposed. It represents one of the emerging methodologies tailored for handling matrix input data. The SMM method preserves the structural information of the matrix data by using the spectral elastic net property which is a combination of the nuclear norm and Frobenius norm. This article provides the first in-depth analysis of the development of the SMM model, which can be used as a thorough summary by both novices and experts. We discuss numerous  SMM variants, such as robust, sparse, class imbalance, and multi-class classification models. We also analyze the applications of the SMM model and conclude the article by outlining potential future research avenues and possibilities that may motivate academics to advance the SMM algorithm.
\end{abstract}




\end{frontmatter}

\section{Introduction}{\label{sec:introduction}}
Support vector machines (SVMs) \cite{cortes1995support} are one of the widely growing and the most popular classification techniques in recent years due to their strong theoretical foundation and high generalization abilities \cite{cervantes2020comprehensive}. These are among the most effective and reliable algorithms for classification and regression in a variety of real-world applications such as economy \cite{wang2005new}, medical \cite{zhu2016efficient}, text classification \cite{zhang2008text}, regression \cite{chuang2007fuzzy}, and so on.  It creates the optimal classification hyperplane with the maximum margin between the two classes of samples on the basis of the structural risk minimization (SRM) principle and the maximum margin principle \cite{deng2012support}. SVM is used with other techniques to improve classification and training efficiency such as SVM with sequential minimal optimization (SMO) \cite{platt1998sequential, keerthi2001improvements} and SVM with successive overrelaxation (SOR) technique \cite{mangasarian1999successive}, SVMlight \cite{joachims1999svmlight} and so on. 

SVMs are appropriate classifiers for the input data in vector form. However, some application generates data as high-order tensors, which is a more natural way to express the data. For example gait silhouette sequences (3-order tensors), gray image (2-order tensors), color video (4-order tensors), and so on. Tucker tensor decomposition (TTD) allows for the transformation of all high-order tensors into matrices \cite{kotsia2011support}. Also, some real applications have data in matrix form such as medical images, photorealistic images of faces, palms and so on. Thus, the study of classification techniques with input data of matrix type is very important. Also, matrix data contains correlations between rows and columns, thus conveying more information than vector data, e.g., correlations between different channels of EEG data \cite{zhou2014regularized}, image data spatial relations of neighbourhood pixels \cite{wolf2007modeling}, and so forth.      

If matrix input data is considered in traditional classifiers such as SVM, the matrices need to be reshaped into vectors which destroy the structural information i.e., spatial correlations present in the matrix \cite{wolf2007modeling}. Furthermore, the process of converting matrices into vectors results in high-dimensional vectors when dealing with a relatively small number of input samples, consequently giving rise to the issue known as the curse of dimensionality.  The high dimensionality contributes to the increased computational complexity of the classification. Some classification models that exploit the spatial correlation of matrix data are proposed, such as rank $k$-SVM \cite{wolf2007modeling} which considers $k$ orthogonal matrices of rank 1 to represent the regression matrix, bilinear SVM \cite{pirsiavash2009bilinear} that generates two low-rank matrices out of the regression matrix. However, in $k$-SVM and bilinear SVM, the ranks of the regression matrix need to be fixed. Also, the aforementioned methods lead to non-convex optimization problems.\par   
In the domain of machine learning, to classify matrix input data by retaining its spatial correlation and addressing the curse of dimensionality, \citet{luo2015support} introduced an advancement in the realm of supervised learning known as support matrix machine (SMM). SMM uses the nuclear norm of a matrix as the convex alternative of the matrix rank \cite{kobayashi2012efficient}. The use of nuclear norm is inspired by matrix completion \cite{huang2013robust, candes2012exact} and low-rank matrix approximation \cite{srebro2005rank}. It exploits the existing correlation between the columns and rows of the input matrix data. Furthermore, SMM also uses the Frobenius norm in the optimization along with penalizing the data points by hinge loss which results in the sparsity of the model. The combination of the squared Frobenius norm and the nuclear norm is known as spectral elastic net which is parallel to the conventional elastic net by \citet{zou2005regularization}. As a result of the spectral elastic net, SMM has improved generalization classification performance for matrix input data and a convex optimization problem.    

\begin{figure*} 
\centering{
\includegraphics[width=1.1\textwidth,keepaspectratio]{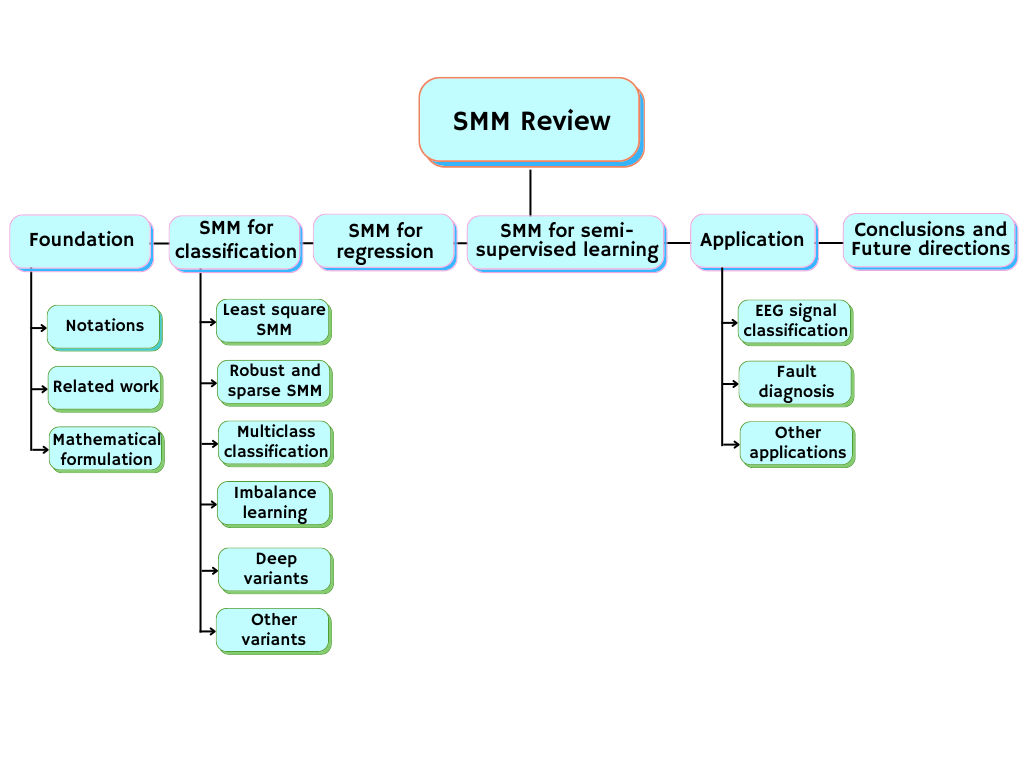}}
\caption{Layout of the paper}
    \label{fig:layout}
\end{figure*}

Since the last few years, there has been a noticeable growth in SMM and its applications in several real-world scenarios. Different variants of SMM introduced in the last decade encompassing robust and sparse models, models designed to handle imbalanced data, least square adaptations of SMM, and the extension of SMM to multi-class classification tasks. SMM is also employed in conjunction with deep variants. It has a major application in the field of electroencephalogram (EEG) data classification and fault diagnosis. In this paper, we discuss the growth of SMM, its variants along with applications. Figure \ref{fig:layout} highlights the overall structure and progression of the paper.      


The rest of the paper is organized as follows: Section \ref{sec:serach_metho} discusses the search methodology related to the review paper. Section \ref{sec:mathematical form} contains the foundation of SMM along with the related work. Different variants of SMM and its development in recent years for classification problems are briefly discussed in Section \ref{sec:variants}. Section \ref{sec:SMR} analyses an SMM model for regression problems and Section \ref{sec:semi_supervised}  discusses SMM in the aspect of semi-supervised learning. We present the various applications of SMM in a real-world domain in Section \ref{sec:application}. Section \ref{sec:conclusion}  concludes the article with several prospective research areas.    

\section{Search methodology}{\label{sec:serach_metho}} 
The papers in the compilation of this review paper were included from two search engines, Google Scholar and Scopus. It was mainly performed in May 2023. The keywords for the search are confined to ``support matrix machine" and ``SMM". The screening was done based on the title of the papers and initial screening excluded studies not concentrating on support matrix machine. It contains in total of 59 papers which got published from 2015 onwards.     

\section{Foundation of SMM}{\label{sec:mathematical form}}
In this section, we delve into the essential components of SMM, beginning with an exploration of notations followed by an examination of relevant prior research, and provide a detailed formulation of the SMM framework.

\subsection{Notations}
First, we discuss the notations used throughout the paper. Following the commonly accepted conventions, we denote scalar values using lowercase letters,
vectors using lowercase bold letters, 
and matrices using uppercase bold letters. 
The singular value decomposition (SVD) for a given matrix $\mathbf{X} \in \mathbb{R}^{p \times q}$ is condensed as $\mathbf{X}=\mathbf{E} \mathbf{\Sigma} \mathbf{F}^{T}$, where $\mathbf{E} \in \mathbb{R}^{p \times r}$ and $\mathbf{F} \in \mathbb{R}^{q \times r}$ are unitary matrices and $\boldsymbol{\Sigma}$ 
 is the diagonal matrix with its diagonal entries being singular values $\nu_{1}, \nu_{2} \ldots \nu_{r}$. These singular values are ordered such that
 $\nu_{1} \geq \nu_{2} \geq \ldots \geq \nu_{r} \geq 0$. The count of non-zero singular values indicates the rank of $\mathbf{X}$, hence $r \leq \min (p, q)$. Further, we let $\|\mathbf{X}\|_{*}=\sum_{i=1}^{r} \nu_{i}$ to be the nuclear norm of $\mathbf{X}$, and $\|\mathbf{X}\|_{F}=\sqrt{\sum_{i, j} x_{i,j}^{2}}=\sqrt{\sum_{i=1}^{r} \nu_{i}^2}$ be the Frobenius norm. Other notations that are utilized throughout the paper are presented in Table \ref{tab:Notation table} along with their descriptions.


\begin{table}[]
\centering
\caption{Notation used in the paper.}
\label{tab:Notation table}
\resizebox{12cm}{!}{%
\begin{tabular}{|l|l|}
\hline
Symbol & Description \\ \hline
    $C$   &    Number of classes in multiclass problems         \\ \hline
      $N$ &   Number of training samples          \\ \hline
       $p \times q$&  Order of input matrices data           \\ \hline
     $\kappa$  & Condition number of the matrix for inversion.            \\ \hline
      $\epsilon$ & Expected accuracy of the output space.             \\ \hline
     $H$  & Hilbert space containing input matrices.            \\ \hline
     $V$  &  Symmetric matrix            \\ \hline
     $\zeta$   & Trade-off parameter             \\ \hline
      $\lambda$ &      Nuclear norm constraint       \\ \hline
     $\rho$  &  Inherent coeffecient of ADMM           \\ \hline
\end{tabular}%
}
\end{table}

\subsection{Related work}
In this subsection, we begin with a simple description of the matrix classification problem. Provided a collection of training matrix samples $\left\{\mathbf{X}_{i}, y_{i}\right\}_{i=1}^{N}$, where $\mathbf{X}_{i} \in \mathbb{R}^{p \times q}$ is the $i^{th}$ input matrix and $y_{i} \in\{-1, 1\}$ is its corresponding class label. The main objective is to train a function $f: \mathbb{R}^{p \times q} \rightarrow \mathbb{R}$ using the provided training samples, which can significantly determine the class of a new unseen matrix sample.

To address the matrix classification problem using conventional techniques, a commonly used heuristic approach involves transforming each matrix data instance $\mathbf{X}_{i}$ into a vector format. Subsequently, the model is trained using this collection of vectorized data. Among the conventional classifiers, one efficient example is the soft margin SVM model \cite{cortes1995support}, whose optimization problem is given as follows:
\begin{align} \label{SVMeqs}
\min _{\mathbf{w}, b}~ \frac{1}{2} \mathbf{w}^{T} \mathbf{w}+\zeta \sum_{i=1}^{N}\left\{1-y_{i}\left(\mathbf{w}^{T} \mathbf{x}_{i}+b\right)\right\}_{+},
\end{align}
where $\mathbf{w} \in \mathbb{R}^{pq}$; $b \in \mathbb{R}$; $\mathbf{x}_{i}=\operatorname{vec}\left(\mathbf{X}_{i}\right)$ shows the vectorized format of matrix $\mathbf{X}_{i}$, $\{h\}_{+}:=\max \{0,h\}$ indicates the classical hinge loss function, and $\zeta > 0$ denotes the regularization parameter.
\par
Since $\operatorname{tr}\left(\mathbf{W}^{T} \mathbf{W}\right)$ is equivalent to
  $\operatorname{vec}(\mathbf{W})^{T} \operatorname{vec}(\mathbf{W})$ and $\operatorname{tr}\left(\mathbf{W}^{T} \mathbf{X}_{i}\right)$ is equivalent to 
  $\operatorname{vec}(\mathbf{W})^{T}$ $\operatorname{vec}\left(\mathbf{X}_{i}\right)$. Thus, in terms of computational considerations, equation (\ref{SVMeqs}) is identical to the subsequent formulation for performing matrix classification directly:
\begin{align} \label{SVMwithMatrix}
\min _{\mathbf{W}, b}~ \frac{1}{2} \operatorname{tr}\left(\mathbf{W}^{T} \mathbf{W}\right)+\zeta \sum_{i=1}^{N}\left\{1-y_{i}\left[\operatorname{tr}\left(\mathbf{W}^{T} \mathbf{X}_{i}\right)+b\right]\right\}_{+} .
\end{align}
This illustrates that directly employing equation (\ref{SVMwithMatrix}) for classification is insufficient to capture the inherent structure present within each input matrix. As a consequence, this approach leads to a loss of information.

\par
In order to consider the structural characteristics, an intuitive strategy involves capturing the correlations inherent within each input matrix by introducing a low-rank restriction on the matrix $\mathbf{W}$. To tackle this issue, a number of approaches are suggested such as the low-rank SVM \cite{wolf2007modeling} and the bi-linear SVM \cite{pirsiavash2009bilinear}. However, these approaches have limitations as they demand manual pre-specification of the latent rank of $\mathbf{W}$ tailored to various applications. To address the aforementioned issues, \citet{luo2015support} proposed an advancement in the realm of supervised learning, SMM, which overcomes the pre-specified rank criteria, preserves the structural information of the input matrix by employing the spectral elastic net penalty for the regression matrix, and produces a better result for matrix-form data.
\begin{figure*}
\centering
     { %
\includegraphics[width=1.0\textwidth,keepaspectratio]{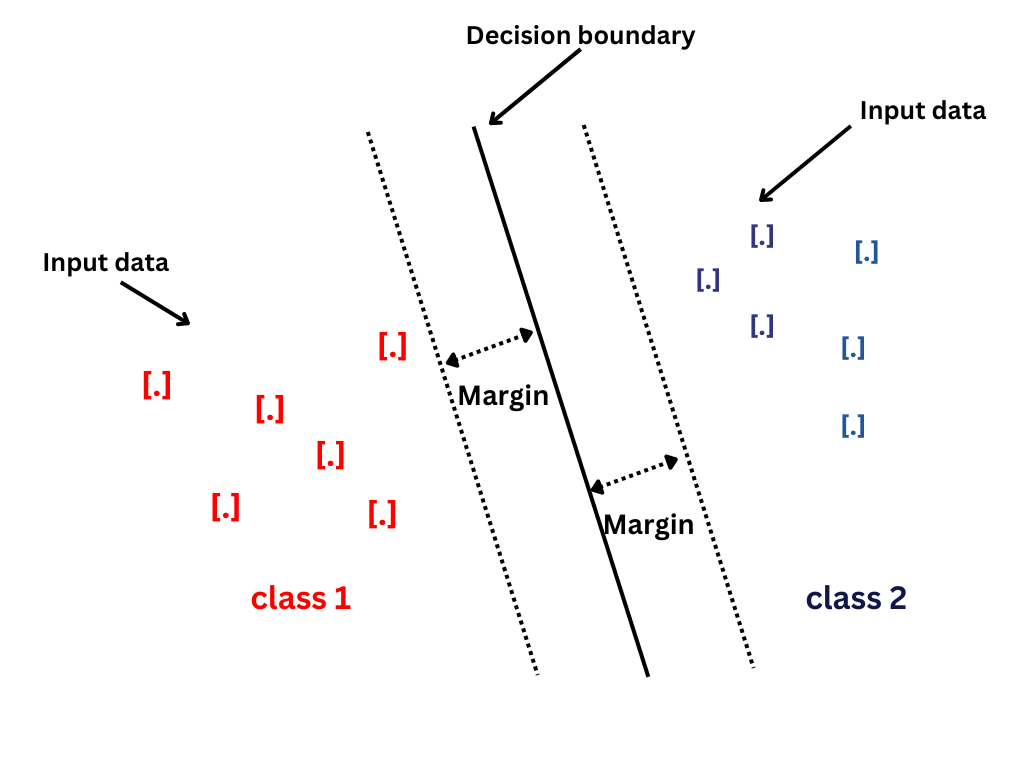}}
\caption{Representation of hyperplanes of SMM.}
    \label{fig:hyperplane_SMM}
      \end{figure*}
\subsection{Mathematical Formulation of SMM}
SMM is a proficient matrix-based adaptation of SVM, capitalizing on the strengths of SVM, which include robust generalization capabilities. Figure \ref{fig:hyperplane_SMM} depicts the hyperplane associated with SMM. Further, it has the ability to comprehensively harness the structural insights embedded within matrix data. The objective function of SMM, under the maximum margin principle, is to find a decision hyperplane, which is given as:
\begin{align} \label{SMMprimary}
\frac{1}{2} \operatorname{tr}\left(\mathbf{W}^{T} \mathbf{W}\right)+\lambda\|\mathbf{W}\|_{*}+\zeta \sum_{i}^{N}\left\{1-y_{i}\left[\operatorname{tr}\left(\mathbf{W}^{T} \mathbf{X}_{i}\right)+b_{i}\right]\right\}_{+}.
\end{align} 
\begin{figure*} 
\centering{
\includegraphics[width=.9\textwidth,keepaspectratio]{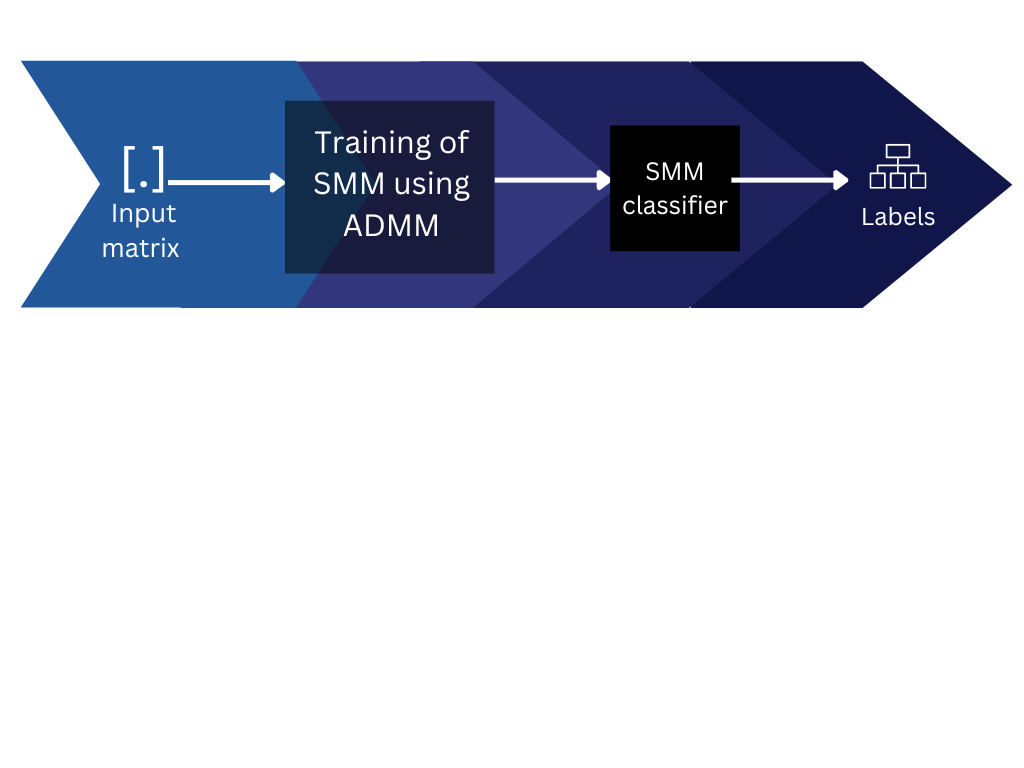}}
\caption{Block representation of SMM. }
    \label{fig:SMM_flow chart}
\end{figure*}
Here, the initial term $\frac{1}{2} \operatorname{tr}\left(\mathbf{W}^{T}\mathbf{W}\right)+\lambda\|\mathbf{W}\|_{*}$ pertains to the utilization of spectral elastic net regularization, which serves the purpose of capturing correlations inherent within individual matrices. On the other hand, the last summation term represents the hinge loss function. The term $\frac{1}{2} \operatorname{tr}\left(\mathbf{W}^{T} \mathbf{W}\right)$ can also be written as  $\frac{1}{2}\|\mathbf{W}\|_{F}^2$, which represents the square Frobenius norm.
\par
The Frobenius norm of matrix $\mathbf{W}$ serves as a regularization term, aiming to find a weight matrix with a reduced rank. It is also crucial to highlight that the nuclear norm serves as a regularization factor to ascertain the rank of matrix $\mathbf{W}$. As estimating the rank of a matrix can be a complex problem with NP-hard characteristics \cite{wang2015visual}, however, the nuclear norm is widely acknowledged as the optimal convex approximation method for assessing the rank of the matrix \cite{zhou2014regularized, candes2012exact}.
Additionally, the low-rank parameter $\lambda$ governs the level of structure information incorporated for constructing the classification hyperplane. The presence of the term $\|\mathbf{W}\|_{*}$ introduces non-smoothness to the objective function of SMM. This characteristic poses a challenge when attempting to directly solve equation (\ref{SMMprimary}). Consequently, the solution for SMM is derived through the application of the alternating direction method of multipliers (ADMM) \cite{goldstein2014fast}. Now, by introducing an auxiliary matrix variable, $\mathbf{Q}$, the objective function of SMM can be reformulated in the following manner:
\begin{align} \label{SMMafterintroducingQ}
\underset{(\mathbf{W}, b), \mathbf{Q}}{\arg \min } ~ F(\mathbf{W}, b)+G(\mathbf{Q}) \nonumber \\
\text { s.t. } \mathbf{W}-\mathbf{Q}=0.
\end{align}
Here, $F(\mathbf{W}, b)=\frac{1}{2} \operatorname{tr}\left(\mathbf{W}^{T} \mathbf{W}\right)+ \zeta \sum_{i}^{N}\left\{1-y_{i}\left[\operatorname{tr}\left(\mathbf{W}^{T} \mathbf{X}_{i}\right)+b_{i}\right]\right\}_{+}$, and $G(\mathbf{Q})=$ $\lambda\|\mathbf{Q}\|_{*}$. Equation (\ref{SMMafterintroducingQ}) is then rewritten with augmented Lagrange multipliers $\mathbf{\beta}$ as follows:
\begin{align} \label{lagrange}
L(\mathbf{Q}, \mathbf{\beta},(\mathbf{W}, b))=F(\mathbf{W}, b)+G(\mathbf{Q})+\frac{\rho}{2}\|\mathbf{Q}-\mathbf{W}\|_{F}^{2}+\operatorname{tr}\left[\mathbf{\beta}^{T}(\mathbf{Q}-\mathbf{W})\right].
\end{align}
Here, $\rho$ denotes the inherent coefficient of ADMM method. Within this framework, we have three variable matrices that need to be determined: $\mathbf{Q}$, $\mathbf{\beta}$, and ($\mathbf{W}$, $b$). The process of obtaining the optimal solutions for these matrices involves an iterative approach. The general steps for updating these variable matrices are outlined as follows:
\begin{align} 
\mathbf{Q}^{(t+1)}=& \underset{\mathbf{Q}}{\arg \min } ~ L\left(\mathbf{Q}, \mathbf{\beta}^{(t)},\left(\mathbf{W}^{(t)}, b^{(t)}\right)\right),  \\
\left(\mathbf{W}^{(t+1)}, b^{(t+1)}\right)=&\underset{(\mathbf{W}, b)}{\arg \min }~ L\left(\mathbf{Q}^{(t+1)}, \mathbf{\beta}^{(t)},(\mathbf{W}, b)\right), \\
\mathbf{\beta}^{(t+1)}=&\mathbf{\beta}^{(t)}+\rho\left(\mathbf{Q}^{(t+1)}-\left(\mathbf{W}^{(t+1)}, b^{(t+1)}\right)\right).
\end{align}

The fundamental steps for solving within this context involve calculating $\mathbf{Q}^{(t)}$ and the pair $\left(\mathbf{W}^{(t)}\right.$, $\left.b^{(t)}\right)$ during each iteration. To update $\mathbf{Q}$ (for the sake of ease, we omit the superscripts in the subsequent explanation), suppose that $\mathbf{\beta}$ and $(\mathbf{W}, b)$ remain constant, then the solution of $\mathbf{Q}$ can be determined using the following equation:
\begin{align} \label{findP1}
 \underset{\mathbf{Q}}{\arg \min } ~G(\mathbf{Q})+\frac{\rho}{2}\|\mathbf{Q}-\mathbf{W}\|_{F}^{2}+\operatorname{tr}\left[\mathbf{\beta}^{T}(\mathbf{Q}-\mathbf{W})\right].
\end{align}

By solving equation (\ref{findP1}), we can derive the updating formula for the matrix $\mathbf{Q}$ during each iteration as follows:
\begin{align*}
\mathbf{Q}=\frac{1}{\rho} \mathcal{S}_{\lambda}(\rho \mathbf{W}-\mathbf{\beta}),
\end{align*}
where $\mathcal{S}_{\lambda}$ is the thresholding operator for singular value \cite{cai2010singular}. In a similar manner, the solution for the pair $(\mathbf{W}, b)$ is acquired by solving the following equation:
\begin{align} \label{w_b_solution}
\underset{(\mathbf{W}, b)}{\arg \min }~ F(\mathbf{W}, b)-\operatorname{tr}\left(\mathbf{\beta}^{T} \mathbf{W}\right)+\frac{\rho}{2}\|\mathbf{W}-\mathbf{Q}\|_{F}^{2}.
\end{align}
Subsequently, through the solution of equation (\ref{w_b_solution}), the formula for updating $(\mathbf{W}, b)$ can be expressed as follows:
\begin{align*}
\mathbf{W}=&\frac{1}{\rho+1}\left(\rho \mathbf{Q}+\mathbf{\beta}+\sum_{i=1}^{N} \gamma_{i} y_{i} \mathbf{X}_{i}\right), \\
b=&\frac{1}{|\mathbf{Q}|} \sum_{i=1}^{N}\left(y_{i}-\operatorname{tr}\left(\mathbf{W}^{T} \mathbf{X}_{i}\right)\right).
\end{align*}
Figure \ref{fig:SMM_flow chart} depicts the block representation of SMM.
The process of updating the parameters is terminated when the maximum specified number of iterations is achieved.
At this point, the optimal solutions for $\mathbf{Q}, \mathbf{\beta}$, and $(\mathbf{W}, b)$  are acquired. Then, the label for a new matrix data $\Tilde{\mathbf{X}}$ is predicted using the following decision function: 
\begin{align}
\Tilde{y}= {\operatorname{\text{sign}}}\left(\operatorname{tr}\left(\Tilde{\mathbf{X}}^{T} \mathbf{W}\right)+b\right).
\end{align}

\section{SMM for classification}{\label{sec:variants}}
SMM preserves the structural information of matrix input data and successfully classifies matrices. This motivated researchers to work in the field of SMM. In this section, we will discuss some of the well-known variants of SMMs which have different advantages over the classical SMM.

\subsection{Least square SMM}
 The classical SMM solves the optimization problem by incorporating the Frobenius and nuclear norm with the hinge loss. To improve the efficiency of SMM, taking motivation from the support tensor machine (STM) \cite{tao2005supervised, cai2006support}, improved least squares twin SMM (ILS-TSMM) \cite{gaoa2015improved} is proposed. ILS-TSMM takes into account the SRM principle by including the regularization term. It considers the twin idea \cite{khemchandani2007twin} and least squares technique \cite{suykens1999least} to speed up the calculation. The incorporation of the twin idea draws two non-parallel hyperplanes. The article \cite{gaoa2015improved} includes both the linear and non-linear cases of classification for matrix data. ILS-TSMM corresponds to the linear case and its incorporation with the matrix kernel function corresponds to the nonlinear ILS-TSMM  (NILS-TSMM). \par
Another least square approach, known as least square SMM on bilevel programming (BP-LSSMM), is introduced by \citet{wenjingXiabilevel2016}.  
The fundamental tenet of bilevel programming (BP) is that the parameters of the lower-level problem are the decision variables for the upper-level problem, and the optimal solution to the lower-level problem is retaliation to the upper-level problem. The formulation of BP-LSSMM  is solved using the least square technology which has reduced time complexity as compared to the classical SMM.  
 
Further, the idea of least square along with non-parallel hyperplanes is proposed, entitled as non-parallel least square SMM for rolling bearing fault diagnosis (NPLSSMM) \cite{li2020non}. It draws two non-parallel hyperplanes in such a way that the distance between the hyperplanes and the samples corresponding to its class should be as near as possible and the distance with the other class of data points should be one. The least-square loss is considered to impose the penalty on the misclassified points. The obtained optimization problem is solved using the ADMM method. Another similar least square approach for SMM named as least square interactive SMM (LSISMM) is considered in \cite{li2022highly}.    

Further, \citet{liang2022adaptive} introduced adaptive multimodal knowledge transfer matrix machine (AMK-TMM) which incorporates the least square loss in SMM along with transfer learning. Using the few available labeled target data and an equality requirement, the AMK-TSMM is capable of automatically detecting the correlative multiple source models and adaptively weighing them. AMK-TMM incorporates a multimodel adaptation approach that follows leave-one-out cross validation strategy and adaptively selects multiple correlated source model knowledge on the available target training data. For the cases with limited training data, AMK-TMM improves the generalization power of LS-SMM.

Thus, we analyzed various existing SMM models based on the concept of least square which corresponds to a reduction in the training time of the models. Moreover, it leads to the simplicity of the models as compared to the classical SMM. Table \ref{Tab:least} contains the summary of the least square variants of SMM. Further, we will delve into discussions about different robust and sparse models within the SMM framework.

\tiny
\newgeometry{left=2cm,bottom=0.1cm,top=0.1cm}
\begin{landscape}
\begin{longtable}[htbp]{|p{0.12\textwidth}|p{0.13\textwidth}|p{0.35\textwidth}|p{0.10\textwidth}|p{0.28\textwidth}|p{0.34\textwidth}|p{0.16\textwidth}|}
\caption[this is]{\bf{Least square variants of SMM}} 
\label{Tab:least} \\

\endfirsthead
\hline
Model&Author&Characteristics &Loss function & Datasets &  Advantages  &Technique to solve\\
\hline
ILS-TSMM (2015) & \citet{gaoa2015improved} &- &- &Face databases ORL and YALE  & Considers SRM principle. Improved efficiency in terms of time than SMM \cite{luo2015support}. &  Solves system of equations. \\
\hline
BP-LSSMM (2016) & \citet{wenjingXiabilevel2016} & Introduces least square method to solve bilevel programming SMM. &- & PALM400, ORL, Yale datasets & Efficient than SMM & Augmented Lagrange multiplier method is used and system of linear equations is solved.\\
\hline
NPLSSMM (2020) & \citet{li2020non} & Solves matrix classification problem by constructing two non-parallel hyperplanes. & Least square loss & Fault dataset from CWRU.  & Distinguishes the classes by obtaining a maximum margin hyperplane in matrix form, reduced complexity as it solves a system of linear equations. & ADMM  \\
\hline
LSISMM (2022) & \citet{li2022highly} & Constructs non-parallel hyperplanes and uses small infrared thermal images for fault diagnosis. &Least square loss && Flexible to maximise the distance between non-parallel hyperplanes and high computational efficiency than SMM.  &ADMM\\
\hline
AMK-TMM $(2022)$ &\citet{liang2022adaptive} & Introduces a novel adaptive multimodel knowledge (AMK) transfer framework and proposed model with equality constraints. & Least square loss & Dataset \RNumU{4}a of BCI \RNumU{3} and \RNumU{2}a of BCI \RNumU{4}.& Utilise cross-validation using a leave-one-out to automatically find the correlated source domains and their corresponding weights.& ADMM \\
\hline
 \end{longtable}
\end{landscape}
\normalsize
\restoregeometry

\subsection{Robust and sparse SMM}
SMM outperforms SVM in terms of performance due to the preservation of the structural information in the input matrix. However, it considers the hinge loss function and $l_2$ norm in the objective function which reduces the robustness \cite{wu2007robust} and sparsity \cite{tanveer2021sparse}, respectively. Moreover, the input data often contains distortions from measurement artifacts, outliers, and unconventional sources of noise. Consequently,
the obtained classifier may have poor performance.  Thus, in order to counter intra-sample outliers, \citet{zheng2018robust} proposed a robust SMM (RSMM). It decomposed the input matrix into a latent low-rank clean matrix plus a sparse noise matrix and used only the clean matrix for training, which makes it robust to intra-sample outliers. Also, to enhance the sparseness, it employs the $l_1$ norm instead of $l_2$ norm. The $l_1$ in the norm optimization problems tend to drive some of the coefficients to exactly zero and encourage sparse solutions \cite{tanveer2021sparse}.
\par
Other approach to build a robust model is by incorporating a robust classification loss function. In light of this, \citet{qian2019robust} proposed robust multicategory SMM (RMSMM) which makes SMM robust by using the truncated hinge loss function \cite{wu2007robust} rather than hinge loss. The hinge loss is unbounded and can grow indefinitely for outliers away from the optimal hyperplane. In contrast, the truncated hinge loss limits the impact of such outliers by capping the loss at a predefined value. As a result, the truncated hinge loss is resistant to outliers. In a similar way, \citet{gu2021ramp} proposed ramp sparse SMM model (RSSMM) to improve the robustness of SMM. RSSMM uses smooth ramp loss function instead of hinge loss, which also limits the maximum loss and weakens the sensitivity to outliers.
\par
\citet{pan2022twin} proposed a non-parallel SMM named as twin robust matrix machine (TRMM). It has two main characteristics: first, it uses the truncated nuclear norm \cite{hong2016online} to obtain important structural information, and second, it employs the ramp loss function \cite{brooks2011support}, which limits the penalty for outliers and makes TRMM robust to outliers. The nuclear norm employed in the original SMM may be suboptimal and might not be the most optimal choice \cite{chen2013reduced}.
As it aggregates all singular values and treats them equally when minimizing the nuclear norm. However, the significance of a singular value is contingent on the importance of the low-rank details in the matrix \cite{liu2015truncated}. This indicates that a larger singular value corresponds to more crucial low-rank information, whereas smaller singular values might result from less relevant information \cite{jia2018bayesian,dixit2020leveraging}.\\
The majority of the aforementioned models use hinge loss function which is sensitive to noise and unstable to resampling \cite{huang2013support}. To overcome this limitation \citet{feng2022support} incorporated the pinball loss function in the framework of SMM and proposed Pin-SMM. It maximizes the quantile distance rather than the shortest distance, which makes it robust to noise and stable under resampling \cite{huang2013support}.
\par
To emphasize sparsity, \citet{zheng2018sparse} proposed a variant known as sparse SMM (SSMM). This approach simultaneously considers the intrinsic structure of each input matrix and the process of feature selection. Additionally, it introduces a novel regularization term, which combines the $l_1$ norm and the nuclear norm, to enhance the sparsity.
However, SSMM is inadequate to merely artificially extract fault features and choose valuable features. 
To address this challenge, \citet{li2021symplectic} introduced a model termed symplectic weighted sparse SMM (SWSSMM). This approach incorporated the principles of symplectic geometry to generate a symplectic coefficient matrix (SCM) serving as the representation of fault features. This method effectively mitigates the influence of noise present within the input matrix. Moreover, a weight coefficient based on variable entropy is introduced and applied to the SCM to amplify the representation of fault features.
\par
Further, to develop sparse SMM for large-scale problems, \citet{wang2022sparse} proposed the sparse norm matrix machine (SNMM), which constructs a pair of non-parallel hyperplanes and uses the hinge loss function and $l_1$ norm distance. The optimization problem of SNMM avoids the need to find the inverse of the matrix, which makes it suitable for large-scale problems. However, it does not incorporate the effect of past information on current detection. To overcome this limitation, \citet{li2022auto} proposed an auto-correlation function based sparse SMM (ACF-SSMM). It also uses the $l_1$ and performs an auto-correlation function on input previous/current data. This process encapsulates how past information affects current detection and accounts for the memory properties inherent in the input matrix. It represents how previous information influences present detection and captures the memory characteristics of the input matrix.
\par
Here, we delved into a comprehensive analysis of robust and sparse variants within the realm of SMM models. By examining these nuanced approaches, we gained valuable insights into enhancing model robustness and promoting sparsity for improved performance across diverse applications. Table \ref{tablesupple:robust and sparse} presents an overview of the various robust and sparse variants of SMM. In the next subsection, we transition our focus to a distinct yet important domain: multiclass classification within the context of SMM models.

\tiny
\newgeometry{left=2cm,bottom=0.1cm,top=0.1cm}
\begin{landscape}
\begin{longtable}[htbp]{|p{0.12\textwidth}|p{0.13\textwidth}|p{0.35\textwidth}|p{0.10\textwidth}|p{0.28\textwidth}|p{0.34\textwidth}|p{0.16\textwidth}|}
\caption[this is]{\bf{Robust and sparse models of SMM}} 
\label{tablesupple:robust and sparse} \\

\endfirsthead
\hline
Model&Author&Characteristics &Loss function & Datasets &  Advantages  &Technique to solve\\
\hline

RSMM $(2018)$ & \citet{zheng2018robust} & Decompose each input signal into low-rank clean signal and sparse intra-sample outliers and employ $l_1$ norm for sparseness.  & Hinge loss & \RNumU{4}a of brain computer interface (BCI) competition \RNumU{3} \cite{dornhege2004boosting}, \RNumU{2}b and \RNumU{2}a of BCI competition \RNumU{4} \cite{leeb2007brain}  &  Enhances the robustness of SMM. & ADMM\\
\hline
SSMM $(2018)$ &\citet{zheng2018sparse}&   Performs feature selection to remove redundant features and involves a new regularization term, which is a linear combination of nuclear norm and $l_1$ norm. & Hinge loss & INRIA Person Dataset \cite{dalal2005histograms}, Caltech Face Dataset \cite{fergus2003object}, \RNumU{2}a and \RNumU{2}b of BCI competetion \RNumU{4} \cite{ang2012filter} & Enhances the sparseness of SMM.  & Generalized forward-backwards (GFB) algorithm is used to solve the convex QPP.\\
\hline
RMSMM $(2019)$ &\citet{qian2019robust} & Constructed in the angle based classification framework and condenses the binary and multiclass problems into a single
framework.   & Truncated hinge loss  & Daily and sports activities dataset \cite{altun2010human} & Enjoys better prediction permormance and faster computation than SMM. &  Inexact proximal DC algorithm is used to solve the nonconvex optimization problem.     \\
\hline
RSSMM $(2021)$ &\citet{gu2021ramp}& Employs $l_1$ norm and sparse constraint into objective function to weaken the redundant information of the input matrix.   & Smooth ramp loss & Fault dataset of roller bearing from AHUT & Reduces the influence of outliers.&  GFB algorithm \\
\hline
SWSSMM $(2021)$ &\citet{li2021symplectic} & Automatically extract inherent fault features from raw signals and use the symplectic coffecient matrix (SCM). Also a variable entropy-based weight coefficient is added into SCM to enhance the fault features. & Hinge loss & Vibration signal dataset from University of Connecticut  & Eliminates the effect of noise in raw signal and enhance the fault features. & GFB algorithm \\
\hline
TRMM $(2022)$ & \citet{pan2022twin} & Employs truncated nuclear norm for low-rank approximation.  & Ramp loss   & Fault dataset of roller bearing from AHUT &  Insensitive and robust to outliers and efficient than RSMM.& Accelerated Proximal Gradient (ALG) algorithm is used to solve the QPP.\\
\hline
Pin-SMM $(2022)$ & \citet{feng2022support} & Maximizes the quantile distance rather than the shortest distance. & Pinball loss & INRIA Person \cite{dalal2005histograms}, Caltech \cite{fei2006one}, \RNumU{2}a of BCI \RNumU{4}   & Robust to noise.& ADMM \\
\hline 

SNMM $(2022)$ & \citet{wang2022sparse} & Employed the $l_1$ norm distance as a constraint of the hyperplane and avoided the need to find inverse matrices. & Hinge loss & Fault dataset of roller bearing from AHUT, HNU, and CWRU&  Improves the robustness and reduces the storage requirements in the calculation process, making it more suitable for large-scale problems. & The alternating iteration method is used to solve the dual QPP.\\
\hline
ACF-SSMM $(2022)$& \citet{li2022auto} & Extend the  the input matrix by adding data through an
auto-correlation function (ACF) transform, which contain data information at previous/current instants.
&  Hinge loss & SEED-VIG fatigue dataset \cite{zheng2017multimodal} & Enhances the generalization performance of SMM.&  GFB algorithm \\
\hline
 \end{longtable}
\end{landscape}
\normalsize
\restoregeometry

\subsection{SMM for multi-class classification}
In the original formulation, SMM was designed for binary classification problems; however, the majority of problems in the real world are based on multiclass classification \cite{franc2002multi}. To address this challenge, \citet{zheng2018multiclass} developed a model named multiclass SMM (MSMM). This approach incorporates a multiclass hinge loss function along with a regularization term that combines the square Frobenius norm and the nuclear norm. The formulation of the multiclass hinge loss function extends the concept of margin rescaling loss \cite{joachims2009cutting} to accommodate matrix-form data.
To improve the classification performance of MSMM, \citet{razzak2019multiclass} proposed a new multiclass SMM (M-SMM). It constitutes a fusion of binary hinge loss and elastic net penalty. The binary hinge loss employs $C$ functions to emulate multiple binary classifiers, thereby avoiding the need to calculate support vectors between every possible pair of classes.
However, MSMM and M-SMM are not robust to outliers. To develop a robust SMM variant for multiclass problems, \citet{qian2019robust} proposed robust multicategory SMM (RMSMM). It is built using the angle-based classification framework \cite{zhang2014multicategory} and embeds a truncated hinge loss function \cite{wu2007robust}. A common method for multiclass classification is to use $C$ classification functions to stand for the $C$ categories. However, the angle-based classification framework needs to train $C-1$ classification functions and thus it adheres to faster computation \cite{sun2017angle}. The non-convex optimization problem of RMSMM is tackled by the DCA algorithm. 
\par
Taking motivation from the cooperative evolution \cite{rosales2018mc2esvm}, \citet{razzak2020cooperative} proposed a multiclass sparse SMM (MSMM-CE) which breaks down the multiclass problem of SMM into binary subproblems in a cooperative manner.
The objective function of MSMM-CE encompasses a fusion of components. Firstly, it integrates a binary hinge loss term to facilitate model fitting. Additionally, it incorporates both Frobenius and nuclear norms as regularization penalties, aimed at encouraging low-rankness and sparsity within the model. Furthermore, the function includes an extra penalty component designed to penalize errors in multiclass classification.
\par
The above-mentioned models ignore the distinct importance of different samples, making class segmentation difficult using the hyperplane. To address this challenge, \citet{pan2022multi} introduced a novel approach in the domain of matrix modeling, leveraging fuzzy theory \cite{hullermeier2005fuzzy} to formulate the multiclass fuzzy SMM(MFSMM).
This innovative framework incorporates the principles of nonparallel hyperplanes in conjunction with fuzzy attributes to optimize the separation intervals between any two fuzzy hyperplanes. The introduction of fuzzy planes assigns varying membership degrees to individual training samples, effectively diminishing the impact of noise in the process. Notably, this marks the inaugural utilization of fuzzy theory in the development of SMM.
\par
In the aforementioned discussion, we examined the landscape of multiclass classification models. A summary of the multiclass classification variants of SMM has been presented in Table \ref{tablesupple:multiclass}.
Now, we shift our attention to the models that evolved to address the imperative challenge of the class imbalance problem.


\tiny
\newgeometry{left=2cm,bottom=0.1cm,top=0.1cm}
\begin{landscape}
\begin{longtable}[htbp]{|p{0.12\textwidth}|p{0.13\textwidth}|p{0.35\textwidth}|p{0.10\textwidth}|p{0.28\textwidth}|p{0.34\textwidth}|p{0.16\textwidth}|}
\caption[this is]{\bf{Variants of SMM for multi-class classification}} 
\label{tablesupple:multiclass} \\

\endfirsthead
\hline
Model&Author&Characteristics &Loss function & Datasets &  Advantages  &Technique to solve\\
\hline
MSMM $(2018)$ &\citet{zheng2018multiclass} & It consists of a multiclass hinge loss term and a regularization term combined with Frobenius and nuclear norm.& Multiclass hinge loss    & Dataset \RNumU{3}a of BCI \RNumU{3} and \RNumU{2}a of BCI \RNumU{4}. & Improves the performance of BCI system with multiple tasks. & ADMM
   \\
\hline
M-SMM $(2019)$& \citet{razzak2019multiclass}&  Maximizes the intra class margin, i.e., maximizes the distance between training point and hyperplane and  employs $C$ functions to simulate all binary classifier rather than computing support vector between every two class.      & Binary hinge loss & Dataset \RNumU{3}a of BCI \RNumU{3} and \RNumU{2}a of BCI \RNumU{4} & Improves the classification performance for multiclass problems. & ADMM      \\
\hline
RMSMM $(2019)$ &\citet{qian2019robust} & Constructed in the angle based classification framework and condenses the binary and multiclass problems into a single
framework.   & Truncated hinge loss  & Daily and sports activities dataset \cite{altun2010human} & Enjoys better prediction permormance and faster computation than SMM. &  Inexact proximal DC algorithm is used to solve the nonconvex optimization problem.     \\
\hline
MSMM-CE (2020) & \citet{razzak2020cooperative} &  Solves multi-class classification problems,  and reduces data redundancy. &Hinge loss &BCI competitions benchmark EEG datasets (\RNumU{3}a and \RNumU{2}a) & Solves multi-class classification problems by finding support vectors in a single step. & Evolutionary technique to break the complicated optimization problem into simpler single-objective optimization problems. \\
\hline
MFSMM $(2022)$& \citet{pan2022multi}& Fuzzy attributes are introduced to assign different membership degrees to different samples.  & Hinge Loss & Fault dataset of roller bearing from AHUT and fault dataset of roller bearing from Hunan University (HNU) &   SOR method solves the dual problem. \\
\hline

 \end{longtable}
\end{landscape}
\normalsize
\restoregeometry


\subsection{SMM for imbalance learning}
In real-world problems, imbalanced datasets are quite common, where one class (the minority class) has significantly fewer data samples compared to the other class or classes (the majority class or classes) \cite{ganaie2022large}. This imbalance pattern can be observed in various contexts, such as medical diagnosis datasets \cite{majid2014prediction}, email spam detection tasks \cite{tang2006fast}, and so forth. The class imbalance can pose difficulties when using the original SMM as it may get biased toward the majority class and struggle to effectively learn from the minority class \cite{richhariya2020reduced}. To deal with the imbalanced datasets,  \citet{zhu2017entropy} proposed an entropy-based SMM (ESMM).
It incorporates an assessment technique for fuzzy memberships based on entropy, with a specific focus on the importance of certainties within patterns. As a result, ESMM not only guarantees the significance of the minority class but also allocates greater consideration to patterns characterized by higher levels of class certainty. Consequently, when confronted with imbalanced datasets, ESMM is capable of generating a decision surface that exhibits greater adaptability in comparison to the conventional SMM.
\par
Another substantial advancement in addressing the challenges posed by imbalanced datasets comes from the work of \citet{li2021fusion}.
They introduced a fusion framework known as the confidence weighted SMM (CWSMM), which employs dynamic penalty factors tailored to individual class samples. Additionally, they integrated prior knowledge related to matrix samples to devise a strategy for assigning confidence weights. This strategic approach significantly contributes to enhancing the overall resilience of the system. Notably, they harnessed the Dempster-Shafer (D-S) evidence theory \cite{tang2020bearing} to harmoniously integrate the probability output derived from CWSMMs with various measurement methods.
\par
The aforementioned imbalance models utilize the nuclear norm to represent the low-rank characteristics within each matrix data. However, the approach of nuclear norm minimization might not be optimal \cite{zhang2021prediction, liu2019saliency}, resulting in certain limitations in capturing weak correlation information. To address this concern, \citet{xu2022dynamic} introduced dynamic penalty adaptive matrix machine (DPAMM). This approach is built upon the adaptive framework for minimizing low-rank approximations \cite{gao2016novel}. The technique dynamically selects and retains singular values that pertain to highly significant correlation data within the input matrix. Moreover, within the loss function component, a dynamic penalty factor is incorporated, enabling the adjustment of the penalty severity for samples based on the degree of imbalance.
\par
 Table \ref{tablesupple:imbalance} contains a synopsis of the various class imbalance variants of SMM. Having traversed an extensive expanse of SMM variants in the preceding subsections that encompass least squares approaches, robust and sparse formulations, multiclass classification paradigms, and strategies to address class imbalance. In the next subsection, we transition our discussion to deep variants of SMM.

\tiny
\newgeometry{left=2cm,bottom=0.1cm,top=0.1cm}
\begin{landscape}
\begin{longtable}[htbp]{|p{0.12\textwidth}|p{0.13\textwidth}|p{0.35\textwidth}|p{0.10\textwidth}|p{0.28\textwidth}|p{0.34\textwidth}|p{0.16\textwidth}|}
\caption[this is]{\bf{SMM for imbalance learning}} 
\label{tablesupple:imbalance} \\

\endfirsthead
\hline
Model&Author&Characteristics &Loss function & Datasets &  Advantages  &Technique to solve\\
\hline
ESMM $(2017)$ & \citet{zhu2017entropy} &  Entropy-based fuzzy membership is employed. & Hinge loss & Keel imbalance datasets \cite{derrac2015keel} & Better generalization performance on imbalance datasets. &   ADMM \\
\hline
CWSMM $(2021)$ &\citet{li2021fusion} &  Distinct penalty factors for various class samples are meticulously designed, and a strategy for assigning confidence weights is formulated based on prior knowledge of the samples. Using D-S evidence theory a fusion CWSMM is proposed. & Hinge loss & A customized rotating machinery dataset produced by Spectra Quest, Inc., Richmond, USA.& Enhances the robustness to imbalanced data. & ADMM \\
\hline
DPAMM (2022) & \citet{xu2022dynamic}& Introduces the adaptive low-rank regularizer to obtain the low rank information and  adaptively chooses and holds on to the singular values relevant to the matrix's highly significant correlation data. &Hinge loss& Dataset of belevel gear roller bearing fault simulation test rig, datasets collected from fixed-shaft roller bearing test rig, bearing dataset provided by CWRU. & The adaptive low-rank operator adaptively selects the greater singular values concerning the strong correlation low-rank information and improves the performance on imbalanced datasets.  &SOR\\
\hline
 \end{longtable}
\end{landscape}
\normalsize
\restoregeometry


\subsection{Deep variant of SMM}
The shallow variants of SMM have good performance in classifying the matrix data, however, they do not explore the powerful stacked generalization principle for the automatic learning of data. In this subsection, we will highlight the deep variants of the SMM model for classification by using the concept of different layers with stacked generalization \cite{wolpert1992stacked,breiman1996stacked}. 
\par The deep stacked SMM (DSSMM) \cite{hang2020deep} has different layers of SMM which preserve the structural information of the data. The random projections of the predictions of each layer modify the original feature and are passed to the next layer of DSSMM. Instead of parameter fine-tuning via backpropagation, DSSMM uses an effective feed-forward method where each layer forms a convex optimization problem. Also, hinge loss for penalizing misclassification leads to the effectiveness of the model \cite{vinyals2012learning}. Though DSSMM has better performance than SMM, the training of DSSMM model has the issue of using the pre-extracted information which may degrade the classification accuracy of the model as the pre-extracted information may not be sufficient or may be corrupted by noise. The following models overcome the said issue.
\begin{itemize}
    \item  Most deep stacking network (DSN)-based models employ pre-extracted features as input, which has a detrimental effect on learning high-level feature representation when the informative neural patterns are not adequately represented by the input features. Also, the classification process is affected if the pre-extracted features are not accurate. \citet{liang2022deep} introduced deep stacked feature representation (DSFR) using common spatial pattern (CSP) and SMM, which addresses the aforementioned problem by allowing the model to acquire high-level representation and abstraction on its own. 
    The fundamental units of DSFR are feature decoding modules with SMM acting as a classifier and CSP acting as a feature extractor. The stacking element is a random projection of the output from the previous layer. Like DSSMM, DSFR uses a feed-forward method instead of backpropagation for fine-tuning.    
\item To overcome the issue of training the model using insufficient data, deep stacked transfer least square support matrix machine (DST-LSSMM) \cite{hang2023deep} is introduced. It uses the LSSMM as the base building blocks of the DSN and random projection of the previous layers as its stacking element. The original input data along with the randomly projected value of the output of the previous layers are fed to the next layer. The required model parameters are taken from the lower levels with the use of an adaptive multi-layer model knowledge transfer learning system, which makes it easier to build models at higher layers. 
It also uses an efficient feed-forward method instead of fine-tuning and parameter pre-training.
\item To enhance the robustness of the deep SMM models to noise, a deep stacked pinball transfer matrix machine (DSPTMM) \cite{pan2023deep} is proposed. The base building blocks and the stacking element of DSPTMM are pinball transfer modules (PTM) and random projections, respectively. In DSPTMM, PTMs are used to obtain the output of the previous layer which is then randomly projected and combined with the original input matrix, and fed to the next layer of DSPTMM. The use of pinball loss maximizes the quantile distance and provides noise robustness \cite{huang2013support}.      
\end{itemize}
Deep variants of SMM improved its classification performance when the training data is insufficient as well as if the training data is contaminated with outliers/noise. Hence, deep SMM variants proved to be an effective development in the field of matrix classification. To give a brief overview of the deep variant model, we have constructed Table \ref{tab:deep_variants}.  Moving our context to different other variants of SMM, we discuss it in the next section.    

\tiny
\newgeometry{left=2cm,bottom=0.1cm,top=0.1cm}
\begin{landscape}
\begin{longtable}[htbp]{|p{0.12\textwidth}|p{0.13\textwidth}|p{0.35\textwidth}|p{0.10\textwidth}|p{0.28\textwidth}|p{0.34\textwidth}|p{0.16\textwidth}|}
\caption[this is]{\bf{Deep variants of SMM}} 
\label{tab:deep_variants} \\

\endfirsthead
\hline
Model&Author&Characteristics &Loss function & Datasets &  Advantages  &Technique to solve\\
\hline

 DSSMM $(2020)$ & \citet{hang2020deep} &
 Inherits the powerful capability of deep representation learning & Hinge Loss &
BCI competition III Dataset IVa, BCI Competition IV Dataset IIb, BCI Competition IV Dataset IIa, Lower Limb MI-BCI Dataset & Involves an efficient feed-forward rather than parameter fine-tuning with backpropagation,  leads to a convex optimization problem. &  ADMM  \\
 \hline 
DSFR $(2022)$ & \citet{liang2022deep} & Base building blocks of DSFR consist of feature decoding modules which have CSP as a feature extractor and SMM as a classifier.& Hinge loss & Dataset IVa of BCI competition III, Dataset IIb of BCI competition IV, Dataset IIa of BCI competition & Instead of relying on pre-extracted EEG features, this method directly accepts the raw EEG data as input and automatically learns the feature representations. To improve classification performance, FDM can collect structural data from the EEG feature matrix. & ADMM \\
\hline
DST-LSSMM (2023) & \citet{hang2023deep} &Deep stacked network uses LSSMM as the base building unit and the projection of the previous layer is used as the stacking element. & Least square loss & MI-based EEG competition datasets which includes dataset \RNumU{3}a, dataset \RNum{4}a in BCI competition \RNumU{3}, and a self-collected dataset which includes lower limb MI-based BCI dataset \cite{lei2019walking}  & Overcomes non-convexity of deep learning models and requires less training data in comparison to deep learning models. Multiple layers take the privilege of adaptive learning. & Alternating iterative method  \\
\hline
DSPTMM (2023) & \citet{pan2023deep} &Deep stacked network uses pinball transfer module as the base building units and random projections as the stacking element. & Pinball loss & Used in the roller bearing fault diagnosis  & Overcomes the non-convexity of deep learning models and requires less training data in comparison to deep learning models. Takes the privilege of adaptive learning, pinball loss leads to a robust model. & ADMM \\
\hline



 \end{longtable}
\end{landscape}
\normalsize
\restoregeometry


\subsection{Other variants of SMM}
In this subsection, our discussion is focused on different existing variants of SMM which play a major role in improving the generalization performance as compared to classical SMM. They can be listed as:
\begin{itemize}
\item  {\textbf{Quantum SMM:}} The time complexity of SMM is $O(\text{poly}(N,pq))$ \cite{duan2017quantum}. 
If the size of the training input matrix or the number of training samples is high, the complexity gets high. To overcome this limitation, quantum algorithm for SMMs (QSMM) \cite{duan2017quantum} is introduced. In QSMM, the QPP of SMM is converted as linear programming problem using the least square technique. The obtained least square formulation is solved by quantum matrix inversion (QMI), i.e., Harrow-Hassidim-Lloyd (HHL) technique \cite{harrow2009quantum}. Then quantum singular value thresholding (QSVT) is employed to solve singular value thresholding (SVT). The time complexity for the two key steps HHL and QSVT are $O(\text{log}(Npq))$ and $O(\text{log}(pq))$ which is exponentially improved over the classical SMM.       
\citet{zhang2021improved} in 2021 proposed an improvement of QSMM (IQSMM) entitled ``an  improved quantum algorithm for support matrix machines", where they used improved QMI inspite of the HHL algorithm. More precisely, the complexity of QMI (HHL) algorithm is $O(\kappa^3 \epsilon^{-3}\text{log}(Npq))$ \cite{zhang2021improved}, which becomes $O(\kappa^2\text{log}^{1.5}(\kappa/\epsilon)\text{log}(Npq))$ in the improve QSMM. 
 \item  {\textbf{Nonlinear Kernel SMM:}} Motivated by matrix Hilbert space \cite{ye2017matrix}, Yunfei Ye introduced nonlinear kernel SMM (KSMM) \cite{ye2019nonlinear}, which includes a matrix form inner product to preserve the structural information in the matrix data. The matrix form inner product is defined as $\bigg\langle \langle X, X \rangle_H, \frac{V}{\norm{V}} \bigg \rangle \geq 0 $ applied to both linear and non-linear cases. The obtained optimization problem of KSMM is solved using the asymptotically convergent algorithm on SMO \cite{platt1999fast} instead of ADMM that classical SMM uses. It makes use of the structural information of the matrix data and constructs a hyperplane by calculating the weighted average distance from the input training matrices which makes it different from conventional kernel methods \cite{luo2015support}.

 \item {\textbf{Wavelet kernels for SMM:}} The wavelet technique has potential for both classification and approximation of nonstationary signals \cite{zhang1992wavelet}, therefore it can be coupled with classification techniques to enhance generalization performance. Taking motivation from wavelet SVM \cite{zhang2004wavelet}, wavelet kernels for SMM (WSMM)  \cite{maboudou2019wavelet}  are proposed. The two SMM variants introduced with wavelet kernels are the SMM with Mexican Hat wavelet kernel which are translation-invariant wavelet kernels and the Morlet wavelet kernel for SMM.  

\item \textbf{Proximal SMM:} Based on the proximal SVM \cite{fung2001proximal}, \citet{zhang2022proximal} introduced another variant of SMM known as proximal SMM (PSMM). The objective of PSMM is to minimize the Euclidean distance of each proximal plane and its corresponding class. Also, it minimizes the nuclear norm of the regression matrix following the concept of SMM. The formulation obtained for PSMM is simpler than the classical SMM \cite{luo2015support} and has an improved efficiency than SMM. These properties make PSMM effective for complex image classification.  

\item {\textbf{Projection twin SMM:}} Regularized projection twin SVM \cite{shao2013regularization} implements SRM principle and finds two projection directions, one for each class so that the projected samples for each class are well separated from one another. \citet{xu2015projection} employed the idea of finding the projection directions to second order tensor and named it as linear projection twin SMMs (PTSMM). For each of the classes, PTSMM finds a projection axis such that the within-class variance of the projected samples is minimum and the projected samples of the other class disperse as wide as possible. The article \cite{xu2015projection} also contains a new matrix kernel function for the non-linear case of PTSMM. The obtained QPPs of PTSMM are solved by employing successive overrelaxation (SOR) technique.    

\item \textbf{Twin multiple rank SMM:} The conversion of high order tensor data to matrix data through tucker tensor decomposition \cite{kotsia2011support} leads to multiple rank matrices. To classify matrices with multiple ranks in linear and non-linear cases, linear twin multiple rank SMM (LTMRSMM) \cite{gao2016novel} and non-linear TMRSMM (NTMRSMM) \cite{gao2016novel}, respectively, are proposed.    
Based on the idea of multiple rank multi linear SVM (MRMLSVM) \cite{hou2014multiple}, LTMRSMM is introduced which replaces the projecting vectors of MRMLSVM by left and right singular vectors of singular value decomposition (SVD) of the regression matrices and forms the QPPs. The NTMRSMM uses the matrix kernel function of non-linear PTSMM \cite{xu2015projection} and solves QPPs using left and right singular vectors.\par
 Another approach to classify matrices of multiple ranks is proposed by \citet{jiang2018multiple}, entitled multiple rank multi-linear twin support matrix classification machine (MRMLTSMCM). Instead of singular vectors as in LTMRSMM, MRMLTSMCM constructs each decision function by using a pair of projecting matrices. Also, it is more efficient than TWSVM \cite{khemchandani2007twin}, overcomes the overfitting issue and improves classification accuracy.  To minimize the square of within-class sample distances, MRMLTSMCM uses a 2-norm distance metric. The distance of outliers gets inflated due to the square operation. Thus, MRMLTSMCM is sensitive to outliers. Besides, the dual of MRMLTSMCM involves the inversion of a matrix, making it unsuitable for large-scale data. To overcome the aforementioned issues, non-parallel bounded SMM (NPBSMM) \cite{pan2023non} is introduced, where the effect of outliers is reduced due to the constraint norm group (CNG) in the optimization problem. CNG consisting of 1-norm distance of within-class samples and hinge loss, suppresses the impact of outliers on the model and leads to a sparse model. Hence, NPBSMM is suitable for large-scale data as it avoids matrix inversion and has better generalization performance than SMM \cite{luo2015support}.

\item \textbf{SMM using transfer learning:} By incorporating transfer learning \cite{weiss2016survey} in SMM, \citet{chen2020novel} proposed knowledge-leaverage-based SMM (KL-SMM). Along with the data of the target domain in training, KL-SMM uses the information of the model from the source domain. Thus, compensates for the deficiency in training due to labeled target data. Also, the indirect use of source domain knowledge leads to privacy protection. Moreover, the propagation of structural information from the source model to the target model enhances the generalization capability of the model. However, KL-SMM uses hinge loss which considers the shortest distance between the plane and data points, and is sensitive to feature noise \cite{huang2013support}.  Incorporation of pinball loss in KL-SMM leads to pinball transfer SMM (Pin-TSMM) \cite{pan2022pinball} which preserves the advantages of KL-SMM along with reducing the influence of noise on the hyperplane.


 \item {\bf{Multisynchrosqeezing Transform (MSST) and Whale optimization algorithm (WOA)-SMM:}} 
    The classical SMM \cite{luo2015support} has three parameters trade-off parameter ($\zeta$), hyperparameter of ADMM method ($\rho$), nuclear norm constraint ($\lambda$). The selection of optimal parameters is a key step for the final performance of the model. WOA \cite{mirjalili2016whale} is an adaptive parameter selection technique and its incorporation with SMM introduced WOA-SMM \cite{zheng2020fault}. The iterative algorithm used to optimize the SMM parameters in WOA-SMM allows for the adaptive acquisition of the optimal values and solves the issue of the subjective setting of parameters in SMM. WOA-SMM has a prevalence of straightforward manipulation, quick convergence speed, good convergence accuracy, and fewer adjustment parameters. The construction of the characteristics of the input matrix to WOA-SMM is important. If the time-frequency (TF) trait of the original signal is fed directly, WOA-SMM may not converge. To handle this, MSST \cite{yu2018multisynchrosqueezing} is used, which is an iterative reassignment procedure to improve the energy concentration of the TF representation by applying multiple synchrosqueezing transform \cite{thakur2011synchrosqueezing} operations. 
   MSST constructs the characteristics matrix by extracting TF domain features with less time consumption and reduced calculation cost. MSST is an iterative reassignment procedure to improve the energy concentration of the TF representation by applying multiple synchrosqueezing transform \cite{thakur2011synchrosqueezing} operations.

    \item {\bf{Multi-distance SMM:}} SMM captures the structural information of the matrix data by regularizing the regression matrix. \citet{ye2019multi} proposed multi-distance SMM (MDSMM) which is another technique to capture the structural information. Geometrically, MDSMM is different from SMM in the aspect of the optimization problem which includes the concept of multi-distance \cite{ye2019multi} and quantifies the cost function and penalty function using a vector-based distance. Further, an appropriate weight assigned to the entries of a multi-distance array determines their relative importance.  MDSMM has improved generalization performance and faster training as compared to SMM. 
\end{itemize}

Hence, different concepts incorporated into classical SMM have led to improvement in its geometric structure along with generalization performances. The different variants are summarized in Table \ref{tab:differentvariants} for better visualisation of the properties of the model.  

\tiny
\newgeometry{left=2cm,bottom=0.1cm,top=0.1cm}
\begin{landscape}
\begin{longtable}[htbp]{|p{0.12\textwidth}|p{0.13\textwidth}|p{0.35\textwidth}|p{0.10\textwidth}|p{0.28\textwidth}|p{0.34\textwidth}|p{0.16\textwidth}|}
\caption[this is]{\bf{Other variants of SMM}} 
\label{tab:differentvariants} \\

\endfirsthead
\hline
Model&Author&Characteristics &Loss function & Datasets &  Advantages  &Technique to solve\\
\hline
SMM (2015) & \citet{luo2015support} & Spectral elastic net penalty having Frobenius and nuclear norm.  &Hinge loss & EEG alcoholism, EEG emotion, the students face and INRIA person  & Preserves the correlation within a matrix. &  ADMM \\
\hline
PTSMM $(2015)$ & \citet{xu2015projection} & 
    Seeks projection axis of both classes with a minimum within-class variance of each and scattered projected samples of other classes as far as possible.
 & Hinge loss & 2d image classification using ORL, YALE and AR face databases  & 
    Deals with non-linear cases using a new matrix kernel function.  Considers SRM principle.
 &  SOR to solve QPP. \\
\hline
LTMRSMM, NTMRSMM (2016) & \citet{gao2016novel} & Deals with matrix data having multiple ranks. & Hinge loss & Feret, ORL, FingerDB, Palm100 and Ar  &  Reduced computational cost than the multi rank matrix vectorisation method. &Iteratively solving the QPPs.\\
\hline
QSMM (2017) & \citet{duan2017quantum} &The QPP of SMM is transformed to the solution of a system of linear equations by incorporating least square loss and solved using quantum matrix inversion (HHL) and QSVT. & Square loss function & -&Exponential increase of speed over classical SMM. Complexity: $O(\kappa^3\epsilon^{-3}(\text{log}(Npq))$ $+O(\text{log}(pq))$, whereas complexity of SMM $O(\text{poly}(N,pq))$ & HHL and QSVT algorithm \\
\hline
MRMLTSMCM (2018) & \citet{jiang2018multiple} & Extension of TWSVM, uses pairs of projecting matrices to obtain the non-parallel hyperplanes. & Hinge loss & UCI datasets: Sonar, CMC, Hill-valley, Ionosphere, Madelon, Pedestrian, Pollen, FingerDB, Binucleate, RGB  & Efficient than TWSVM, implements SRM principle.  & Optimizing the obtained QPPs alternatively.\\
\hline
KSMM (2018) & \citet{ye2019nonlinear}& Generates a matrix-based hyperplane by computing the weighted average distance. & Hinge loss & ORL face database, the Sheffield Face dataset, Columbia Object Image Library (COIL-20) and the binary alpha digits   & The matrix form inner product exploits the structural information of matrix data and solves optimization problem without using alternating projection method. & SMO\\
\hline
MDSMM $(2019)$& \citet{ye2019multi} & Introduces multi-distance to extract the intrinsic information of input matrix and used vector-based distance to quantify the cost function and penalty function. & Hinge loss & IMM face dataset, the Japanese female facial expression (JAFFE) dataset \cite{lyons1998japanese}, the jochen triesch static hand posture dataset \cite{von1996robust}, the Columbia object image library COIL-$20$ \cite{nene1996columbia}, and the Columbia Object Image Library COIL-$100$ \cite{nene1996columbia}.& Improves the generalization performance. & Alternating projection method is used to solve the optimization problem.\\
\hline 
  WSMM (2019) & \citet{maboudou2019wavelet} & Wavelet kernels introduced for non-linear case. & Hinge loss & EEG alcoholism dataset, INRIA person dataset  &  Obtains Mercer kernel in the matrix space.  Improves performance on the EEG and INRIA datasets.  &QPP is solved using quadratic programming software.\\
\hline
KL-SMM (2020) & \citet{chen2020novel} &Uses the concept of transfer learning.&Hinge loss & Motor Imagery (MI) based EEG datasets & Improved generalization capability of a target domain by leveraging information from the source domain. &ADMM\\
\hline
WOA-SMM $(2020)$ &\citet{zheng2020fault} & Time-frequency domain features are extracted using multisynchrosqueezing transform (MSST) to construct the feature matrix.& Hinge loss & Fault dataset from Case Western Reserve University (CWRU) and Anhui University of Technology (AHUT).& Improves the classification performance, consumes less time and lower calculation cost.& WOA is used to solve the optimization problem.       \\
\hline
IQSMM (2021) & \citet{zhang2021improved} &The QPP of SMM is transformed to the solution of the system of linear equations by incorporating least square loss and solved using improved quantum matrix inversion and QSVT. &Square loss function & -& complexity: $O(\kappa^2(\text{log}^{1.5}(\kappa/\epsilon)$ $\text{log}(Npq))$ $+O(\text{log}(pq))$&  Quantum matrix inversion  and QSVT \\
\hline
PSMM $(2022)$ & \citet{zhang2022proximal} &Constructs proximal hyperplane for the different classes. &-& minst digital database \cite{lecun1998gradient}, MIT face database, INRIA person database \cite{dalal2005histograms}, students face database \cite{nazir2010feature}, JAFFE \cite{lyons1998coding}  &   Simpler formulation than SMM, efficient than SMM in time complexity. &ADMM \\
\hline
NPBSMM (2023) & \citet{pan2023non} &Constrain norm group is introduced in the optimization problem & Hinge loss & AHUT fault dataset of roller bearing.  & Leads to robust and sparse model. Suitable for large-scale data as matrix inversion is not required.   & Dual coordinate descent (DCD)\\
\hline

 \end{longtable}
\end{landscape}
\normalsize
\restoregeometry


\section{SMM for regression}{\label{sec:SMR}}
The concept of plane-based learning applied to regression problems is termed as support vector regression (SVR) \cite{smola2004tutorial}. It is rigorous and has convex QPP applied to find the global optimal solution, which resolves the local minima issue that the neural network model faces \cite{tang2019real}. Taking motivation from SVR,  \citet{yuan2021support} introduced support matrix regression (SMR).
SMR \cite{yuan2021support} applies the idea of matrix input to the regression problems along with preserving the structural information of the matrix data.  The objective of SMR maximizes the margin and minimise the squared Frobenius norm of the matrix, hence reducing the sensitivity of the regression to noisy data and leading to robustness against noise. SMR overcomes the lack of physical degradability in the input data and is effective for time asynchronization issues. 

\section{SMM for semi-supervised learning}{\label{sec:semi_supervised}}
The original SMM model is highly reliant on a large number of labeled datasets. However, in real-world applications, labeled data is not commonly available, which can deteriorate the performance of the model due to a lack of supervised information \cite{bennett1998semi}. To overcome this limitation, semi-supervised learning (SSL) \cite{zhu2005semi} approaches receive great attention from researchers. It uses both labeled and unlabeled data. By employing the SSL approach, \citet{li2023intelligent} first proposed a novel semi-supervised probability SMM (SPSMM).
In SPSMM, a strategy for probability output is designed to compute the class-specific probability for each input data. Also, to address the issue of insufficiently labeled samples, an SSL-based framework is employed to carefully choose unlabeled samples with a significant level of confidence to allocate pseudo labels. 


\tiny
\newgeometry{left=2cm,bottom=0.1cm,top=0.1cm}
\begin{landscape}
\begin{longtable}[htbp]{|p{0.12\textwidth}|p{0.13\textwidth}|p{0.35\textwidth}|p{0.10\textwidth}|p{0.28\textwidth}|p{0.34\textwidth}|p{0.16\textwidth}|}
\caption[this is]{\bf{SMM for regression and semi-supervised learning}} 
\label{tablesupple:averageRaFBIO} \\

\endfirsthead
\hline
Model&Author&Characteristics &Loss function & Datasets &  Advantages  &Technique to solve\\
\hline
SMR (2021) & \citet{yuan2021support} & Incorporates the idea of matrix learning for regression problems & $\epsilon$-insensitive loss & Test distribution grid: single phase IEEE 8-bus system \cite{liao2018urban}, IEEE 123-bus system. utility distribution network modified from \cite{osti_1171386}  & Used for learning power flow mapping and overcomes the lack of physical degradability, thus overfitting. Robust to noise/outliers. effective for time asynchronization issues.& \\
\hline 
SPSMM $(2023)$ & \citet{li2023intelligent} & A strategy based on probability output is utilized to estimate the posterior class probabilities for matrix inputs. Furthermore, a semi-supervised learning framework is implemented to facilitate the transfer of knowledge from unlabeled samples to labeled ones & - & An infrared thermal imaging dataset.    & Mitigate the issue of limited labeled samples and bolster the generalization performance. & SMO \\
\hline 
 \end{longtable}
\end{landscape}
\normalsize
\restoregeometry


\begin{figure*} 
\centering{
\includegraphics[width=.65\textwidth,keepaspectratio]{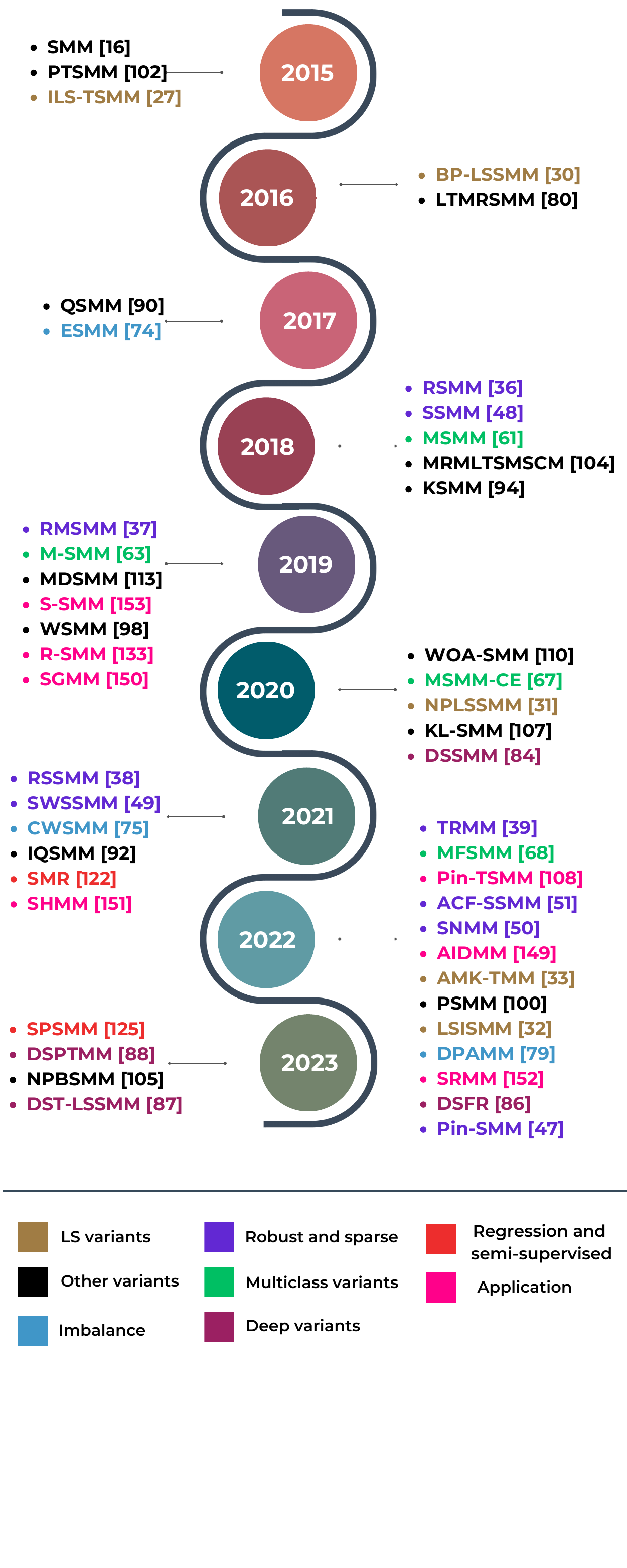}}
\caption{Timeline}
    \label{fig:Timeline}
\end{figure*}

\section{Application}{\label{sec:application}}
SMM has gained significant attention due to its exceptional capabilities in handling complex and high-dimensional data. The application of SMM spreads across a wide range of fields that deal with multi-dimensional data which can be represented as matrices. In many classification problems such as EEG classification, fault diagnosis and image classification, the input features present high dimensionality and are represented as matrices. SMM helps to encapsulate the structural information of the feature matrix by correlating the useful information provided by the rows and columns. In this section, we will discuss the diverse applications of SMM across various domains, showcasing its effectiveness in solving real-world problems and its potential for future advancements.

\begin{figure*} 
   \centering  { %
\includegraphics[width=1.1\textwidth,keepaspectratio]{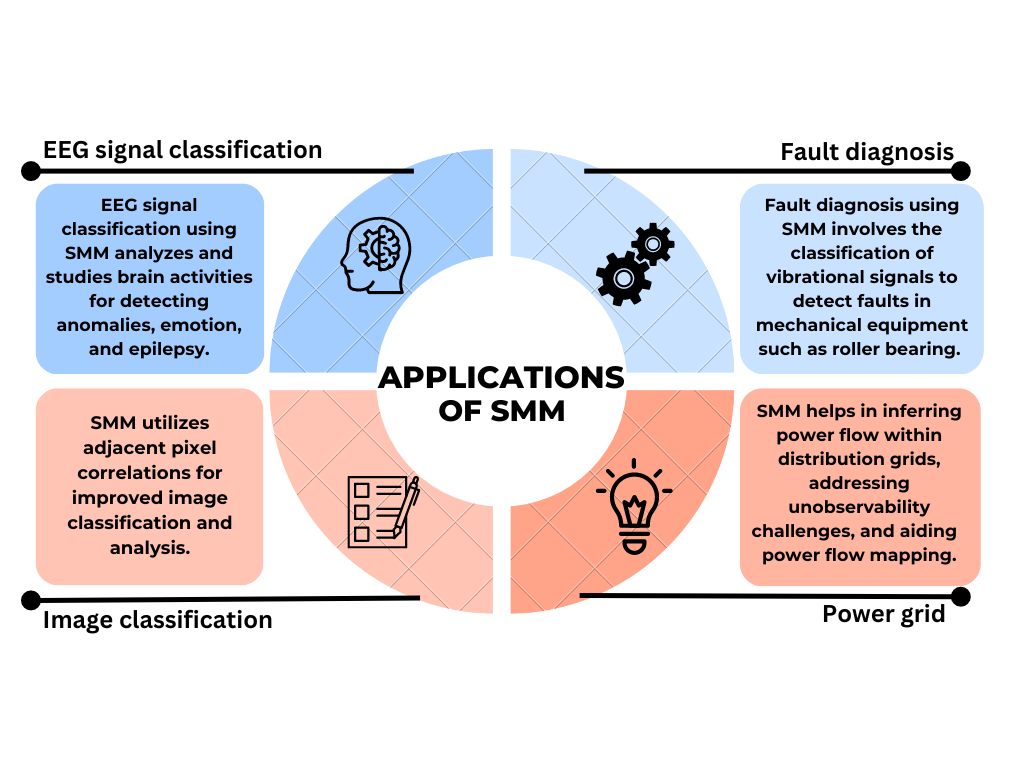}}
\caption{Applications of SMM}
    \label{fig:Application}
      \end{figure*}

\subsection{EEG signal classification}
EEG classification plays a crucial role in various domains, including neuroscience \cite{srinivasan2007cognitive}, clinical diagnostics \cite{praline2007emergent}, BCI \cite{varbu2022past}, and so forth. The application of SMM in classifying EEG signals has shown promising results, which help in analyzing and studying brain activities.\par
EEG utilizes sensors to capture the dynamic electric activity of the brain produced by the synchronous activities of billions of neurons in the brain. Over time, several techniques developed for acquiring EEG signals, and various sensor types are available, such as wet electrodes, dry electrodes, and wireless EEG systems \cite{tyagi2012review}.
 The multidimensional nature of EEG data makes it challenging to comprehend and analyze, requiring a comprehensive understanding and learning process \cite{mumtaz2021review}. This makes SMM a fitting choice for EEG classification tasks. Different EEG datasets used to carry out the application of SMM variants are listed in Table \ref{Table:EEG_datasets}.

\newgeometry{left=2cm,bottom=0.1cm,top=1.5cm}
\begin{footnotesize}
\begin{longtable}[t]{|p{0.18\textwidth}|p{0.28\textwidth}|p{0.10\textwidth}|p{0.14\textwidth}|p{0.14\textwidth}|}
\caption[this is]{\bf{ EEG Datasets}} 
\label{Table:EEG_datasets} \\

\endfirsthead
\hline
Dataset& Description & Sampling Frequency & Papers & Link \\
\hline
BCI competition III dataset IVa (BCIC34a) & 118-channel, 5 subjects (280 trials per subject) during motor imagery tasks involving right-hand or foot movements & 1000Hz & \cite{zheng2018robust,razzak2019robust,hang2020deep,liang2022deep,liang2022adaptive,hang2023deep} & \href{https://www.bbci.de/competition/iii/desc_IVa.html}{Link}  \\
\hline
BCI competition III dataset IIIa (BCIC33a) & 60-channel, single-trail 3 subjects with 4 classes: right hand, left hand, tongue and feet & 250Hz  &  \cite{zheng2018multiclass,razzak2019robust,razzak2019multiclass,hang2023deep} & \href{https://www.bbci.de/competition/iii/desc_IIIa.pdf}{Link}  \\
\hline

BCI competition IV dataset IIa (BCIC32a) & 22-channel, 4 classes: left hand, right hand, feet and tongue, 9 subjects & 250Hz  &  \cite{zheng2018robust,zheng2018multiclass,razzak2019robust,razzak2019multiclass,hang2020deep,chen2020novel,liang2022deep} & \href{https://www.bbci.de/competition/iv/desc_2a.pdf}{Link}  \\
\hline

BCI competition IV dataset IIb (BCIC32b) & 3 bi-polar EEG channels, 2 classes: left hand, right hand, feet and tongue, 9 subjects  & 250Hz  &  \cite{zheng2018robust,razzak2019robust,hang2020deep,chen2020novel,liang2022deep} & \href{https://www.bbci.de/competition/iv/desc_2b.pdf}{Link}  \\
\hline
Lower Limb MI-BCI dataset (LLMI-BCI) & 32-channel self-collected EEG signals from 10 subjects  & 512Hz  &  \cite{hang2020deep,hang2023deep} & -  \\
\hline

The SJTU emotion EEG dataset (SEED-VIG) Fatigue & 17-channel 3 classes: awake, tired, drowsy from 10 subjects  & 512Hz  &  \cite{li2022auto} & \href{https://bcmi.sjtu.edu.cn/~seed/seed-vig.html}{Link} \\
\hline
 \end{longtable}
    \end{footnotesize}

\restoregeometry

Classifying single-trial EEG signals is a challenging task that requires the implementation of different techniques to enhance signal quality. These techniques aim to minimize noise, measurement artifacts, outliers, and irrelevant information in the EEG data \cite{lotte2007review}, ultimately leading to improved accuracy in subsequent analysis. To improve the quality of signals, filtering plays a significant role. Common spatial pattern (CSP) \cite{devlaminck2011multisubject} is used as a preprocessing technique for BCIs. Another preprocessing technique is downsampling the EEG data to reduce the high-computational costs \cite{bischof2021geometric}. 
After preprocessing, the EEG data undergoes feature extraction or feature selection in order to extract the most helpful information from the input data matrix. The standard methods of feature extraction are time domain parameters (TDP) \cite{vidaurre2009time},
fast Fourier transform (FFT) \cite{shakshi2016brain}, principle component analysis (PCA) \cite{kuncheva2013pca}, and so forth \cite{hu2019eeg}. 

The presence of noise and outliers in the EEG data affects the classification of data. To handle this,
 \citet{zheng2018robust} proposed a classifier entitled RSMM for single-trial EEG classification. The preprocessing techniques of the raw data used are Chebyshev Type II filter \cite{sen2023review} followed by spatial filtering using the CSP algorithm. The feature extraction is done using TDP algorithm, which leads to the robustness of the model \cite{nicolas2012brain}. 
The EEG data is highly complex because of the high dimensionality. To overcome this complexity, \citet{razzak2019robust} proposed efficient feature extraction and showed comparative studies on the PCA algorithms, namely robust joint sparse PCA (RJSPCA) and outliers robust PCA (ORPCA) for dimensionality reduction (for simplicity, we denote the model as R-SMM in further paper). The preprocessing technique used is filter bank CSP (FBCSP) followed by the TDP algorithm for feature extraction. PCA is then applied to select the robust features from TDP which is beneficial in dimensionality reduction. \citet{li2022auto} discusses the application of EEG-based fatigue and attention detection by using  SEED-VIG for experimentation. The EEG fatigue signals were classified using the ACF-SSMM \cite{li2022auto} method. This method involves compressing the redundant features by use of the sparse principle.

The aforementioned methods involve binary classification only. However, to address the multiclass classification, a multiclass SMM (MSMM) is developed that aims to improve EEG-based BCIs' performance involving multiple activities \cite{zheng2018multiclass}. 
The preprocessing is based on the techniques used by \citet{ang2012filter}, which employs non-overlapping bandpass filters of the sixth-order Butterworth filter \cite{pise2021comparative} to filter out the artifacts and unrelated signals, followed by CSP to select the most dominant channels. Several techniques are experimented upon for feature extraction, including band powers (BPO), power spectral density (PSD) and TPD, among which TPD led to the best results. 
MSMM \cite{zheng2018multiclass} is the first attempt to handle a multiple-class EEG data classification that promotes a broader range of applications in BCI technology.
To increase the generalization performance in multiclass SMM, 
\citet{razzak2019multiclass} proposed a multiclass SMM (M-SMM) which enhanced and increased the inter-class margins to focus on the single-trial multi-class classification of EEG signals. 
 To mitigate the effect of outliers/noise, spatial filtering has been employed as an effective preprocessing technique to identify discriminative spatial patterns and eliminate uncorrelated information. Specifically, the FBCSP algorithm is used to filter out unrelated sensorimotor rhythms and artifacts by autonomously selecting a subject-specific frequency range for band-pass filtering of the EEG measurements. 

The collection of EEG data is extremely time-consuming and challenging to obtain from the clinical point of view due to the intricacies of recording the data and privacy laws \cite{vaid2015eeg}. Thus, techniques are developed to leverage the available little source data and apply it to the target domain. \citet{chen2020novel} proposed KL-SMM to improve the performance of the EEG signal classification when very little data on the target domain is available. KL-SMM is a five-order Butterworth filter followed by spatial filters as part of the preprocessing techniques.
Another article that addresses classification in cases of insufficient data is AMK-TMM based on the LS-SMM \cite{liang2022adaptive}. The AMK-TMM framework introduces an adaptive approach that uses the leave-one-out cross-validation strategy to identify many correlated source models and their corresponding weights. This enables construction of the target classifier and the identification of the correlated source models to be integrated into a single learning framework. 

Inspired by the deep learning and transfer learning techniques for increasing the performance of the model with less amount of data, \citet{hang2020deep} proposed SMM to be a basic building block of a DSN and introduced DSSMM which uses first-ordered band pass filter for the preprocessing technique. Similar to DSSMM, \citet{liang2022deep} proposed a DSFR method that takes the input in the form of raw EEG data and performs learning directly from it. DSFR method reduces the reliance of the model on pre-extracted EEG features and can extract the features more effectively than CSP followed by classification. \citet{liang2022deep} suggested the filtering of EEG signals using a five-ordered band-pass filter as preprocessing technique.
Further, \citet{hang2023deep} proposed the deep stacked method termed as DST-LSSMM with the building block module made of LSSMM \cite{wenjingXiabilevel2016}. The author performed five-order Butterworth bandpass filtering followed by CSP on it as a preprocessing technique. The spatial filters of the preserved CSP consisted of the initial filter along with the last three filters. Subsequently, the dynamic logarithmic power of the filtered EEG signals is calculated over time. This process yielded a matrix-based representation of EEG features.

The accurate identification of distinct brain states using SMM can contribute to more accurate diagnoses of neurological disorders and assist in patient monitoring. Table \ref{table_Application_EEG} represents different variants of SMM to classify EEG signals. The success achieved in accurately categorizing EEG patterns underscores the significance of this approach in advancing our understanding of brain activity and its applications in healthcare and beyond.
 
\subsection{Fault diagnosis}
Fault diagnosis is essential in various industries, including manufacturing \cite{yan2023review}, automotive \cite{pernestaal2009probabilistic}, aerospace \cite{patton1990fault}, power systems \cite{sekine1992fault} and so forth. The health of these mechanical instruments and equipment directly impacts the life of a machine and production safety. The application of SMM in fault diagnosis has shown significant potential, providing a robust and practical framework for identifying and classifying faults in complex systems. Different datasets used for fault diagnosis are enlisted in Table \ref{tab:Fault diagnosis_dataset}.






\newgeometry{left=2cm,bottom=0.1cm,top=0.1cm}
\begin{landscape}
\begin{footnotesize}
\begin{longtable}[t]{|p{0.16\textwidth}|p{0.16\textwidth}|p{0.20\textwidth}|p{0.20\textwidth}|p{0.20\textwidth}|p{0.28\textwidth}|p{0.24\textwidth}|}
\caption[this is]{\bf{ Application on EEG classification.}} 
\label{table_Application_EEG} \\

\endfirsthead
\hline
Model& Author & Dataset & Metric & Feature extraction & Preprocessing technique & Involves future directions \\
\hline
RSMM (2018) & \citet{zheng2018robust} &BCIC34a, BCIC42b, and BCIC32a & Accuracy & TDP algorithm &  Chebyshev type 2 filter and CSP & Yes \\
\hline
MSMM (2018) & \citet{zheng2018multiclass} &BCIC34a, BCIC42b, and BCIC32a & Kappa coefficient, precision, recall and F-measure & TDP algorithm& Based on \cite{ang2012filter} non-overlapping bandpass ﬁlters of six-order Butter-worth and CSP& Yes \\
\hline
R-SMM (2019) & \citet{razzak2019robust} &BCIC34a, BCIC42a and BCIC42b & Kappa coefficient, precision, recall and F-measure & JSPCA over TDP algorithm & FBCSP & No \\
\hline
M-SMM (2019) & \citet{razzak2019multiclass} &BCIC33a and BCIC42b& Recall, precision, F-measure and kappa coefﬁcient. &TDP algorithm  & FBCSP& No \\
\hline
DSSMM (2020) & \citet{hang2020deep} & BCIC34a, BCIC42b, BCIC42a, LLMI-BCI & Accuracy, F1,
AUC & TDP algorithm & Band-pass ﬁlter using a ﬁfth-order Butterworth ﬁlter and spatial filtering & Yes \\
\hline
KL-SMM (2020) & \citet{chen2020novel} & BCIC42a and BCIC42b & Accuracy, F1,
AUC & - & Five-order Butterworth filter followed by spatial filters & Yes \\
\hline
DSFR (2022) & \citet{liang2022deep} & BCIC34a, BCIC42b, and BCIC42a & Accuracy, F1,
AUC & - & Five-ordered band-pass filter and CSP & Yes \\
\hline
AMK-TMM (2022) & \citet{liang2022adaptive} & BCIC34a, BCIC42a, and LLMI-BCI & Accuracy,
standard deviation, AUC & - & Fifth-order Butterworth filter and CSP& Yes \\
\hline
ACF-SSMM (2022) & \citet{li2022auto} & SEED-VIG dataset & Accuracy, F1,
AUC & TDP Algorithm  &  Bandpass filter and CSP & No \\
\hline
DST-LSSMM (2023) & \citet{hang2023deep} & BCIC33a, BCIC34a and LLMI-BCI & Accuracy, recall, kappa, F-Score & - & Five-order butterworth bandpass filtering and CSP & Yes \\
\hline

 \end{longtable}
    \end{footnotesize}
\end{landscape}
\restoregeometry

Fault diagnosis uses a two-dimensional vibrational signal as an input. However, sometimes the feature matrix may be contaminated which leads to noise and outliers. To mitigate these contaminations, \citet{gu2021ramp} suggested RSSMM which uses MSST as its feature extraction technique and leads to robustness in fault diagnosis in roller bearings.  Similarly, adaptive interactive deviation matrix machine (AIDMM) \cite{pan2022intelligent} also contributes to the sparseness and robustness. A similar application is targetted by TRMM \cite{pan2022twin} which is a non-parallel classifier for fault diagnosis. TRMM uses symplectic geometry similarity transformation (SGST) to extract the two-dimensional feature matrix and is insensitive as well as robust to the outliers, which helps to improve the health of the mechanical equipment by accurate fault diagnosis. Meanwhile, \citet{pan2022multi} proposed MFSMM for the fault diagnosis of roller bearings to diagnose the working state of the roller bearing accurately by the classification of outliers using the concept of fuzzy hyperplanes. To handle complicated roller bearing faults, i.e., the compound roller bearing faults, \citet{li2020non} proposed NPLSSMM which used continuous wavelet transform (CWT) as its feature extraction technique and minimized the effect of outliers. Further, its applications can also be extended to other rotating machinery for fault diagnosis. Application on different rotating machinery includes SWSSMM \cite{li2021symplectic}, which extracts the distinct fault features in the gears directly from the raw vibration signal. Gear fault diagnosis requires professional expertise and knowledge, however the proposed model extracts a symplectic weighted coefficient matrix using symplectic similar transform (SST).  
 
Binary classification SMM models can be applied to binary datasets. \citet{pan2019fault} suggested the multiclass classification model symplectic geometry matrix machine (SGMM) for fault diagnosis of roller bearing, which is robust to noise and outliers. SGST is used to obtain a symplectic geometry coefficient matrix in SGMM that preserves the structural information and removes noise interference while also preventing the convergence problem. 
The time-frequency features of the roller bearings are insufficient to represent the whole information and complete functioning. For a faster and high convergence activity of the optimization algorithm, \citet{zheng2020fault} proposed the combination of the WOA and SMM, which involves MSST time-frequency analysis. The vibration signature from the drive-end bearing is chosen for analysis. An SKF bearing is employed for this purpose, and artificial defects are introduced at individual points on the ball, inner race, and outer race using spark machining techniques.
\citet{pan2021intelligent} proposed an improved version of SMM called symplectic hyperdisk matrix machine (SHMM) for fault diagnosis which suggests the use of SGST to obtain the input matrix in the form of a dimensionless feature matrix. Further, hyperdisk is used in SHMM to cluster the different kinds of data which makes the whole process robust and efficient.

In fault diagnosis, the combination of vibrational data and infrared images as the input data is also explored. \citet{li2021fusion} discussed CWSMM which employs dynamic penalty factors to tackle class imbalance problem by giving different class samples appropriate weights during training. CWSMM is made more robust by using the prior knowledge of matrix samples to design a confidence weight assignment strategy. Vibration data and infrared thermography (IRT) images are gathered from a specialized test setup for diagnosing faults in rotating machinery. 

During industrial practices, there must be a shortage of annotated samples. To mitigate this challenge, \citet{pan2022pinball} discussed the application of Pin-TSMM on fault diagnosis in roller bearings. When dealing with a shortage of annotated samples, the Pin-TSMM approach involves initially training a model with a substantial volume of labeled data from a source domain. Subsequently, this pre-trained model undergoes fine-tuning using a limited amount of data from the target domain. Furthermore, \citet{li2023intelligent} discussed SPSMM based on semi-supervised probability, and infrared imaging is used for gearbox fault diagnosis instead of the vibration signals susceptible to noise. The SSL strategy elevated the problem of insufficient data samples and outliers, and this makes the model robust in the field of fault detection. However, if the data is of large size, NPBSMM \cite{pan2023non} can be used, as it consists of the CNG that enables the large data handling. The CNG also tries to suppress the effect of outliers and helps make the results more sparse. Therefore, the model has an increased ability to fit the given data accurately.

Classical SMM has insufficient probability information which is addressed by symplectic relevance matrix machine (SRMM) \cite{pan2022novel} for the roller bearing fault diagnosis scheme. In the SRMM approach, the input to the classifier is the sample signal matrix. 
The inherent robustness of symplectic geometry analysis contributes to the resilience of SRMM. Further, classical SMM performs poorly in cases of imbalance. Thus to tackle the challenge of imbalance, \citet{xu2022dynamic} proposed DPAMM which can learn from the feature matrix containing the structural information of the vibration signals.

Fault diagnosis can not only be leveraged using vibrational data but also using thermal images. 
Taking this into account, \citet{li2022highly} introduced LSISMM with infrared thermal images. This design effectively harnesses the structural information present in infrared thermal images. SNMM \cite{wang2022sparse} constructs a pair of hyperplanes for solving the fault diagnosis classification problem. \citet{pan2023deep} proposed DSPTMM uses superposition generalization to capture the matrix structural data and arrive at the prediction. DSPTMM employed SGST to process the raw signals as the data is subject to noise to improve the performance on fault diagnosis.

SMM can effectively capture the complex relationships and interactions between features and instances. It allows for accurate fault identification even in systems with intricate fault patterns and can help experts make an informed decision for maintenance and system optimization. Several applications of SMM on fault diagnosis are represented in Table \ref{table:Fault_diagnosis}.

\begin{table}[]
    \centering
    \begin{tabular}{|c|l|}
        \hline
Dataset& Papers  \\
\hline
AHUT Dataset  & \cite{zheng2020fault,gu2021ramp,pan2021intelligent,pan2022twin,pan2022novel,wang2022sparse,pan2023deep,pan2022pinball,pan2022multi,pan2022intelligent,pan2023non}   \\
\hline
CRWU Dataset   &  \cite{pan2019fault,zheng2020fault,pan2022twin,li2020non,pan2021intelligent,wang2022sparse,pan2023deep,pan2022pinball,pan2023non}   \\
\hline

HNU Dataset  &  \cite{li2020non,pan2022twin,pan2022intelligent,pan2023non}   \\
\hline
  Vibration signal dataset & \cite{li2021symplectic}   \\
  from University of Connecticut (UCONN)& \\
\hline
Dataset of Suzhou University (SUZ) &  \cite{gu2021ramp}   \\
\hline
Custom dataset &  \cite{pan2021intelligent,li2023intelligent}   \\
\hline
    \end{tabular}
    \caption{Roller Bearing Fault Diagnosis  Datasets}
    \label{tab:Fault diagnosis_dataset}
\end{table}

\subsection{Other Applications}
SMM can also be applied to image classification tasks and power grid applications \cite{xu2015projection,yuan2021support}, leveraging the relationships between image features, matrix data, and class labels. The image data as an input used to go as a vector in the SVM algorithms but with the advent of SMM, we can now leverage the structural nature of the images and produce better results. For 2d image classification, \citet{xu2015projection} proposed linear projection twin support matrix machine (PTSMM). The author also discussed the non-linear version of PTSMM by introducing a new matrix kernel function which gave good results. The experiments were performed on the ORL, YALE, and AR databases. Similarly, \citet{liu2019polsar} proposed an image classification technique specifically for polarimetric synthetic aperture radar (PolSAR) images. It involves the conversion of the PolSAR images into the matrix by polarimetric scattering coding followed by the matrix classification using S-SMM \cite{liu2019polsar}. The author used PolSAR images from an airborne system (NASA/JPL-Caltech AIRSAR) for experiments. Some other applications include regression techniques. \citet{yuan2021support} proposed a model variational SMR for the imputation of the power flow in the distribution grid. This is the only work that has applications for regression. We have enlisted the applications in Table \ref{table:other_application}.  

The applications of SMM can be extended to different data domains that can be represented in the form of a matrix. It performs well in cases, where the structural information of the input data has to be leveraged and incorporated. As the complexity of the data increases, SMM provides a way to incorporate the correlation of the data in the training, thus resulting in improved performance. Figure \ref{fig:Application} depicts the various applications of SMM.

\newgeometry{left=2cm,bottom=0.1cm,top=0.1cm}
\begin{landscape}
\begin{footnotesize}
\begin{longtable}[t]{|p{0.16\textwidth}|p{0.16\textwidth}|p{0.20\textwidth}|p{0.20\textwidth}|p{0.12\textwidth}|p{0.12\textwidth}|p{0.20\textwidth}|p{0.20\textwidth}|}
\caption[this is]{\bf{ Application on Fault Diagnosis.}} 
\label{table:Fault_diagnosis} \\

\endfirsthead
\hline
Model& Author & Dataset & Metric&Feature extraction & Optimization technique & Hyperparameter tuning & Involves future directions \\
\hline

SGMM (2019) & \citet{pan2019fault} & CWRU dataset & Accuracy& SGST & ADMM & 5-fold cross-validation & Yes  \\
\hline
WOA-SMM (2020) & \citet{zheng2020fault} & CWRU and AHUT datasets & Accuracy& MSST & WOA & 5-fold cross-validation & Yes  \\
\hline
NPLSSMM (2020) & \citet{li2020non} & CWRU and HNU dataset & Accuracy &CWT & ADMM &5-fold cross-validation & Yes  \\
\hline
SWSSMM (2021) &\citet{li2021symplectic} & UCONN & Accuracy& SST& ADMM & 5-fold cross-validation & Yes \\
\hline
RSSMM (2021) & \citet{gu2021ramp} & AHUT and SUZ datasets & Accuracy& MSST& ADMM & 5-fold cross-validation& Yes \\
\hline
SHMM (2021) & \citet{pan2021intelligent} & CWRU, 6 types of roller-bearing data, and 12 types of roller-bearing data of AHUT & Kappa, recall,
precision and
F1, accuracy & &ADMM &5-fold cross-validation & No \\
\hline
CWSMM (2021) & \citet{li2021fusion} &Vibration data and infrared thermography (IRT) images Spectra Quest, Inc., located in Richmond, VA, USA & Accuracy & & ADMM & - & Yes \\
\hline
TRMM (2022) & \citet{pan2022twin} & AHUT, CWRU and HNU datasets & Accuracy,
G-mean, F-
measure,
AUC, kappa,
precision,
sensitivity,
specificity &  SGST& APG & 5-fold cross-validation & Yes \\
\hline
SRMM (2022) & \citet{pan2022novel} & AHUT dataset & Recognition
rate, time,
kappa, accuracy, recall
rate and F1
and statistical
test & SGST & - & 5-fold cross-validation & Yes \\
\hline
DPAMM (2022) & \citet{xu2022dynamic} & Two custom datasets and CWRU dataset. & Specificity,
Gmean, and recall& SGST &ADMM &Grid-search & No \\
\hline
LSISMM (2022) & \citet{li2022highly} & Custom dataset. & Specificity,
Gmean, and recall& & ADMM & 5-fold cross-validation& Yes \\
\hline
MFSMM (2022) & \citet{pan2022multi} & AHUT dataset & 
Precision, recall, F-score, kappa, accuracy and operational
efficiency
 & &ADMM & - & Yes \\
\hline
Pin-TSMM (2022) & \citet{pan2022pinball} & AHUT and CWRU dataset & Accuracy, precision, recall, F1-score, and kappa coefficient &  &ADMM & Grid-search  & Yes \\
\hline

SNMM (2022) & \citet{wang2022sparse} & AHUT and CRWU dataset &  Accuracy, kappa, recall, F1 score and precision. &  &ADMM &5-fold cross-validation & Yes \\
\hline

AIDMM (2022) & \citet{pan2022intelligent} & AHUT and HNU dataset &  Accuracy, kappa, recall, F1 score and time. & & AIDMM & 5-fold cross-validation & Yes  \\
\hline
DSPTMM (2023) & \citet{pan2023deep} & AHUT dataset & Kappa, recall,
precision and
F1, accuracy & &  ADMM &5-fold cross-validation & Yes  \\
\hline
SPSMM (2023) & \citet{li2023intelligent} & Custom dataset & Accuracy & &  ADMM & 5-fold cross-validation & Yes \\
\hline

NPBSMM (2023) & \citet{pan2023non} & AHUT, CWRU, and HNU datasets & Accuracy, kappa, recall and
F1score& &Dual coordinate descent (DCD) algorithm & Grid-search & Yes \\
\hline

 \end{longtable}
    \end{footnotesize}
\end{landscape}

\newgeometry{left=2cm,bottom=0.1cm,top=0.1cm}
\begin{landscape}
\begin{footnotesize}
\begin{longtable}{|p{0.16\textwidth}|p{0.16\textwidth}|p{0.27\textwidth}|p{0.20\textwidth}|p{0.25\textwidth}|}
\caption[this is]{\bf{ Other Applications of SMM}} 
\label{table:other_application} \\

\endfirsthead
\hline
Model& Author & Dataset & Metric & Application  \\
\hline
PTSMM (2015) & \citet{xu2015projection} & ORL, YALE, AR dataset & Accuracy and running time & 2d Image classification \\
\hline
S-SMM (2019) & \citet{liu2019polsar} &  PolSAR images from an airborne system (NASA/JPL-Caltech AIRSAR). &  Accuracy and kappa coefficient & 2d Image classification  \\
\hline

SMR (2021) & \citet{yuan2021support} & - & Accuracy & Calculate the power flow in the distribution grid automatically without any observability   \\
\hline

 \end{longtable}
    \end{footnotesize}
\end{landscape}

\restoregeometry

\section{Conclusion and Future direction}{\label{sec:conclusion}}
\begin{figure*} 
   \centering  { %
\includegraphics[width=1.1\textwidth,keepaspectratio]{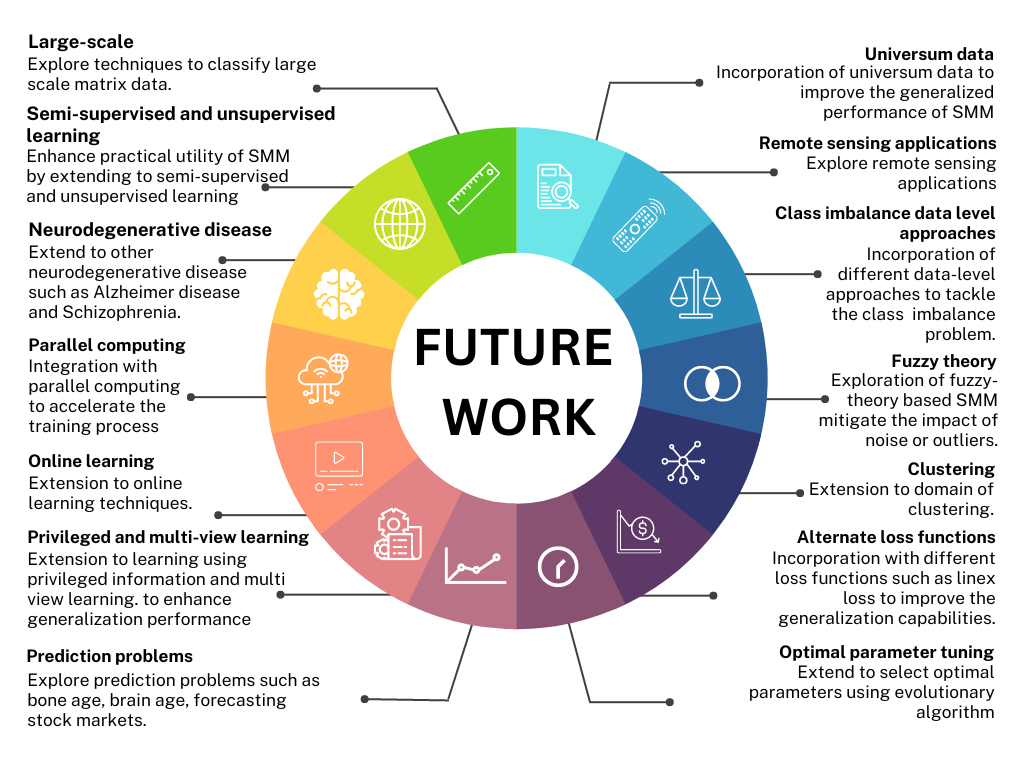}}
\caption{Future work for encompassing challenges and opportunities in SMM.}
    \label{fig:Future work}
      \end{figure*}
      
This paper contains a thorough analysis of the support matrix machine (SMM) algorithm, its variants along with real-world applications. SMM considers Frobenius and nuclear norms in the optimization problem which preserves the structural information of matrix input data. 
The SMM-based models have achieved significant success in various real-world domains because of their property of preserving spatial correlation of input matrices and avoiding the curse of dimensionality as in support vector machine. SMM also achieved cutting-edge performance broadly in the domain of  EEG signal classification and fault diagnosis. The evolution of SMM is discussed in this review study, from its theoretical roots to its diverse uses. To our knowledge, this is the first review article devoted to SMM. We hope that this paper provides vital information about SMM to the researchers. 
While examining the literature, we identified several prospective research areas that the researchers can investigate in the future which are shown in Figure \ref{fig:Future work} and written in detail as follows:
 \begin{itemize}
     \item While SMM exhibits strong performance in classifying matrix data, their applicability to the classification of large-scale data is constrained and warrants additional exploration. Drawing a parallel to the approach involving twin SVM with pinball loss \cite{tanveer2021large}, it is conceivable to formulate a variant of SMM that is resilient to noise and specifically tailored to address the challenges associated with classifying extensive datasets. 
      
     \item Much of the existing research in SMM focuses on supervised learning scenarios, despite the prevalence of unlabeled real-world data. Therefore, there is a promising avenue for extending SMM to address semi-supervised learning, an area with limited exploration, as well as unsupervised learning. This expansion of SMM's scope to encompass unsupervised and semi-supervised problems has the potential to enhance its practicality and utility. 
     \item While SMM has found applications in the biomedical field primarily involving EEG data for tasks like emotion detection, limb/leg movement analysis, and epilepsy dataset analysis, there exists an opportunity to broaden its scope. Specifically, its application could be extended to encompass the detection of other neurodegenerative diseases such as Alzheimer's disease and Schizophrenia. 
     \item The memory demand and computational intricacy stand out as key constraints of SMM. To expedite the training process, parallel computing \cite{zhang2005parallel} can be integrated with SMM. 
     \item While SMM and its variations rely on offline training, which assumes access to the complete training data simultaneously, the context of online scenarios involves sequential streaming data. Consequently, there's an opportunity to concentrate on the advancement of SMM using online learning techniques.
     \item  The notion of incorporating privileged information \cite{vapnik2009new} into the learning process can be extended to SMM which enhance its performance by utilizing supplementary data. 
     Further, multi-view learning is another well-known paradigm \cite{houthuys2018multi} which enhances classification performance by leveraging insights from various perspectives. As a result, extensive investigation of the possibilities of combining SMM with multi-view learning is required in future research. 
     \item The application of SMM in regression problems, remains relatively underexplored which presents an avenue for expansion, potentially by drawing inspiration from approaches like twin support vector regression \cite{huang2022overview}. Moreover, the versatility of SMM can be harnessed for addressing a variety of real-world continuous-variable challenges, such as estimating brain age \cite{ganaie2022brain}, predicting bone age \cite{iglovikov2018paediatric}, and forecasting stock market trends \cite{naeini2010stock}, and so forth. 
     \item    Currently, SMM has not yet advanced to the realm of clustering for categorizing similar data points \cite{winters2007svm}, leaving an unexplored potential for future investigation. 
    \item A wide range of remote sensing image analysis methods are flourishing and being evaluated \cite{mountrakis2011support}. However, SMM at present has no applications for remote sensing data. Consequently, there exists a promising avenue to explore a proficient classifier specifically tailored for remote sensing applications. 
    \item In machine learning, class imbalance is a general problem where one class has significantly fewer samples than the other. This imbalance in data can result in biased models that perform poorly for minority groups \cite{rezvani2023broad}. Moreover, only a few SMM algorithms have been established in view of the class imbalance task. 
    Also, the existing SMM variants effective to handle imbalance class learning are algorithmic level approaches.  Nonetheless, the integration of data-level techniques, including undersampling \cite{laurikkala2001improving, hart1968condensed} and oversampling \cite{chawla2002smote} can enhance the effectiveness of SMM in handling imbalanced datasets. Consequently, to effectively solve the problems caused by class imbalance, a more thorough investigation of these options through greater research efforts is required. 
    \item Fuzzy membership functions can improve the robustness of algorithms by allowing them to deal with noisy or ambiguous data \cite{rezvani2019intuitionistic}. It can capture complex relationships between features and classes that traditional crisp boundaries may not be able to model, leading to improved generalization and predictive accuracy \cite{lee2004first}. The performance of SMM is highly influenced by outliers or noisy samples, resulting in their poor performance. Taking motivation from diverse fuzzy SVM variations, fuzzy theory-based SMM adaptations could offer solutions that mitigate the impact of outliers and noise, thus advancing the field.  
    \item Diverse forms of input data samples offer insights into the distribution of data. By integrating universum data points \cite{weston2006inference}, SMM gets equipped with preexisting knowledge regarding data, leading to improvement in the performance of the classifier. In the future, robust and efficient universum-based SMM can be developed by incorporating prior information through the use of universum data points.
    \item In the context of loss functions, there is limited research on SMM. Only a few loss functions, such as hinge loss \cite{cortes1995support} and pinball loss \cite{huang2013support}, are currently applied in SMM. However, in order to further improve the efficiency, robustness, and generalization capabilities of SMM, there is a compelling need to investigate and incorporate smooth and robust loss functions such as RoBoSS loss \cite{akhtar2023roboss}, Blinex loss \cite{tian2023kernel}, generalized exponential loss \cite{feng2016robust}, and so forth.
    \item The tuning process to select the optimal parameters has a significant impact on the performance of SMM. An adaptive selection of parameters using the whale optimization algorithm (WOA) is proposed in \cite{zheng2020fault}, which circumvents the issue of setting parameters subjectively and enhances classification performance. In further research, more advanced optimization algorithms, such as the evolutionary algorithm \cite{das2010differential} can be combined with WOA to improve classification performance.

 \end{itemize}

\section*{Acknowledgment}
We would like to acknowledge Science and Engineering Research Board (SERB) under
Mathematical Research Impact-Centric Support (MATRICS) scheme grant
no. MTR/2021/000787 for supporting and funding the work. Ms. Anuradha
Kumari (File no - 09/1022 (12437)/2021-EMR-I) and Mr. Mushir Akhtar (File no -09/1022 (13849)/2022-EMR-I) would like to express their appreciation to the Council of Scientific and Industrial Research (CSIR) in
New Delhi, India, for the financial assistance provided as a fellowship.
\bibliographystyle{elsarticle-num-names}
\bibliography{refs.bib}

\begin{thebibliography}{174}
\expandafter\ifx\csname natexlab\endcsname\relax\def\natexlab#1{#1}\fi
\providecommand{\url}[1]{\texttt{#1}}
\providecommand{\href}[2]{#2}
\providecommand{\path}[1]{#1}
\providecommand{\DOIprefix}{doi:}
\providecommand{\ArXivprefix}{arXiv:}
\providecommand{\URLprefix}{URL: }
\providecommand{\Pubmedprefix}{pmid:}
\providecommand{\doi}[1]{\href{http://dx.doi.org/#1}{\path{#1}}}
\providecommand{\Pubmed}[1]{\href{pmid:#1}{\path{#1}}}
\providecommand{\bibinfo}[2]{#2}
\ifx\xfnm\relax \def\xfnm[#1]{\unskip,\space#1}\fi
\bibitem[{Cortes and Vapnik(1995)}]{cortes1995support}
\bibinfo{author}{C.~Cortes}, \bibinfo{author}{V.~Vapnik},
\newblock \bibinfo{title}{Support-vector networks},
\newblock \bibinfo{journal}{Machine Learning} \bibinfo{volume}{20} (\bibinfo{year}{1995}) \bibinfo{pages}{273--297}.
\bibitem[{Cervantes et~al.(2020)Cervantes, Garcia-Lamont, Rodr{\'\i}guez-Mazahua, and Lopez}]{cervantes2020comprehensive}
\bibinfo{author}{J.~Cervantes}, \bibinfo{author}{F.~Garcia-Lamont}, \bibinfo{author}{L.~Rodr{\'\i}guez-Mazahua}, \bibinfo{author}{A.~Lopez},
\newblock \bibinfo{title}{A comprehensive survey on support vector machine classification: Applications, challenges and trends},
\newblock \bibinfo{journal}{Neurocomputing} \bibinfo{volume}{408} (\bibinfo{year}{2020}) \bibinfo{pages}{189--215}.
\bibitem[{Wang et~al.(2005)Wang, Wang, and Lai}]{wang2005new}
\bibinfo{author}{Y.~Wang}, \bibinfo{author}{S.~Wang}, \bibinfo{author}{K.~K. Lai},
\newblock \bibinfo{title}{A new fuzzy support vector machine to evaluate credit risk},
\newblock \bibinfo{journal}{IEEE Transactions on Fuzzy Systems} \bibinfo{volume}{13} (\bibinfo{year}{2005}) \bibinfo{pages}{820--831}.
\bibitem[{Zhu et~al.(2016)Zhu, Liu, Lu, and Li}]{zhu2016efficient}
\bibinfo{author}{H.~Zhu}, \bibinfo{author}{X.~Liu}, \bibinfo{author}{R.~Lu}, \bibinfo{author}{H.~Li},
\newblock \bibinfo{title}{Efficient and privacy-preserving online medical prediagnosis framework using nonlinear {SVM}},
\newblock \bibinfo{journal}{IEEE Journal of Biomedical and Health Informatics} \bibinfo{volume}{21} (\bibinfo{year}{2016}) \bibinfo{pages}{838--850}.
\bibitem[{Zhang et~al.(2008)Zhang, Yoshida, and Tang}]{zhang2008text}
\bibinfo{author}{W.~Zhang}, \bibinfo{author}{T.~Yoshida}, \bibinfo{author}{X.~Tang},
\newblock \bibinfo{title}{Text classification based on multi-word with support vector machine},
\newblock \bibinfo{journal}{Knowledge-Based Systems} \bibinfo{volume}{21} (\bibinfo{year}{2008}) \bibinfo{pages}{879--886}.
\bibitem[{Chuang(2007)}]{chuang2007fuzzy}
\bibinfo{author}{C.-C. Chuang},
\newblock \bibinfo{title}{Fuzzy weighted support vector regression with a fuzzy partition},
\newblock \bibinfo{journal}{IEEE Transactions on Systems, Man, and Cybernetics, Part B (Cybernetics)} \bibinfo{volume}{37} (\bibinfo{year}{2007}) \bibinfo{pages}{630--640}.
\bibitem[{Deng et~al.(2012)Deng, Tian, and Zhang}]{deng2012support}
\bibinfo{author}{N.~Deng}, \bibinfo{author}{Y.~Tian}, \bibinfo{author}{C.~Zhang}, \bibinfo{title}{Support vector machines: optimization based theory, algorithms, and extensions}, \bibinfo{publisher}{CRC press}, \bibinfo{year}{2012}.
\bibitem[{Platt(1998)}]{platt1998sequential}
\bibinfo{author}{J.~Platt},
\newblock \bibinfo{title}{Sequential minimal optimization: A fast algorithm for training support vector machines}  (\bibinfo{year}{1998}).
\bibitem[{Keerthi et~al.(2001)Keerthi, Shevade, Bhattacharyya, and Murthy}]{keerthi2001improvements}
\bibinfo{author}{S.~S. Keerthi}, \bibinfo{author}{S.~K. Shevade}, \bibinfo{author}{C.~Bhattacharyya}, \bibinfo{author}{K.~R.~K. Murthy},
\newblock \bibinfo{title}{Improvements to platt's {SMO} algorithm for {SVM} classifier design},
\newblock \bibinfo{journal}{Neural Computation} \bibinfo{volume}{13} (\bibinfo{year}{2001}) \bibinfo{pages}{637--649}.
\bibitem[{Mangasarian and Musicant(1999)}]{mangasarian1999successive}
\bibinfo{author}{O.~L. Mangasarian}, \bibinfo{author}{D.~R. Musicant},
\newblock \bibinfo{title}{Successive overrelaxation for support vector machines},
\newblock \bibinfo{journal}{IEEE Transactions on Neural Networks} \bibinfo{volume}{10} (\bibinfo{year}{1999}) \bibinfo{pages}{1032--1037}.
\bibitem[{Joachims(1999)}]{joachims1999svmlight}
\bibinfo{author}{T.~Joachims},
\newblock \bibinfo{title}{Svmlight: Support vector machine},
\newblock \bibinfo{journal}{SVM-Light Support Vector Machine http://svmlight. joachims. org/, University of Dortmund} \bibinfo{volume}{19} (\bibinfo{year}{1999}) \bibinfo{pages}{25}.
\bibitem[{Kotsia and Patras(2011)}]{kotsia2011support}
\bibinfo{author}{I.~Kotsia}, \bibinfo{author}{I.~Patras},
\newblock \bibinfo{title}{Support tucker machines},
\newblock in: \bibinfo{booktitle}{CVPR 2011}, \bibinfo{organization}{IEEE}, \bibinfo{year}{2011}, pp. \bibinfo{pages}{633--640}.
\bibitem[{Zhou and Li(2014)}]{zhou2014regularized}
\bibinfo{author}{H.~Zhou}, \bibinfo{author}{L.~Li},
\newblock \bibinfo{title}{Regularized matrix regression},
\newblock \bibinfo{journal}{Journal of the Royal Statistical Society Series B: Statistical Methodology} \bibinfo{volume}{76} (\bibinfo{year}{2014}) \bibinfo{pages}{463--483}.
\bibitem[{Wolf et~al.(2007)Wolf, Jhuang, and Hazan}]{wolf2007modeling}
\bibinfo{author}{L.~Wolf}, \bibinfo{author}{H.~Jhuang}, \bibinfo{author}{T.~Hazan},
\newblock \bibinfo{title}{Modeling appearances with low-rank {SVM}},
\newblock in: \bibinfo{booktitle}{2007 IEEE Conference on Computer Vision and Pattern Recognition}, \bibinfo{organization}{IEEE}, \bibinfo{year}{2007}, pp. \bibinfo{pages}{1--6}.
\bibitem[{Pirsiavash et~al.(2009)Pirsiavash, Ramanan, and Fowlkes}]{pirsiavash2009bilinear}
\bibinfo{author}{H.~Pirsiavash}, \bibinfo{author}{D.~Ramanan}, \bibinfo{author}{C.~Fowlkes},
\newblock \bibinfo{title}{Bilinear classifiers for visual recognition},
\newblock \bibinfo{journal}{Advances in Neural Information Processing Systems} \bibinfo{volume}{22} (\bibinfo{year}{2009}).
\bibitem[{Luo et~al.(2015)Luo, Xie, Zhang, and Li}]{luo2015support}
\bibinfo{author}{L.~Luo}, \bibinfo{author}{Y.~Xie}, \bibinfo{author}{Z.~Zhang}, \bibinfo{author}{W.-J. Li},
\newblock \bibinfo{title}{Support matrix machines},
\newblock in: \bibinfo{booktitle}{International Conference on Machine Learning}, \bibinfo{organization}{PMLR}, \bibinfo{year}{2015}, pp. \bibinfo{pages}{938--947}.
\bibitem[{Kobayashi and Otsu(2012)}]{kobayashi2012efficient}
\bibinfo{author}{T.~Kobayashi}, \bibinfo{author}{N.~Otsu},
\newblock \bibinfo{title}{Efficient optimization for low-rank integrated bilinear classifiers},
\newblock in: \bibinfo{booktitle}{Computer Vision--ECCV 2012: 12th European Conference on Computer Vision, Florence, Italy, October 7-13, 2012, Proceedings, Part II 12}, \bibinfo{organization}{Springer}, \bibinfo{year}{2012}, pp. \bibinfo{pages}{474--487}.
\bibitem[{Huang et~al.(2013)Huang, Nie, and Huang}]{huang2013robust}
\bibinfo{author}{J.~Huang}, \bibinfo{author}{F.~Nie}, \bibinfo{author}{H.~Huang},
\newblock \bibinfo{title}{Robust discrete matrix completion},
\newblock in: \bibinfo{booktitle}{Proceedings of the AAAI Conference on Artificial Intelligence}, volume~\bibinfo{volume}{27}, \bibinfo{year}{2013}, pp. \bibinfo{pages}{424--430}.
\bibitem[{Candes and Recht(2012)}]{candes2012exact}
\bibinfo{author}{E.~Candes}, \bibinfo{author}{B.~Recht},
\newblock \bibinfo{title}{Exact matrix completion via convex optimization},
\newblock \bibinfo{journal}{Communications of the ACM} \bibinfo{volume}{55} (\bibinfo{year}{2012}) \bibinfo{pages}{111--119}.
\bibitem[{Srebro and Shraibman(2005)}]{srebro2005rank}
\bibinfo{author}{N.~Srebro}, \bibinfo{author}{A.~Shraibman},
\newblock \bibinfo{title}{Rank, trace-norm and max-norm},
\newblock in: \bibinfo{booktitle}{International Conference on Computational Learning Theory}, \bibinfo{organization}{Springer}, \bibinfo{year}{2005}, pp. \bibinfo{pages}{545--560}.
\bibitem[{Zou and Hastie(2005)}]{zou2005regularization}
\bibinfo{author}{H.~Zou}, \bibinfo{author}{T.~Hastie},
\newblock \bibinfo{title}{Regularization and variable selection via the elastic net},
\newblock \bibinfo{journal}{Journal of the Royal Statistical Society Series B: Statistical Methodology} \bibinfo{volume}{67} (\bibinfo{year}{2005}) \bibinfo{pages}{301--320}.
\bibitem[{Wang et~al.(2015)Wang, Wang, Hu, and Yan}]{wang2015visual}
\bibinfo{author}{J.~Wang}, \bibinfo{author}{M.~Wang}, \bibinfo{author}{X.~Hu}, \bibinfo{author}{S.~Yan},
\newblock \bibinfo{title}{Visual data denoising with a unified schatten-$p$ norm and $l_q$ norm regularized principal component pursuit},
\newblock \bibinfo{journal}{Pattern Recognition} \bibinfo{volume}{48} (\bibinfo{year}{2015}) \bibinfo{pages}{3135--3144}.
\bibitem[{Goldstein et~al.(2014)Goldstein, O'Donoghue, Setzer, and Baraniuk}]{goldstein2014fast}
\bibinfo{author}{T.~Goldstein}, \bibinfo{author}{B.~O'Donoghue}, \bibinfo{author}{S.~Setzer}, \bibinfo{author}{R.~Baraniuk},
\newblock \bibinfo{title}{Fast alternating direction optimization methods},
\newblock \bibinfo{journal}{SIAM Journal on Imaging Sciences} \bibinfo{volume}{7} (\bibinfo{year}{2014}) \bibinfo{pages}{1588--1623}.
\bibitem[{Cai et~al.(2010)Cai, Cand{\`e}s, and Shen}]{cai2010singular}
\bibinfo{author}{J.-F. Cai}, \bibinfo{author}{E.~J. Cand{\`e}s}, \bibinfo{author}{Z.~Shen},
\newblock \bibinfo{title}{A singular value thresholding algorithm for matrix completion},
\newblock \bibinfo{journal}{SIAM Journal on Optimization} \bibinfo{volume}{20} (\bibinfo{year}{2010}) \bibinfo{pages}{1956--1982}.
\bibitem[{Tao et~al.(2005)Tao, Li, Hu, Maybank, and Wu}]{tao2005supervised}
\bibinfo{author}{D.~Tao}, \bibinfo{author}{X.~Li}, \bibinfo{author}{W.~Hu}, \bibinfo{author}{S.~Maybank}, \bibinfo{author}{X.~Wu},
\newblock \bibinfo{title}{Supervised tensor learning},
\newblock in: \bibinfo{booktitle}{Fifth IEEE International Conference on Data Mining (ICDM'05)}, \bibinfo{organization}{IEEE}, \bibinfo{year}{2005}, pp. \bibinfo{pages}{8--pp}.
\bibitem[{Cai et~al.(2006)Cai, He, Wen, Han, and Ma}]{cai2006support}
\bibinfo{author}{D.~Cai}, \bibinfo{author}{X.~He}, \bibinfo{author}{J.-R. Wen}, \bibinfo{author}{J.~Han}, \bibinfo{author}{W.-Y. Ma},
\newblock \bibinfo{title}{Support tensor machines for text categorization}  (\bibinfo{year}{2006}).
\bibitem[{Gaoa et~al.(2015)Gaoa, Fanb, and Xub}]{gaoa2015improved}
\bibinfo{author}{X.~Gaoa}, \bibinfo{author}{L.~Fanb}, \bibinfo{author}{H.~Xub},
\newblock \bibinfo{title}{Improved least squares twin support matrix machines}  (\bibinfo{year}{2015}).
\bibitem[{Khemchandani et~al.(2007)Khemchandani, Chandra et~al.}]{khemchandani2007twin}
\bibinfo{author}{R.~Khemchandani}, \bibinfo{author}{S.~Chandra}, et~al.,
\newblock \bibinfo{title}{Twin support vector machines for pattern classification},
\newblock \bibinfo{journal}{IEEE Transactions on Pattern Analysis and Machine Intelligence} \bibinfo{volume}{29} (\bibinfo{year}{2007}) \bibinfo{pages}{905--910}.
\bibitem[{Suykens and Vandewalle(1999)}]{suykens1999least}
\bibinfo{author}{J.~A. Suykens}, \bibinfo{author}{J.~Vandewalle},
\newblock \bibinfo{title}{Least squares support vector machine classifiers},
\newblock \bibinfo{journal}{Neural Processing Letters} \bibinfo{volume}{9} (\bibinfo{year}{1999}) \bibinfo{pages}{293--300}.
\bibitem[{Xia and Fan(2016)}]{wenjingXiabilevel2016}
\bibinfo{author}{W.~Xia}, \bibinfo{author}{L.~Fan},
\newblock \bibinfo{title}{Least squares support matrix machines based on bilevel programming},
\newblock \bibinfo{journal}{International Journal of Applied Mathematics and Machine Learning (IJAMML)}  (\bibinfo{year}{2016}) \bibinfo{pages}{1--18}.
\bibitem[{Li et~al.(2020)Li, Yang, Pan, Cheng, and Cheng}]{li2020non}
\bibinfo{author}{X.~Li}, \bibinfo{author}{Y.~Yang}, \bibinfo{author}{H.~Pan}, \bibinfo{author}{J.~Cheng}, \bibinfo{author}{J.~Cheng},
\newblock \bibinfo{title}{Non-parallel least squares support matrix machine for rolling bearing fault diagnosis},
\newblock \bibinfo{journal}{Mechanism and Machine Theory} \bibinfo{volume}{145} (\bibinfo{year}{2020}) \bibinfo{pages}{103676}.
\bibitem[{Li et~al.(2022)Li, Shao, Lu, Xiang, and Cai}]{li2022highly}
\bibinfo{author}{X.~Li}, \bibinfo{author}{H.~Shao}, \bibinfo{author}{S.~Lu}, \bibinfo{author}{J.~Xiang}, \bibinfo{author}{B.~Cai},
\newblock \bibinfo{title}{Highly efficient fault diagnosis of rotating machinery under time-varying speeds using {LSISMM} and small infrared thermal images},
\newblock \bibinfo{journal}{IEEE Transactions on Systems, Man, and Cybernetics: Systems} \bibinfo{volume}{52} (\bibinfo{year}{2022}) \bibinfo{pages}{7328--7340}.
\bibitem[{Liang et~al.(2022)Liang, Hang, Lei, Wang, Qin, Choi, and Zhang}]{liang2022adaptive}
\bibinfo{author}{S.~Liang}, \bibinfo{author}{W.~Hang}, \bibinfo{author}{B.~Lei}, \bibinfo{author}{J.~Wang}, \bibinfo{author}{J.~Qin}, \bibinfo{author}{K.-S. Choi}, \bibinfo{author}{Y.~Zhang},
\newblock \bibinfo{title}{Adaptive multimodel knowledge transfer matrix machine for {EEG} classification},
\newblock \bibinfo{journal}{IEEE Transactions on Neural Networks and Learning Systems}  (\bibinfo{year}{2022}).
\bibitem[{Wu and Liu(2007)}]{wu2007robust}
\bibinfo{author}{Y.~Wu}, \bibinfo{author}{Y.~Liu},
\newblock \bibinfo{title}{Robust truncated hinge loss support vector machines},
\newblock \bibinfo{journal}{Journal of the American Statistical Association} \bibinfo{volume}{102} (\bibinfo{year}{2007}) \bibinfo{pages}{974--983}.
\bibitem[{Tanveer et~al.(2021)Tanveer, Sharma, Rastogi, and Anand}]{tanveer2021sparse}
\bibinfo{author}{M.~Tanveer}, \bibinfo{author}{S.~Sharma}, \bibinfo{author}{R.~Rastogi}, \bibinfo{author}{P.~Anand},
\newblock \bibinfo{title}{Sparse support vector machine with pinball loss},
\newblock \bibinfo{journal}{Transactions on Emerging Telecommunications Technologies} \bibinfo{volume}{32} (\bibinfo{year}{2021}) \bibinfo{pages}{e3820}.
\bibitem[{Zheng et~al.(2018)Zheng, Zhu, and Heng}]{zheng2018robust}
\bibinfo{author}{Q.~Zheng}, \bibinfo{author}{F.~Zhu}, \bibinfo{author}{P.-A. Heng},
\newblock \bibinfo{title}{Robust support matrix machine for single trial {EEG} classification},
\newblock \bibinfo{journal}{IEEE Transactions on Neural Systems and Rehabilitation Engineering} \bibinfo{volume}{26} (\bibinfo{year}{2018}) \bibinfo{pages}{551--562}.
\bibitem[{Qian et~al.(2019)Qian, Tran-Dinh, Fu, Zou, and Liu}]{qian2019robust}
\bibinfo{author}{C.~Qian}, \bibinfo{author}{Q.~Tran-Dinh}, \bibinfo{author}{S.~Fu}, \bibinfo{author}{C.~Zou}, \bibinfo{author}{Y.~Liu},
\newblock \bibinfo{title}{Robust multicategory support matrix machines},
\newblock \bibinfo{journal}{Mathematical Programming} \bibinfo{volume}{176} (\bibinfo{year}{2019}) \bibinfo{pages}{429--463}.
\bibitem[{Gu et~al.(2021)Gu, Zheng, Pan, and Tong}]{gu2021ramp}
\bibinfo{author}{M.~Gu}, \bibinfo{author}{J.~Zheng}, \bibinfo{author}{H.~Pan}, \bibinfo{author}{J.~Tong},
\newblock \bibinfo{title}{Ramp sparse support matrix machine and its application in roller bearing fault diagnosis},
\newblock \bibinfo{journal}{Applied Soft Computing} \bibinfo{volume}{113} (\bibinfo{year}{2021}) \bibinfo{pages}{107928}.
\bibitem[{Pan et~al.(2022)Pan, Xu, Zheng, Tong, and Cheng}]{pan2022twin}
\bibinfo{author}{H.~Pan}, \bibinfo{author}{H.~Xu}, \bibinfo{author}{J.~Zheng}, \bibinfo{author}{J.~Tong}, \bibinfo{author}{J.~Cheng},
\newblock \bibinfo{title}{Twin robust matrix machine for intelligent fault identification of outlier samples in roller bearing},
\newblock \bibinfo{journal}{Knowledge-Based Systems} \bibinfo{volume}{252} (\bibinfo{year}{2022}) \bibinfo{pages}{109391}.
\bibitem[{Hong et~al.(2016)Hong, Wei, Hu, Cai, and He}]{hong2016online}
\bibinfo{author}{B.~Hong}, \bibinfo{author}{L.~Wei}, \bibinfo{author}{Y.~Hu}, \bibinfo{author}{D.~Cai}, \bibinfo{author}{X.~He},
\newblock \bibinfo{title}{Online robust principal component analysis via truncated nuclear norm regularization},
\newblock \bibinfo{journal}{Neurocomputing} \bibinfo{volume}{175} (\bibinfo{year}{2016}) \bibinfo{pages}{216--222}.
\bibitem[{Brooks(2011)}]{brooks2011support}
\bibinfo{author}{J.~P. Brooks},
\newblock \bibinfo{title}{Support vector machines with the ramp loss and the hard margin loss},
\newblock \bibinfo{journal}{Operations research} \bibinfo{volume}{59} (\bibinfo{year}{2011}) \bibinfo{pages}{467--479}.
\bibitem[{Chen et~al.(2013)Chen, Dong, and Chan}]{chen2013reduced}
\bibinfo{author}{K.~Chen}, \bibinfo{author}{H.~Dong}, \bibinfo{author}{K.-S. Chan},
\newblock \bibinfo{title}{Reduced rank regression via adaptive nuclear norm penalization},
\newblock \bibinfo{journal}{Biometrika} \bibinfo{volume}{100} (\bibinfo{year}{2013}) \bibinfo{pages}{901--920}.
\bibitem[{Liu et~al.(2015)Liu, Lai, Zhou, Kuang, and Jin}]{liu2015truncated}
\bibinfo{author}{Q.~Liu}, \bibinfo{author}{Z.~Lai}, \bibinfo{author}{Z.~Zhou}, \bibinfo{author}{F.~Kuang}, \bibinfo{author}{Z.~Jin},
\newblock \bibinfo{title}{A truncated nuclear norm regularization method based on weighted residual error for matrix completion},
\newblock \bibinfo{journal}{IEEE Transactions on Image Processing} \bibinfo{volume}{25} (\bibinfo{year}{2015}) \bibinfo{pages}{316--330}.
\bibitem[{Jia et~al.(2018)Jia, Feng, Wang, Xu, and Zhang}]{jia2018bayesian}
\bibinfo{author}{X.~Jia}, \bibinfo{author}{X.~Feng}, \bibinfo{author}{W.~Wang}, \bibinfo{author}{C.~Xu}, \bibinfo{author}{L.~Zhang},
\newblock \bibinfo{title}{Bayesian inference for adaptive low rank and sparse matrix estimation},
\newblock \bibinfo{journal}{Neurocomputing} \bibinfo{volume}{291} (\bibinfo{year}{2018}) \bibinfo{pages}{71--83}.
\bibitem[{Dixit et~al.(2020)Dixit, Verma, and Raj}]{dixit2020leveraging}
\bibinfo{author}{V.~Dixit}, \bibinfo{author}{P.~Verma}, \bibinfo{author}{P.~Raj},
\newblock \bibinfo{title}{Leveraging tacit knowledge for shipyard facility layout selection using fuzzy set theory},
\newblock \bibinfo{journal}{Expert Systems with Applications} \bibinfo{volume}{158} (\bibinfo{year}{2020}) \bibinfo{pages}{113423}.
\bibitem[{Huang et~al.(2013)Huang, Shi, and Suykens}]{huang2013support}
\bibinfo{author}{X.~Huang}, \bibinfo{author}{L.~Shi}, \bibinfo{author}{J.~A. Suykens},
\newblock \bibinfo{title}{Support vector machine classifier with pinball loss},
\newblock \bibinfo{journal}{IEEE Transactions on Pattern Analysis and Machine Intelligence} \bibinfo{volume}{36} (\bibinfo{year}{2013}) \bibinfo{pages}{984--997}.
\bibitem[{Feng and Xu(2022)}]{feng2022support}
\bibinfo{author}{R.~Feng}, \bibinfo{author}{Y.~Xu},
\newblock \bibinfo{title}{Support matrix machine with pinball loss for classification},
\newblock \bibinfo{journal}{Neural Computing and Applications} \bibinfo{volume}{34} (\bibinfo{year}{2022}) \bibinfo{pages}{18643--18661}.
\bibitem[{Zheng et~al.(2018)Zheng, Zhu, Qin, Chen, and Heng}]{zheng2018sparse}
\bibinfo{author}{Q.~Zheng}, \bibinfo{author}{F.~Zhu}, \bibinfo{author}{J.~Qin}, \bibinfo{author}{B.~Chen}, \bibinfo{author}{P.-A. Heng},
\newblock \bibinfo{title}{Sparse support matrix machine},
\newblock \bibinfo{journal}{Pattern Recognition} \bibinfo{volume}{76} (\bibinfo{year}{2018}) \bibinfo{pages}{715--726}.
\bibitem[{Li et~al.(2021)Li, Yang, Shao, Zhong, Cheng, and Cheng}]{li2021symplectic}
\bibinfo{author}{X.~Li}, \bibinfo{author}{Y.~Yang}, \bibinfo{author}{H.~Shao}, \bibinfo{author}{X.~Zhong}, \bibinfo{author}{J.~Cheng}, \bibinfo{author}{J.~Cheng},
\newblock \bibinfo{title}{Symplectic weighted sparse support matrix machine for gear fault diagnosis},
\newblock \bibinfo{journal}{Measurement} \bibinfo{volume}{168} (\bibinfo{year}{2021}) \bibinfo{pages}{108392}.
\bibitem[{Wang et~al.(2022)Wang, Xu, Pan, Xie, and Zheng}]{wang2022sparse}
\bibinfo{author}{M.~Wang}, \bibinfo{author}{H.~Xu}, \bibinfo{author}{H.~Pan}, \bibinfo{author}{N.~Xie}, \bibinfo{author}{J.~Zheng},
\newblock \bibinfo{title}{Sparse norm matrix machine and its application in roller bearing fault diagnosis},
\newblock \bibinfo{journal}{Measurement Science and Technology} \bibinfo{volume}{33} (\bibinfo{year}{2022}) \bibinfo{pages}{115114}.
\bibitem[{Li et~al.(2022)Li, Wang, and Liu}]{li2022auto}
\bibinfo{author}{Y.~Li}, \bibinfo{author}{D.~Wang}, \bibinfo{author}{F.~Liu},
\newblock \bibinfo{title}{The auto-correlation function aided sparse support matrix machine for {EEG}-based fatigue detection},
\newblock \bibinfo{journal}{IEEE Transactions on Circuits and Systems II: Express Briefs}  (\bibinfo{year}{2022}).
\bibitem[{Dornhege et~al.(2004)Dornhege, Blankertz, Curio, and Muller}]{dornhege2004boosting}
\bibinfo{author}{G.~Dornhege}, \bibinfo{author}{B.~Blankertz}, \bibinfo{author}{G.~Curio}, \bibinfo{author}{K.-R. Muller},
\newblock \bibinfo{title}{Boosting bit rates in noninvasive {EEG} single-trial classifications by feature combination and multiclass paradigms},
\newblock \bibinfo{journal}{IEEE Transactions on Biomedical Engineering} \bibinfo{volume}{51} (\bibinfo{year}{2004}) \bibinfo{pages}{993--1002}.
\bibitem[{Leeb et~al.(2007)Leeb, Lee, Keinrath, Scherer, Bischof, and Pfurtscheller}]{leeb2007brain}
\bibinfo{author}{R.~Leeb}, \bibinfo{author}{F.~Lee}, \bibinfo{author}{C.~Keinrath}, \bibinfo{author}{R.~Scherer}, \bibinfo{author}{H.~Bischof}, \bibinfo{author}{G.~Pfurtscheller},
\newblock \bibinfo{title}{Brain--computer communication: motivation, aim, and impact of exploring a virtual apartment},
\newblock \bibinfo{journal}{IEEE Transactions on Neural Systems and Rehabilitation Engineering} \bibinfo{volume}{15} (\bibinfo{year}{2007}) \bibinfo{pages}{473--482}.
\bibitem[{Dalal and Triggs(2005)}]{dalal2005histograms}
\bibinfo{author}{N.~Dalal}, \bibinfo{author}{B.~Triggs},
\newblock \bibinfo{title}{Histograms of oriented gradients for human detection},
\newblock in: \bibinfo{booktitle}{2005 IEEE computer Society Conference on Computer Vision and Pattern Recognition (CVPR'05)}, volume~\bibinfo{volume}{1}, \bibinfo{organization}{Ieee}, \bibinfo{year}{2005}, pp. \bibinfo{pages}{886--893}.
\bibitem[{Fergus et~al.(2003)Fergus, Perona, and Zisserman}]{fergus2003object}
\bibinfo{author}{R.~Fergus}, \bibinfo{author}{P.~Perona}, \bibinfo{author}{A.~Zisserman},
\newblock \bibinfo{title}{Object class recognition by unsupervised scale-invariant learning},
\newblock in: \bibinfo{booktitle}{2003 IEEE Computer Society Conference on Computer Vision and Pattern Recognition, 2003. Proceedings.}, volume~\bibinfo{volume}{2}, \bibinfo{organization}{IEEE}, \bibinfo{year}{2003}, pp. \bibinfo{pages}{II--II}.
\bibitem[{Ang et~al.(2012)Ang, Chin, Wang, Guan, and Zhang}]{ang2012filter}
\bibinfo{author}{K.~K. Ang}, \bibinfo{author}{Z.~Y. Chin}, \bibinfo{author}{C.~Wang}, \bibinfo{author}{C.~Guan}, \bibinfo{author}{H.~Zhang},
\newblock \bibinfo{title}{Filter bank common spatial pattern algorithm on {BCI} competition iv datasets 2a and 2b},
\newblock \bibinfo{journal}{Frontiers in Neuroscience} \bibinfo{volume}{6} (\bibinfo{year}{2012}) \bibinfo{pages}{39}.
\bibitem[{Altun and Barshan(2010)}]{altun2010human}
\bibinfo{author}{K.~Altun}, \bibinfo{author}{B.~Barshan},
\newblock \bibinfo{title}{Human activity recognition using inertial/magnetic sensor units},
\newblock in: \bibinfo{booktitle}{Human Behavior Understanding: First International Workshop, HBU 2010, Istanbul, Turkey, August 22, 2010. Proceedings 1}, \bibinfo{organization}{Springer}, \bibinfo{year}{2010}, pp. \bibinfo{pages}{38--51}.
\bibitem[{Fei-Fei et~al.(2006)Fei-Fei, Fergus, and Perona}]{fei2006one}
\bibinfo{author}{L.~Fei-Fei}, \bibinfo{author}{R.~Fergus}, \bibinfo{author}{P.~Perona},
\newblock \bibinfo{title}{One-shot learning of object categories},
\newblock \bibinfo{journal}{IEEE Transactions on Pattern Analysis and Machine Intelligence} \bibinfo{volume}{28} (\bibinfo{year}{2006}) \bibinfo{pages}{594--611}.
\bibitem[{Zheng and Lu(2017)}]{zheng2017multimodal}
\bibinfo{author}{W.-L. Zheng}, \bibinfo{author}{B.-L. Lu},
\newblock \bibinfo{title}{A multimodal approach to estimating vigilance using {EEG} and forehead eog},
\newblock \bibinfo{journal}{Journal of Neural Engineering} \bibinfo{volume}{14} (\bibinfo{year}{2017}) \bibinfo{pages}{026017}.
\bibitem[{Franc and Hlav{\'a}c(2002)}]{franc2002multi}
\bibinfo{author}{V.~Franc}, \bibinfo{author}{V.~Hlav{\'a}c},
\newblock \bibinfo{title}{Multi-class support vector machine},
\newblock in: \bibinfo{booktitle}{2002 International Conference on Pattern Recognition}, volume~\bibinfo{volume}{2}, \bibinfo{organization}{IEEE}, \bibinfo{year}{2002}, pp. \bibinfo{pages}{236--239}.
\bibitem[{Zheng et~al.(2018)Zheng, Zhu, Qin, and Heng}]{zheng2018multiclass}
\bibinfo{author}{Q.~Zheng}, \bibinfo{author}{F.~Zhu}, \bibinfo{author}{J.~Qin}, \bibinfo{author}{P.-A. Heng},
\newblock \bibinfo{title}{Multiclass support matrix machine for single-trial {EEG} classification},
\newblock \bibinfo{journal}{Neurocomputing} \bibinfo{volume}{275} (\bibinfo{year}{2018}) \bibinfo{pages}{869--880}.
\bibitem[{Joachims et~al.(2009)Joachims, Finley, and Yu}]{joachims2009cutting}
\bibinfo{author}{T.~Joachims}, \bibinfo{author}{T.~Finley}, \bibinfo{author}{C.-N.~J. Yu},
\newblock \bibinfo{title}{Cutting-plane training of structural {SVM}s},
\newblock \bibinfo{journal}{Machine learning} \bibinfo{volume}{77} (\bibinfo{year}{2009}) \bibinfo{pages}{27--59}.
\bibitem[{Razzak et~al.(2019)Razzak, Blumenstein, and Xu}]{razzak2019multiclass}
\bibinfo{author}{I.~Razzak}, \bibinfo{author}{M.~Blumenstein}, \bibinfo{author}{G.~Xu},
\newblock \bibinfo{title}{Multiclass support matrix machines by maximizing the inter-class margin for single trial {EEG} classification},
\newblock \bibinfo{journal}{IEEE Transactions on Neural Systems and Rehabilitation Engineering} \bibinfo{volume}{27} (\bibinfo{year}{2019}) \bibinfo{pages}{1117--1127}.
\bibitem[{Zhang and Liu(2014)}]{zhang2014multicategory}
\bibinfo{author}{C.~Zhang}, \bibinfo{author}{Y.~Liu},
\newblock \bibinfo{title}{Multicategory angle-based large-margin classification},
\newblock \bibinfo{journal}{Biometrika} \bibinfo{volume}{101} (\bibinfo{year}{2014}) \bibinfo{pages}{625--640}.
\bibitem[{Sun et~al.(2017)Sun, Craig, and Zhang}]{sun2017angle}
\bibinfo{author}{H.~Sun}, \bibinfo{author}{B.~A. Craig}, \bibinfo{author}{L.~Zhang},
\newblock \bibinfo{title}{Angle-based multicategory distance-weighted {SVM}},
\newblock \bibinfo{journal}{The Journal of Machine Learning Research} \bibinfo{volume}{18} (\bibinfo{year}{2017}) \bibinfo{pages}{2981--3001}.
\bibitem[{Rosales-Perez et~al.(2018)Rosales-Perez, Garc{\'\i}a, Terashima-Marin, Coello, and Herrera}]{rosales2018mc2esvm}
\bibinfo{author}{A.~Rosales-Perez}, \bibinfo{author}{S.~Garc{\'\i}a}, \bibinfo{author}{H.~Terashima-Marin}, \bibinfo{author}{C.~A.~C. Coello}, \bibinfo{author}{F.~Herrera},
\newblock \bibinfo{title}{{MC2ESVM}: Multiclass classification based on cooperative evolution of support vector machines},
\newblock \bibinfo{journal}{IEEE Computational Intelligence Magazine} \bibinfo{volume}{13} (\bibinfo{year}{2018}) \bibinfo{pages}{18--29}.
\bibitem[{Razzak(2020)}]{razzak2020cooperative}
\bibinfo{author}{I.~Razzak},
\newblock \bibinfo{title}{Cooperative evolution multiclass support matrix machines},
\newblock in: \bibinfo{booktitle}{2020 International Joint Conference on Neural Networks (IJCNN)}, \bibinfo{organization}{IEEE}, \bibinfo{year}{2020}, pp. \bibinfo{pages}{1--8}.
\bibitem[{Pan et~al.(2022)Pan, Xu, Zheng, Su, and Tong}]{pan2022multi}
\bibinfo{author}{H.~Pan}, \bibinfo{author}{H.~Xu}, \bibinfo{author}{J.~Zheng}, \bibinfo{author}{J.~Su}, \bibinfo{author}{J.~Tong},
\newblock \bibinfo{title}{Multi-class fuzzy support matrix machine for classification in roller bearing fault diagnosis},
\newblock \bibinfo{journal}{Advanced Engineering Informatics} \bibinfo{volume}{51} (\bibinfo{year}{2022}) \bibinfo{pages}{101445}.
\bibitem[{H{\"u}llermeier(2005)}]{hullermeier2005fuzzy}
\bibinfo{author}{E.~H{\"u}llermeier},
\newblock \bibinfo{title}{Fuzzy methods in machine learning and data mining: Status and prospects},
\newblock \bibinfo{journal}{Fuzzy Sets and Systems} \bibinfo{volume}{156} (\bibinfo{year}{2005}) \bibinfo{pages}{387--406}.
\bibitem[{Ganaie et~al.(2022)Ganaie, Tanveer, and Lin}]{ganaie2022large}
\bibinfo{author}{M.~Ganaie}, \bibinfo{author}{M.~Tanveer}, \bibinfo{author}{C.-T. Lin},
\newblock \bibinfo{title}{Large-scale fuzzy least squares twin {SVMs} for class imbalance learning},
\newblock \bibinfo{journal}{IEEE Transactions on Fuzzy Systems} \bibinfo{volume}{30} (\bibinfo{year}{2022}) \bibinfo{pages}{4815--4827}.
\bibitem[{Majid et~al.(2014)Majid, Ali, Iqbal, and Kausar}]{majid2014prediction}
\bibinfo{author}{A.~Majid}, \bibinfo{author}{S.~Ali}, \bibinfo{author}{M.~Iqbal}, \bibinfo{author}{N.~Kausar},
\newblock \bibinfo{title}{Prediction of human breast and colon cancers from imbalanced data using nearest neighbor and support vector machines},
\newblock \bibinfo{journal}{Computer Methods and Programs in Biomedicine} \bibinfo{volume}{113} (\bibinfo{year}{2014}) \bibinfo{pages}{792--808}.
\bibitem[{Tang et~al.(2006)Tang, Krasser, Judge, and Zhang}]{tang2006fast}
\bibinfo{author}{Y.~Tang}, \bibinfo{author}{S.~Krasser}, \bibinfo{author}{P.~Judge}, \bibinfo{author}{Y.-Q. Zhang},
\newblock \bibinfo{title}{Fast and effective spam sender detection with granular {SVM} on highly imbalanced mail server behavior data},
\newblock in: \bibinfo{booktitle}{2006 International Conference on Collaborative Computing: Networking, Applications and Worksharing}, \bibinfo{organization}{IEEE}, \bibinfo{year}{2006}, pp. \bibinfo{pages}{1--6}.
\bibitem[{Richhariya and Tanveer(2020)}]{richhariya2020reduced}
\bibinfo{author}{B.~Richhariya}, \bibinfo{author}{M.~Tanveer},
\newblock \bibinfo{title}{A reduced universum twin support vector machine for class imbalance learning},
\newblock \bibinfo{journal}{Pattern Recognition} \bibinfo{volume}{102} (\bibinfo{year}{2020}) \bibinfo{pages}{107150}.
\bibitem[{Zhu(2017)}]{zhu2017entropy}
\bibinfo{author}{C.~Zhu},
\newblock \bibinfo{title}{Entropy-based support matrix machine},
\newblock in: \bibinfo{booktitle}{Intelligence Science I: Second IFIP TC 12 International Conference, ICIS 2017, Shanghai, China, October 25-28, 2017, Proceedings 2}, \bibinfo{organization}{Springer}, \bibinfo{year}{2017}, pp. \bibinfo{pages}{200--211}.
\bibitem[{Li et~al.(2021)Li, Cheng, Shao, Liu, and Cai}]{li2021fusion}
\bibinfo{author}{X.~Li}, \bibinfo{author}{J.~Cheng}, \bibinfo{author}{H.~Shao}, \bibinfo{author}{K.~Liu}, \bibinfo{author}{B.~Cai},
\newblock \bibinfo{title}{A fusion {CWSMM}-based framework for rotating machinery fault diagnosis under strong interference and imbalanced case},
\newblock \bibinfo{journal}{IEEE Transactions on Industrial Informatics} \bibinfo{volume}{18} (\bibinfo{year}{2021}) \bibinfo{pages}{5180--5189}.
\bibitem[{Tang et~al.(2020)Tang, Gu, Wang, He, Zhang, and Lu}]{tang2020bearing}
\bibinfo{author}{X.~Tang}, \bibinfo{author}{X.~Gu}, \bibinfo{author}{J.~Wang}, \bibinfo{author}{Q.~He}, \bibinfo{author}{F.~Zhang}, \bibinfo{author}{J.~Lu},
\newblock \bibinfo{title}{A bearing fault diagnosis method based on feature selection feedback network and improved {DS} evidence fusion},
\newblock \bibinfo{journal}{IEEE Access} \bibinfo{volume}{8} (\bibinfo{year}{2020}) \bibinfo{pages}{20523--20536}.
\bibitem[{Zhang et~al.(2021)Zhang, Lei, Pan, and Pedrycz}]{zhang2021prediction}
\bibinfo{author}{Y.~Zhang}, \bibinfo{author}{X.~Lei}, \bibinfo{author}{Y.~Pan}, \bibinfo{author}{W.~Pedrycz},
\newblock \bibinfo{title}{Prediction of disease-associated circ{RNA}s via circ{RNA}--disease pair graph and weighted nuclear norm minimization},
\newblock \bibinfo{journal}{Knowledge-Based Systems} \bibinfo{volume}{214} (\bibinfo{year}{2021}) \bibinfo{pages}{106694}.
\bibitem[{Liu et~al.(2019)Liu, Zhou, Liu, Sun, Li, and Fang}]{liu2019saliency}
\bibinfo{author}{B.~Liu}, \bibinfo{author}{Y.~Zhou}, \bibinfo{author}{P.~Liu}, \bibinfo{author}{W.~Sun}, \bibinfo{author}{S.~Li}, \bibinfo{author}{X.~Fang},
\newblock \bibinfo{title}{Saliency detection via double nuclear norm maximization and ensemble manifold regularization},
\newblock \bibinfo{journal}{Knowledge-Based Systems} \bibinfo{volume}{183} (\bibinfo{year}{2019}) \bibinfo{pages}{104850}.
\bibitem[{Xu et~al.(2022)Xu, Pan, Zheng, Liu, and Tong}]{xu2022dynamic}
\bibinfo{author}{H.~Xu}, \bibinfo{author}{H.~Pan}, \bibinfo{author}{J.~Zheng}, \bibinfo{author}{Q.~Liu}, \bibinfo{author}{J.~Tong},
\newblock \bibinfo{title}{Dynamic penalty adaptive matrix machine for the intelligent detection of unbalanced faults in roller bearing},
\newblock \bibinfo{journal}{Knowledge-Based Systems} \bibinfo{volume}{247} (\bibinfo{year}{2022}) \bibinfo{pages}{108779}.
\bibitem[{Gao et~al.(2016)Gao, Fan, and Xu}]{gao2016novel}
\bibinfo{author}{X.~Gao}, \bibinfo{author}{L.~Fan}, \bibinfo{author}{H.~Xu},
\newblock \bibinfo{title}{A novel method for classification of matrix data using twin multiple rank {SMM}s},
\newblock \bibinfo{journal}{Applied Soft Computing} \bibinfo{volume}{48} (\bibinfo{year}{2016}) \bibinfo{pages}{546--562}.
\bibitem[{Derrac et~al.(2015)Derrac, Garcia, Sanchez, and Herrera}]{derrac2015keel}
\bibinfo{author}{J.~Derrac}, \bibinfo{author}{S.~Garcia}, \bibinfo{author}{L.~Sanchez}, \bibinfo{author}{F.~Herrera},
\newblock \bibinfo{title}{{KEEL} data-mining software tool: Data set repository, integration of algorithms and experimental analysis framework},
\newblock \bibinfo{journal}{J. Mult. Valued Logic Soft Comput} \bibinfo{volume}{17} (\bibinfo{year}{2015}).
\bibitem[{Wolpert(1992)}]{wolpert1992stacked}
\bibinfo{author}{D.~H. Wolpert},
\newblock \bibinfo{title}{Stacked generalization},
\newblock \bibinfo{journal}{Neural Networks} \bibinfo{volume}{5} (\bibinfo{year}{1992}) \bibinfo{pages}{241--259}.
\bibitem[{Breiman(1996)}]{breiman1996stacked}
\bibinfo{author}{L.~Breiman},
\newblock \bibinfo{title}{Stacked regressions},
\newblock \bibinfo{journal}{Machine Learning} \bibinfo{volume}{24} (\bibinfo{year}{1996}) \bibinfo{pages}{49--64}.
\bibitem[{Hang et~al.(2020)Hang, Feng, Liang, Wang, Liu, and Choi}]{hang2020deep}
\bibinfo{author}{W.~Hang}, \bibinfo{author}{W.~Feng}, \bibinfo{author}{S.~Liang}, \bibinfo{author}{Q.~Wang}, \bibinfo{author}{X.~Liu}, \bibinfo{author}{K.-S. Choi},
\newblock \bibinfo{title}{Deep stacked support matrix machine-based representation learning for motor imagery {EEG} classification},
\newblock \bibinfo{journal}{Computer Methods and Programs in Biomedicine} \bibinfo{volume}{193} (\bibinfo{year}{2020}) \bibinfo{pages}{105466}.
\bibitem[{Vinyals et~al.(2012)Vinyals, Jia, Deng, and Darrell}]{vinyals2012learning}
\bibinfo{author}{O.~Vinyals}, \bibinfo{author}{Y.~Jia}, \bibinfo{author}{L.~Deng}, \bibinfo{author}{T.~Darrell},
\newblock \bibinfo{title}{Learning with recursive perceptual representations},
\newblock \bibinfo{journal}{Advances in Neural Information Processing Systems} \bibinfo{volume}{25} (\bibinfo{year}{2012}).
\bibitem[{Liang et~al.(2022)Liang, Hang, Yin, Shen, Wang, Qin, Choi, and Zhang}]{liang2022deep}
\bibinfo{author}{S.~Liang}, \bibinfo{author}{W.~Hang}, \bibinfo{author}{M.~Yin}, \bibinfo{author}{H.~Shen}, \bibinfo{author}{Q.~Wang}, \bibinfo{author}{J.~Qin}, \bibinfo{author}{K.-S. Choi}, \bibinfo{author}{Y.~Zhang},
\newblock \bibinfo{title}{Deep {EEG} feature learning via stacking common spatial pattern and support matrix machine},
\newblock \bibinfo{journal}{Biomedical Signal Processing and Control} \bibinfo{volume}{74} (\bibinfo{year}{2022}) \bibinfo{pages}{103531}.
\bibitem[{Hang et~al.(2023)Hang, Li, Yin, Liang, Shen, Wang, Qin, and Choi}]{hang2023deep}
\bibinfo{author}{W.~Hang}, \bibinfo{author}{Z.~Li}, \bibinfo{author}{M.~Yin}, \bibinfo{author}{S.~Liang}, \bibinfo{author}{H.~Shen}, \bibinfo{author}{Q.~Wang}, \bibinfo{author}{J.~Qin}, \bibinfo{author}{K.-S. Choi},
\newblock \bibinfo{title}{Deep stacked least square support matrix machine with adaptive multi-layer transfer for {EEG} classification},
\newblock \bibinfo{journal}{Biomedical Signal Processing and Control} \bibinfo{volume}{82} (\bibinfo{year}{2023}) \bibinfo{pages}{104579}.
\bibitem[{Pan et~al.(2023)Pan, Sheng, Xu, Zheng, Tong, and Niu}]{pan2023deep}
\bibinfo{author}{H.~Pan}, \bibinfo{author}{L.~Sheng}, \bibinfo{author}{H.~Xu}, \bibinfo{author}{J.~Zheng}, \bibinfo{author}{J.~Tong}, \bibinfo{author}{L.~Niu},
\newblock \bibinfo{title}{Deep stacked pinball transfer matrix machine with its application in roller bearing fault diagnosis},
\newblock \bibinfo{journal}{Engineering Applications of Artificial Intelligence} \bibinfo{volume}{121} (\bibinfo{year}{2023}) \bibinfo{pages}{105991}.
\bibitem[{Lei et~al.(2019)Lei, Liu, Liang, Hang, Wang, Choi, and Qin}]{lei2019walking}
\bibinfo{author}{B.~Lei}, \bibinfo{author}{X.~Liu}, \bibinfo{author}{S.~Liang}, \bibinfo{author}{W.~Hang}, \bibinfo{author}{Q.~Wang}, \bibinfo{author}{K.-S. Choi}, \bibinfo{author}{J.~Qin},
\newblock \bibinfo{title}{Walking imagery evaluation in brain computer interfaces via a multi-view multi-level deep polynomial network},
\newblock \bibinfo{journal}{IEEE Transactions on Neural Systems and Rehabilitation Engineering} \bibinfo{volume}{27} (\bibinfo{year}{2019}) \bibinfo{pages}{497--506}.
\bibitem[{Duan et~al.(2017)Duan, Yuan, Liu, and Li}]{duan2017quantum}
\bibinfo{author}{B.~Duan}, \bibinfo{author}{J.~Yuan}, \bibinfo{author}{Y.~Liu}, \bibinfo{author}{D.~Li},
\newblock \bibinfo{title}{Quantum algorithm for support matrix machines},
\newblock \bibinfo{journal}{Physical Review A} \bibinfo{volume}{96} (\bibinfo{year}{2017}) \bibinfo{pages}{032301}.
\bibitem[{Harrow et~al.(2009)Harrow, Hassidim, and Lloyd}]{harrow2009quantum}
\bibinfo{author}{A.~W. Harrow}, \bibinfo{author}{A.~Hassidim}, \bibinfo{author}{S.~Lloyd},
\newblock \bibinfo{title}{Quantum algorithm for linear systems of equations},
\newblock \bibinfo{journal}{Physical Review Letters} \bibinfo{volume}{103} (\bibinfo{year}{2009}) \bibinfo{pages}{150502}.
\bibitem[{Zhang et~al.(2021)Zhang, Song, and Wu}]{zhang2021improved}
\bibinfo{author}{Y.~Zhang}, \bibinfo{author}{T.~Song}, \bibinfo{author}{Z.~Wu},
\newblock \bibinfo{title}{An improved quantum algorithm for support matrix machines},
\newblock \bibinfo{journal}{Quantum Information Processing} \bibinfo{volume}{20} (\bibinfo{year}{2021}) \bibinfo{pages}{1--12}.
\bibitem[{Ye(2017)}]{ye2017matrix}
\bibinfo{author}{Y.~Ye},
\newblock \bibinfo{title}{The matrix hilbert space and its application to matrix learning},
\newblock \bibinfo{journal}{arXiv preprint arXiv:1706.08110}  (\bibinfo{year}{2017}).
\bibitem[{Ye(2019)}]{ye2019nonlinear}
\bibinfo{author}{Y.~Ye},
\newblock \bibinfo{title}{A nonlinear kernel support matrix machine for matrix learning},
\newblock \bibinfo{journal}{International Journal of Machine Learning and Cybernetics} \bibinfo{volume}{10} (\bibinfo{year}{2019}) \bibinfo{pages}{2725--2738}.
\bibitem[{Platt(1999)}]{platt1999fast}
\bibinfo{author}{J.~C. Platt},
\newblock \bibinfo{title}{Fast training of support vector machines using sequential minimal optimization, advances in kernel methods},
\newblock \bibinfo{journal}{Support Vector Learning}  (\bibinfo{year}{1999}) \bibinfo{pages}{185--208}.
\bibitem[{Zhang and Benveniste(1992)}]{zhang1992wavelet}
\bibinfo{author}{Q.~Zhang}, \bibinfo{author}{A.~Benveniste},
\newblock \bibinfo{title}{Wavelet networks},
\newblock \bibinfo{journal}{IEEE Transactions on Neural Networks} \bibinfo{volume}{3} (\bibinfo{year}{1992}) \bibinfo{pages}{889--898}.
\bibitem[{Zhang et~al.(2004)Zhang, Zhou, and Jiao}]{zhang2004wavelet}
\bibinfo{author}{L.~Zhang}, \bibinfo{author}{W.~Zhou}, \bibinfo{author}{L.~Jiao},
\newblock \bibinfo{title}{Wavelet support vector machine},
\newblock \bibinfo{journal}{IEEE Transactions on Systems, Man, and Cybernetics, Part B (Cybernetics)} \bibinfo{volume}{34} (\bibinfo{year}{2004}) \bibinfo{pages}{34--39}.
\bibitem[{Maboudou-Tchao(2019)}]{maboudou2019wavelet}
\bibinfo{author}{E.~M. Maboudou-Tchao},
\newblock \bibinfo{title}{Wavelet kernels for support matrix machines},
\newblock \bibinfo{journal}{Modern Statistical Methods for Spatial and Multivariate Data}  (\bibinfo{year}{2019}) \bibinfo{pages}{75--93}.
\bibitem[{Fung and Mangasarian(2001)}]{fung2001proximal}
\bibinfo{author}{G.~Fung}, \bibinfo{author}{O.~L. Mangasarian},
\newblock \bibinfo{title}{Proximal support vector machine classifiers},
\newblock in: \bibinfo{booktitle}{Proceedings of the Seventh ACM SIGKDD International Conference on Knowledge Discovery and Data mining}, \bibinfo{year}{2001}, pp. \bibinfo{pages}{77--86}.
\bibitem[{Zhang and Liu(2022)}]{zhang2022proximal}
\bibinfo{author}{W.~Zhang}, \bibinfo{author}{Y.~Liu},
\newblock \bibinfo{title}{Proximal support matrix machine},
\newblock \bibinfo{journal}{Journal of Applied Mathematics and Physics} \bibinfo{volume}{10} (\bibinfo{year}{2022}) \bibinfo{pages}{2268--2291}.
\bibitem[{Shao et~al.(2013)Shao, Wang, Chen, and Deng}]{shao2013regularization}
\bibinfo{author}{Y.-H. Shao}, \bibinfo{author}{Z.~Wang}, \bibinfo{author}{W.-J. Chen}, \bibinfo{author}{N.-Y. Deng},
\newblock \bibinfo{title}{A regularization for the projection twin support vector machine},
\newblock \bibinfo{journal}{Knowledge-Based Systems} \bibinfo{volume}{37} (\bibinfo{year}{2013}) \bibinfo{pages}{203--210}.
\bibitem[{Xu et~al.(2015)Xu, Fan, and Gao}]{xu2015projection}
\bibinfo{author}{H.~Xu}, \bibinfo{author}{L.~Fan}, \bibinfo{author}{X.~Gao},
\newblock \bibinfo{title}{Projection twin {SMMs} for 2d image data classification},
\newblock \bibinfo{journal}{Neural Computing and Applications} \bibinfo{volume}{26} (\bibinfo{year}{2015}) \bibinfo{pages}{91--100}.
\bibitem[{Hou et~al.(2014)Hou, Nie, Zhang, Yi, and Wu}]{hou2014multiple}
\bibinfo{author}{C.~Hou}, \bibinfo{author}{F.~Nie}, \bibinfo{author}{C.~Zhang}, \bibinfo{author}{D.~Yi}, \bibinfo{author}{Y.~Wu},
\newblock \bibinfo{title}{Multiple rank multi-linear {SVM} for matrix data classification},
\newblock \bibinfo{journal}{Pattern Recognition} \bibinfo{volume}{47} (\bibinfo{year}{2014}) \bibinfo{pages}{454--469}.
\bibitem[{Jiang and Yang(2018)}]{jiang2018multiple}
\bibinfo{author}{R.~Jiang}, \bibinfo{author}{Z.-X. Yang},
\newblock \bibinfo{title}{Multiple rank multi-linear twin support matrix classification machine},
\newblock \bibinfo{journal}{Journal of Intelligent \& Fuzzy Systems} \bibinfo{volume}{35} (\bibinfo{year}{2018}) \bibinfo{pages}{5741--5754}.
\bibitem[{Pan et~al.(2023)Pan, Xu, Zheng, and Tong}]{pan2023non}
\bibinfo{author}{H.~Pan}, \bibinfo{author}{H.~Xu}, \bibinfo{author}{J.~Zheng}, \bibinfo{author}{J.~Tong},
\newblock \bibinfo{title}{Non-parallel bounded support matrix machine and its application in roller bearing fault diagnosis},
\newblock \bibinfo{journal}{Information Sciences}  (\bibinfo{year}{2023}).
\bibitem[{Weiss et~al.(2016)Weiss, Khoshgoftaar, and Wang}]{weiss2016survey}
\bibinfo{author}{K.~Weiss}, \bibinfo{author}{T.~M. Khoshgoftaar}, \bibinfo{author}{D.~Wang},
\newblock \bibinfo{title}{A survey of transfer learning},
\newblock \bibinfo{journal}{Journal of Big Data} \bibinfo{volume}{3} (\bibinfo{year}{2016}) \bibinfo{pages}{1--40}.
\bibitem[{Chen et~al.(2020)Chen, Hang, Liang, Liu, Li, Wang, Qin, and Choi}]{chen2020novel}
\bibinfo{author}{Y.~Chen}, \bibinfo{author}{W.~Hang}, \bibinfo{author}{S.~Liang}, \bibinfo{author}{X.~Liu}, \bibinfo{author}{G.~Li}, \bibinfo{author}{Q.~Wang}, \bibinfo{author}{J.~Qin}, \bibinfo{author}{K.-S. Choi},
\newblock \bibinfo{title}{A novel transfer support matrix machine for motor imagery-based brain computer interface},
\newblock \bibinfo{journal}{Frontiers in Neuroscience} \bibinfo{volume}{14} (\bibinfo{year}{2020}) \bibinfo{pages}{606949}.
\bibitem[{Pan et~al.(2022)Pan, Sheng, Xu, Tong, Zheng, and Liu}]{pan2022pinball}
\bibinfo{author}{H.~Pan}, \bibinfo{author}{L.~Sheng}, \bibinfo{author}{H.~Xu}, \bibinfo{author}{J.~Tong}, \bibinfo{author}{J.~Zheng}, \bibinfo{author}{Q.~Liu},
\newblock \bibinfo{title}{Pinball transfer support matrix machine for roller bearing fault diagnosis under limited annotation data},
\newblock \bibinfo{journal}{Applied Soft Computing} \bibinfo{volume}{125} (\bibinfo{year}{2022}) \bibinfo{pages}{109209}.
\bibitem[{Mirjalili and Lewis(2016)}]{mirjalili2016whale}
\bibinfo{author}{S.~Mirjalili}, \bibinfo{author}{A.~Lewis},
\newblock \bibinfo{title}{The whale optimization algorithm},
\newblock \bibinfo{journal}{Advances in Engineering Software} \bibinfo{volume}{95} (\bibinfo{year}{2016}) \bibinfo{pages}{51--67}.
\bibitem[{Zheng et~al.(2020)Zheng, Gu, Pan, and Tong}]{zheng2020fault}
\bibinfo{author}{J.~Zheng}, \bibinfo{author}{M.~Gu}, \bibinfo{author}{H.~Pan}, \bibinfo{author}{J.~Tong},
\newblock \bibinfo{title}{A fault classification method for rolling bearing based on multisynchrosqueezing transform and {WOA-SMM}},
\newblock \bibinfo{journal}{IEEE Access} \bibinfo{volume}{8} (\bibinfo{year}{2020}) \bibinfo{pages}{215355--215364}.
\bibitem[{Yu et~al.(2018)Yu, Wang, and Zhao}]{yu2018multisynchrosqueezing}
\bibinfo{author}{G.~Yu}, \bibinfo{author}{Z.~Wang}, \bibinfo{author}{P.~Zhao},
\newblock \bibinfo{title}{Multisynchrosqueezing transform},
\newblock \bibinfo{journal}{IEEE Transactions on Industrial Electronics} \bibinfo{volume}{66} (\bibinfo{year}{2018}) \bibinfo{pages}{5441--5455}.
\bibitem[{Thakur and Wu(2011)}]{thakur2011synchrosqueezing}
\bibinfo{author}{G.~Thakur}, \bibinfo{author}{H.-T. Wu},
\newblock \bibinfo{title}{Synchrosqueezing-based recovery of instantaneous frequency from nonuniform samples},
\newblock \bibinfo{journal}{SIAM Journal on Mathematical Analysis} \bibinfo{volume}{43} (\bibinfo{year}{2011}) \bibinfo{pages}{2078--2095}.
\bibitem[{Ye and Han(2019)}]{ye2019multi}
\bibinfo{author}{Y.~Ye}, \bibinfo{author}{D.~Han},
\newblock \bibinfo{title}{Multi-distance support matrix machines},
\newblock \bibinfo{journal}{Pattern Recognition Letters} \bibinfo{volume}{128} (\bibinfo{year}{2019}) \bibinfo{pages}{237--243}.
\bibitem[{Lyons et~al.(1998)Lyons, Akamatsu, Kamachi, Gyoba, and Budynek}]{lyons1998japanese}
\bibinfo{author}{M.~J. Lyons}, \bibinfo{author}{S.~Akamatsu}, \bibinfo{author}{M.~Kamachi}, \bibinfo{author}{J.~Gyoba}, \bibinfo{author}{J.~Budynek},
\newblock \bibinfo{title}{The japanese female facial expression {(JAFFE)} database},
\newblock in: \bibinfo{booktitle}{Proceedings of Third International Conference on Automatic Face and Gesture Recognition}, \bibinfo{year}{1998}, pp. \bibinfo{pages}{14--16}.
\bibitem[{von~der Malsburg(1996)}]{von1996robust}
\bibinfo{author}{C.~von~der Malsburg},
\newblock \bibinfo{title}{Robust classification of hand postures against complex background},
\newblock in: \bibinfo{booktitle}{Proceedings of the Second International Workshop on Automatic Face and Gesture Recognition, Vermont}, \bibinfo{year}{1996}, pp. \bibinfo{pages}{170--175}.
\bibitem[{Nene et~al.(1996)Nene, Nayar, Murase et~al.}]{nene1996columbia}
\bibinfo{author}{S.~A. Nene}, \bibinfo{author}{S.~K. Nayar}, \bibinfo{author}{H.~Murase}, et~al.,
\newblock \bibinfo{title}{Columbia object image library (coil-20)}  (\bibinfo{year}{1996}).
\bibitem[{LeCun et~al.(1998)LeCun, Bottou, Bengio, and Haffner}]{lecun1998gradient}
\bibinfo{author}{Y.~LeCun}, \bibinfo{author}{L.~Bottou}, \bibinfo{author}{Y.~Bengio}, \bibinfo{author}{P.~Haffner},
\newblock \bibinfo{title}{Gradient-based learning applied to document recognition},
\newblock \bibinfo{journal}{Proceedings of the IEEE} \bibinfo{volume}{86} (\bibinfo{year}{1998}) \bibinfo{pages}{2278--2324}.
\bibitem[{Nazir et~al.(2010)Nazir, Ishtiaq, Batool, Jaffar, and Mirza}]{nazir2010feature}
\bibinfo{author}{M.~Nazir}, \bibinfo{author}{M.~Ishtiaq}, \bibinfo{author}{A.~Batool}, \bibinfo{author}{M.~A. Jaffar}, \bibinfo{author}{A.~M. Mirza},
\newblock \bibinfo{title}{Feature selection for efficient gender classification},
\newblock in: \bibinfo{booktitle}{Proceedings of the 11th WSEAS International Conference}, \bibinfo{year}{2010}, pp. \bibinfo{pages}{70--75}.
\bibitem[{Lyons et~al.(1998)Lyons, Akamatsu, Kamachi, and Gyoba}]{lyons1998coding}
\bibinfo{author}{M.~Lyons}, \bibinfo{author}{S.~Akamatsu}, \bibinfo{author}{M.~Kamachi}, \bibinfo{author}{J.~Gyoba},
\newblock \bibinfo{title}{Coding facial expressions with gabor wavelets},
\newblock in: \bibinfo{booktitle}{Proceedings Third IEEE International Conference on Automatic Face and Gesture Recognition}, \bibinfo{organization}{IEEE}, \bibinfo{year}{1998}, pp. \bibinfo{pages}{200--205}.
\bibitem[{Smola and Sch{\"o}lkopf(2004)}]{smola2004tutorial}
\bibinfo{author}{A.~J. Smola}, \bibinfo{author}{B.~Sch{\"o}lkopf},
\newblock \bibinfo{title}{A tutorial on support vector regression},
\newblock \bibinfo{journal}{Statistics and Computing} \bibinfo{volume}{14} (\bibinfo{year}{2004}) \bibinfo{pages}{199--222}.
\bibitem[{Tang et~al.(2019)Tang, Ma, Hu, and Tang}]{tang2019real}
\bibinfo{author}{X.~Tang}, \bibinfo{author}{Z.~Ma}, \bibinfo{author}{Q.~Hu}, \bibinfo{author}{W.~Tang},
\newblock \bibinfo{title}{A real-time arrhythmia heartbeats classification algorithm using parallel delta modulations and rotated linear-kernel support vector machines},
\newblock \bibinfo{journal}{IEEE Transactions on Biomedical Engineering} \bibinfo{volume}{67} (\bibinfo{year}{2019}) \bibinfo{pages}{978--986}.
\bibitem[{Yuan and Weng(2021)}]{yuan2021support}
\bibinfo{author}{J.~Yuan}, \bibinfo{author}{Y.~Weng},
\newblock \bibinfo{title}{Support matrix regression for learning power flow in distribution grid with unobservability},
\newblock \bibinfo{journal}{IEEE Transactions on Power Systems} \bibinfo{volume}{37} (\bibinfo{year}{2021}) \bibinfo{pages}{1151--1161}.
\bibitem[{Bennett and Demiriz(1998)}]{bennett1998semi}
\bibinfo{author}{K.~Bennett}, \bibinfo{author}{A.~Demiriz},
\newblock \bibinfo{title}{Semi-supervised support vector machines},
\newblock \bibinfo{journal}{Advances in Neural Information Processing Systems} \bibinfo{volume}{11} (\bibinfo{year}{1998}).
\bibitem[{Zhu(2005)}]{zhu2005semi}
\bibinfo{author}{X.~J. Zhu},
\newblock \bibinfo{title}{Semi-supervised learning literature survey}  (\bibinfo{year}{2005}).
\bibitem[{Li et~al.(2023)Li, Li, Yan, Shao, and Lin}]{li2023intelligent}
\bibinfo{author}{X.~Li}, \bibinfo{author}{Y.~Li}, \bibinfo{author}{K.~Yan}, \bibinfo{author}{H.~Shao}, \bibinfo{author}{J.~J. Lin},
\newblock \bibinfo{title}{Intelligent fault diagnosis of bevel gearboxes using semi-supervised probability support matrix machine and infrared imaging},
\newblock \bibinfo{journal}{Reliability Engineering \& System Safety} \bibinfo{volume}{230} (\bibinfo{year}{2023}) \bibinfo{pages}{108921}.
\bibitem[{Liao et~al.(2018)Liao, Weng, Liu, and Rajagopal}]{liao2018urban}
\bibinfo{author}{Y.~Liao}, \bibinfo{author}{Y.~Weng}, \bibinfo{author}{G.~Liu}, \bibinfo{author}{R.~Rajagopal},
\newblock \bibinfo{title}{Urban {MV} and {LV} distribution grid topology estimation via group lasso},
\newblock \bibinfo{journal}{IEEE Transactions on Power Systems} \bibinfo{volume}{34} (\bibinfo{year}{2018}) \bibinfo{pages}{12--27}.
\bibitem[{Narang et~al.(2015)Narang, Ayyanar, Gemin, Baggu, and Srinivasan}]{osti_1171386}
\bibinfo{author}{D.~Narang}, \bibinfo{author}{R.~Ayyanar}, \bibinfo{author}{P.~Gemin}, \bibinfo{author}{M.~Baggu}, \bibinfo{author}{D.~Srinivasan},
\newblock \bibinfo{title}{High penetration of photovoltaic generation study – flagstaff community power (final technical report, results of phases 2-5)}  (\bibinfo{year}{2015}). \URLprefix \url{https://www.osti.gov/biblio/1171386}. \DOIprefix\doi{10.2172/1171386}.
\bibitem[{Srinivasan(2007)}]{srinivasan2007cognitive}
\bibinfo{author}{N.~Srinivasan},
\newblock \bibinfo{title}{Cognitive neuroscience of creativity: {EEG} based approaches},
\newblock \bibinfo{journal}{Methods} \bibinfo{volume}{42} (\bibinfo{year}{2007}) \bibinfo{pages}{109--116}.
\bibitem[{Praline et~al.(2007)Praline, Grujic, Corcia, Lucas, Hommet, Autret, and De~Toffol}]{praline2007emergent}
\bibinfo{author}{J.~Praline}, \bibinfo{author}{J.~Grujic}, \bibinfo{author}{P.~Corcia}, \bibinfo{author}{B.~Lucas}, \bibinfo{author}{C.~Hommet}, \bibinfo{author}{A.~Autret}, \bibinfo{author}{B.~De~Toffol},
\newblock \bibinfo{title}{Emergent {EEG} in clinical practice},
\newblock \bibinfo{journal}{Clinical Neurophysiology} \bibinfo{volume}{118} (\bibinfo{year}{2007}) \bibinfo{pages}{2149--2155}.
\bibitem[{V{\"a}rbu et~al.(2022)V{\"a}rbu, Muhammad, and Muhammad}]{varbu2022past}
\bibinfo{author}{K.~V{\"a}rbu}, \bibinfo{author}{N.~Muhammad}, \bibinfo{author}{Y.~Muhammad},
\newblock \bibinfo{title}{Past, present, and future of {EEG}-based bci applications},
\newblock \bibinfo{journal}{Sensors} \bibinfo{volume}{22} (\bibinfo{year}{2022}) \bibinfo{pages}{3331}.
\bibitem[{Tyagi et~al.(2012)Tyagi, Semwal, and Shah}]{tyagi2012review}
\bibinfo{author}{A.~Tyagi}, \bibinfo{author}{S.~Semwal}, \bibinfo{author}{G.~Shah},
\newblock \bibinfo{title}{A review of {EEG} sensors used for data acquisition},
\newblock \bibinfo{journal}{Journal of Computer Applications (IJCA)}  (\bibinfo{year}{2012}) \bibinfo{pages}{13--17}.
\bibitem[{Mumtaz et~al.(2021)Mumtaz, Rasheed, and Irfan}]{mumtaz2021review}
\bibinfo{author}{W.~Mumtaz}, \bibinfo{author}{S.~Rasheed}, \bibinfo{author}{A.~Irfan},
\newblock \bibinfo{title}{Review of challenges associated with the {EEG} artifact removal methods},
\newblock \bibinfo{journal}{Biomedical Signal Processing and Control} \bibinfo{volume}{68} (\bibinfo{year}{2021}) \bibinfo{pages}{102741}.
\bibitem[{Razzak et~al.(2019)Razzak, Hameed, and Xu}]{razzak2019robust}
\bibinfo{author}{I.~Razzak}, \bibinfo{author}{I.~A. Hameed}, \bibinfo{author}{G.~Xu},
\newblock \bibinfo{title}{Robust sparse representation and multiclass support matrix machines for the classification of motor imagery {EEG} signals},
\newblock \bibinfo{journal}{IEEE Journal of Translational Engineering in Health and Medicine} \bibinfo{volume}{7} (\bibinfo{year}{2019}) \bibinfo{pages}{1--8}.
\bibitem[{Lotte et~al.(2007)Lotte, Congedo, L{\'e}cuyer, Lamarche, and Arnaldi}]{lotte2007review}
\bibinfo{author}{F.~Lotte}, \bibinfo{author}{M.~Congedo}, \bibinfo{author}{A.~L{\'e}cuyer}, \bibinfo{author}{F.~Lamarche}, \bibinfo{author}{B.~Arnaldi},
\newblock \bibinfo{title}{A review of classification algorithms for {EEG}-based brain--computer interfaces},
\newblock \bibinfo{journal}{Journal of Neural Engineering} \bibinfo{volume}{4} (\bibinfo{year}{2007}) \bibinfo{pages}{R1}.
\bibitem[{Devlaminck et~al.(2011)Devlaminck, Wyns, Grosse-Wentrup, Otte, and Santens}]{devlaminck2011multisubject}
\bibinfo{author}{D.~Devlaminck}, \bibinfo{author}{B.~Wyns}, \bibinfo{author}{M.~Grosse-Wentrup}, \bibinfo{author}{G.~Otte}, \bibinfo{author}{P.~Santens},
\newblock \bibinfo{title}{Multisubject learning for common spatial patterns in motor-imagery {BCI}},
\newblock \bibinfo{journal}{Computational Intelligence and Neuroscience} \bibinfo{volume}{2011} (\bibinfo{year}{2011}) \bibinfo{pages}{8--8}.
\bibitem[{Bischof and Bunch(2021)}]{bischof2021geometric}
\bibinfo{author}{B.~Bischof}, \bibinfo{author}{E.~Bunch},
\newblock \bibinfo{title}{Geometric feature performance under downsampling for {EEG} classification tasks},
\newblock \bibinfo{journal}{arXiv preprint arXiv:2102.07669}  (\bibinfo{year}{2021}).
\bibitem[{Vidaurre et~al.(2009)Vidaurre, Kr{\"a}mer, Blankertz, and Schl{\"o}gl}]{vidaurre2009time}
\bibinfo{author}{C.~Vidaurre}, \bibinfo{author}{N.~Kr{\"a}mer}, \bibinfo{author}{B.~Blankertz}, \bibinfo{author}{A.~Schl{\"o}gl},
\newblock \bibinfo{title}{Time domain parameters as a feature for {EEG}-based brain--computer interfaces},
\newblock \bibinfo{journal}{Neural Networks} \bibinfo{volume}{22} (\bibinfo{year}{2009}) \bibinfo{pages}{1313--1319}.
\bibitem[{Shakshi and Jaswal(2016)}]{shakshi2016brain}
\bibinfo{author}{R.~J. Shakshi}, \bibinfo{author}{R.~Jaswal},
\newblock \bibinfo{title}{Brain wave classification and feature extraction of {EEG} signal by using {FFT} on lab view},
\newblock \bibinfo{journal}{Int. Res. J. Eng. Technol} \bibinfo{volume}{3} (\bibinfo{year}{2016}) \bibinfo{pages}{1208--1212}.
\bibitem[{Kuncheva and Faithfull(2013)}]{kuncheva2013pca}
\bibinfo{author}{L.~I. Kuncheva}, \bibinfo{author}{W.~J. Faithfull},
\newblock \bibinfo{title}{{PCA} feature extraction for change detection in multidimensional unlabeled data},
\newblock \bibinfo{journal}{IEEE Transactions on Neural Networks and Learning Systems} \bibinfo{volume}{25} (\bibinfo{year}{2013}) \bibinfo{pages}{69--80}.
\bibitem[{Hu and Zhang(2019)}]{hu2019eeg}
\bibinfo{author}{L.~Hu}, \bibinfo{author}{Z.~Zhang}, \bibinfo{title}{{EEG} signal processing and feature extraction}, \bibinfo{publisher}{Springer}, \bibinfo{year}{2019}.
\bibitem[{Sen et~al.(2023)Sen, Mishra, and Pattnaik}]{sen2023review}
\bibinfo{author}{D.~Sen}, \bibinfo{author}{B.~B. Mishra}, \bibinfo{author}{P.~K. Pattnaik},
\newblock \bibinfo{title}{A review of the filtering techniques used in {EEG} signal processing},
\newblock in: \bibinfo{booktitle}{2023 7th International Conference on Trends in Electronics and Informatics (ICOEI)}, \bibinfo{organization}{IEEE}, \bibinfo{year}{2023}, pp. \bibinfo{pages}{270--277}.
\bibitem[{Nicolas-Alonso and Gomez-Gil(2012)}]{nicolas2012brain}
\bibinfo{author}{L.~F. Nicolas-Alonso}, \bibinfo{author}{J.~Gomez-Gil},
\newblock \bibinfo{title}{Brain computer interfaces, a review},
\newblock \bibinfo{journal}{Sensors} \bibinfo{volume}{12} (\bibinfo{year}{2012}) \bibinfo{pages}{1211--1279}.
\bibitem[{Pise and Rege(2021)}]{pise2021comparative}
\bibinfo{author}{A.~W. Pise}, \bibinfo{author}{P.~P. Rege},
\newblock \bibinfo{title}{Comparative analysis of various filtering techniques for denoising {EEG} signals},
\newblock in: \bibinfo{booktitle}{2021 6th International Conference for Convergence in Technology (I2CT)}, \bibinfo{organization}{IEEE}, \bibinfo{year}{2021}, pp. \bibinfo{pages}{1--4}.
\bibitem[{Vaid et~al.(2015)Vaid, Singh, and Kaur}]{vaid2015eeg}
\bibinfo{author}{S.~Vaid}, \bibinfo{author}{P.~Singh}, \bibinfo{author}{C.~Kaur},
\newblock \bibinfo{title}{{EEG} signal analysis for bci interface: A review},
\newblock in: \bibinfo{booktitle}{2015 Fifth International Conference on Advanced Computing \& Communication Technologies}, \bibinfo{organization}{IEEE}, \bibinfo{year}{2015}, pp. \bibinfo{pages}{143--147}.
\bibitem[{Yan et~al.(2023)Yan, Wang, Lu, Zhou, and Peng}]{yan2023review}
\bibinfo{author}{W.~Yan}, \bibinfo{author}{J.~Wang}, \bibinfo{author}{S.~Lu}, \bibinfo{author}{M.~Zhou}, \bibinfo{author}{X.~Peng},
\newblock \bibinfo{title}{A review of real-time fault diagnosis methods for industrial smart manufacturing},
\newblock \bibinfo{journal}{Processes} \bibinfo{volume}{11} (\bibinfo{year}{2023}) \bibinfo{pages}{369}.
\bibitem[{Pernest{\aa}l(2009)}]{pernestaal2009probabilistic}
\bibinfo{author}{A.~Pernest{\aa}l}, \bibinfo{title}{Probabilistic fault diagnosis with automotive applications}, Ph.D. thesis, Link{\"o}ping University Electronic Press, \bibinfo{year}{2009}.
\bibitem[{Patton(1990)}]{patton1990fault}
\bibinfo{author}{R.~J. Patton},
\newblock \bibinfo{title}{Fault detection and diagnosis in aerospace systems using analytical redundancy},
\newblock in: \bibinfo{booktitle}{IEE Colloquium on Condition Monitoring and Fault Tolerance}, \bibinfo{organization}{IET}, \bibinfo{year}{1990}, pp. \bibinfo{pages}{1--1}.
\bibitem[{Sekine et~al.(1992)Sekine, Akimoto, Kunugi, Fukui, and Fukui}]{sekine1992fault}
\bibinfo{author}{Y.~Sekine}, \bibinfo{author}{Y.~Akimoto}, \bibinfo{author}{M.~Kunugi}, \bibinfo{author}{C.~Fukui}, \bibinfo{author}{S.~Fukui},
\newblock \bibinfo{title}{Fault diagnosis of power systems},
\newblock \bibinfo{journal}{Proceedings of the IEEE} \bibinfo{volume}{80} (\bibinfo{year}{1992}) \bibinfo{pages}{673--683}.
\bibitem[{Pan et~al.(2022)Pan, Xu, Zheng, Liu, and Tong}]{pan2022intelligent}
\bibinfo{author}{H.~Pan}, \bibinfo{author}{H.~Xu}, \bibinfo{author}{J.~Zheng}, \bibinfo{author}{Q.~Liu}, \bibinfo{author}{J.~Tong},
\newblock \bibinfo{title}{An intelligent fault diagnosis method for roller bearings using an adaptive interactive deviation matrix machine},
\newblock \bibinfo{journal}{Measurement Science and Technology} \bibinfo{volume}{33} (\bibinfo{year}{2022}) \bibinfo{pages}{075103}.
\bibitem[{Pan et~al.(2019)Pan, Yang, Zheng, Li, and Cheng}]{pan2019fault}
\bibinfo{author}{H.~Pan}, \bibinfo{author}{Y.~Yang}, \bibinfo{author}{J.~Zheng}, \bibinfo{author}{X.~Li}, \bibinfo{author}{J.~Cheng},
\newblock \bibinfo{title}{A fault diagnosis approach for roller bearing based on symplectic geometry matrix machine},
\newblock \bibinfo{journal}{Mechanism and Machine Theory} \bibinfo{volume}{140} (\bibinfo{year}{2019}) \bibinfo{pages}{31--43}.
\bibitem[{Pan and Zheng(2021)}]{pan2021intelligent}
\bibinfo{author}{H.~Pan}, \bibinfo{author}{J.~Zheng},
\newblock \bibinfo{title}{An intelligent fault diagnosis method for roller bearing using symplectic hyperdisk matrix machine},
\newblock \bibinfo{journal}{Applied Soft Computing} \bibinfo{volume}{105} (\bibinfo{year}{2021}) \bibinfo{pages}{107284}.
\bibitem[{Pan et~al.(2022)Pan, Xu, and Zheng}]{pan2022novel}
\bibinfo{author}{H.~Pan}, \bibinfo{author}{H.~Xu}, \bibinfo{author}{J.~Zheng},
\newblock \bibinfo{title}{A novel symplectic relevance matrix machine method for intelligent fault diagnosis of roller bearing},
\newblock \bibinfo{journal}{Expert Systems with Applications} \bibinfo{volume}{192} (\bibinfo{year}{2022}) \bibinfo{pages}{116400}.
\bibitem[{Liu et~al.(2019)Liu, Jiao, Zhang, and Liu}]{liu2019polsar}
\bibinfo{author}{X.~Liu}, \bibinfo{author}{L.~Jiao}, \bibinfo{author}{D.~Zhang}, \bibinfo{author}{F.~Liu},
\newblock \bibinfo{title}{Polsar image classification based on polarimetric scattering coding and sparse support matrix machine},
\newblock in: \bibinfo{booktitle}{IGARSS 2019-2019 IEEE International Geoscience and Remote Sensing Symposium}, \bibinfo{organization}{IEEE}, \bibinfo{year}{2019}, pp. \bibinfo{pages}{3181--3184}.
\bibitem[{Tanveer et~al.(2021)Tanveer, Tiwari, Choudhary, and Ganaie}]{tanveer2021large}
\bibinfo{author}{M.~Tanveer}, \bibinfo{author}{A.~Tiwari}, \bibinfo{author}{R.~Choudhary}, \bibinfo{author}{M.~Ganaie},
\newblock \bibinfo{title}{Large-scale pinball twin support vector machines},
\newblock \bibinfo{journal}{Machine Learning}  (\bibinfo{year}{2021}) \bibinfo{pages}{1--24}.
\bibitem[{Zhang et~al.(2005)Zhang, Li, and Yang}]{zhang2005parallel}
\bibinfo{author}{J.-P. Zhang}, \bibinfo{author}{Z.-W. Li}, \bibinfo{author}{J.~Yang},
\newblock \bibinfo{title}{A parallel {SVM} training algorithm on large-scale classification problems},
\newblock in: \bibinfo{booktitle}{2005 International Conference on Machine Learning and Cybernetics}, volume~\bibinfo{volume}{3}, \bibinfo{organization}{IEEE}, \bibinfo{year}{2005}, pp. \bibinfo{pages}{1637--1641}.
\bibitem[{Vapnik and Vashist(2009)}]{vapnik2009new}
\bibinfo{author}{V.~Vapnik}, \bibinfo{author}{A.~Vashist},
\newblock \bibinfo{title}{A new learning paradigm: Learning using privileged information},
\newblock \bibinfo{journal}{Neural Networks} \bibinfo{volume}{22} (\bibinfo{year}{2009}) \bibinfo{pages}{544--557}.
\bibitem[{Houthuys et~al.(2018)Houthuys, Langone, and Suykens}]{houthuys2018multi}
\bibinfo{author}{L.~Houthuys}, \bibinfo{author}{R.~Langone}, \bibinfo{author}{J.~A. Suykens},
\newblock \bibinfo{title}{Multi-view least squares support vector machines classification},
\newblock \bibinfo{journal}{Neurocomputing} \bibinfo{volume}{282} (\bibinfo{year}{2018}) \bibinfo{pages}{78--88}.
\bibitem[{Huang et~al.(2022)Huang, Wei, and Zhou}]{huang2022overview}
\bibinfo{author}{H.~Huang}, \bibinfo{author}{X.~Wei}, \bibinfo{author}{Y.~Zhou},
\newblock \bibinfo{title}{An overview on twin support vector regression},
\newblock \bibinfo{journal}{Neurocomputing} \bibinfo{volume}{490} (\bibinfo{year}{2022}) \bibinfo{pages}{80--92}.
\bibitem[{Ganaie et~al.(2022)Ganaie, Tanveer, and Beheshti}]{ganaie2022brain}
\bibinfo{author}{M.~Ganaie}, \bibinfo{author}{M.~Tanveer}, \bibinfo{author}{I.~Beheshti},
\newblock \bibinfo{title}{Brain age prediction with improved least squares twin svr},
\newblock \bibinfo{journal}{IEEE Journal of Biomedical and Health Informatics} \bibinfo{volume}{27} (\bibinfo{year}{2022}) \bibinfo{pages}{1661--1669}.
\bibitem[{Iglovikov et~al.(2018)Iglovikov, Rakhlin, Kalinin, and Shvets}]{iglovikov2018paediatric}
\bibinfo{author}{V.~I. Iglovikov}, \bibinfo{author}{A.~Rakhlin}, \bibinfo{author}{A.~A. Kalinin}, \bibinfo{author}{A.~A. Shvets},
\newblock \bibinfo{title}{Paediatric bone age assessment using deep convolutional neural networks},
\newblock in: \bibinfo{booktitle}{Deep Learning in Medical Image Analysis and Multimodal Learning for Clinical Decision Support: 4th International Workshop, DLMIA 2018, and 8th International Workshop, ML-CDS 2018, Held in Conjunction with MICCAI 2018, Granada, Spain, September 20, 2018, Proceedings 4}, \bibinfo{organization}{Springer}, \bibinfo{year}{2018}, pp. \bibinfo{pages}{300--308}.
\bibitem[{Naeini et~al.(2010)Naeini, Taremian, and Hashemi}]{naeini2010stock}
\bibinfo{author}{M.~P. Naeini}, \bibinfo{author}{H.~Taremian}, \bibinfo{author}{H.~B. Hashemi},
\newblock \bibinfo{title}{Stock market value prediction using neural networks},
\newblock in: \bibinfo{booktitle}{2010 International Conference on Computer Information Systems and Industrial Management Applications (CISIM)}, \bibinfo{organization}{IEEE}, \bibinfo{year}{2010}, pp. \bibinfo{pages}{132--136}.
\bibitem[{Winters-Hilt and Merat(2007)}]{winters2007svm}
\bibinfo{author}{S.~Winters-Hilt}, \bibinfo{author}{S.~Merat},
\newblock \bibinfo{title}{{SVM} clustering},
\newblock in: \bibinfo{booktitle}{BMC Bioinformatics}, volume~\bibinfo{volume}{8}, \bibinfo{organization}{BioMed Central}, \bibinfo{year}{2007}, pp. \bibinfo{pages}{1--12}.
\bibitem[{Mountrakis et~al.(2011)Mountrakis, Im, and Ogole}]{mountrakis2011support}
\bibinfo{author}{G.~Mountrakis}, \bibinfo{author}{J.~Im}, \bibinfo{author}{C.~Ogole},
\newblock \bibinfo{title}{Support vector machines in remote sensing: A review},
\newblock \bibinfo{journal}{ISPRS Journal of Photogrammetry and Remote Sensing} \bibinfo{volume}{66} (\bibinfo{year}{2011}) \bibinfo{pages}{247--259}.
\bibitem[{Rezvani and Wang(2023)}]{rezvani2023broad}
\bibinfo{author}{S.~Rezvani}, \bibinfo{author}{X.~Wang},
\newblock \bibinfo{title}{A broad review on class imbalance learning techniques},
\newblock \bibinfo{journal}{Applied Soft Computing}  (\bibinfo{year}{2023}) \bibinfo{pages}{110415}.
\bibitem[{Laurikkala(2001)}]{laurikkala2001improving}
\bibinfo{author}{J.~Laurikkala},
\newblock \bibinfo{title}{Improving identification of difficult small classes by balancing class distribution},
\newblock in: \bibinfo{booktitle}{Artificial Intelligence in Medicine: 8th Conference on Artificial Intelligence in Medicine in Europe, AIME, Proceedings 8}, \bibinfo{organization}{Springer}, \bibinfo{year}{2001}, pp. \bibinfo{pages}{63--66}.
\bibitem[{Hart(1968)}]{hart1968condensed}
\bibinfo{author}{P.~Hart},
\newblock \bibinfo{title}{The condensed nearest neighbor rule (corresp.)},
\newblock \bibinfo{journal}{IEEE Transactions on Information Theory} \bibinfo{volume}{14} (\bibinfo{year}{1968}) \bibinfo{pages}{515--516}.
\bibitem[{Chawla et~al.(2002)Chawla, Bowyer, Hall, and Kegelmeyer}]{chawla2002smote}
\bibinfo{author}{N.~V. Chawla}, \bibinfo{author}{K.~W. Bowyer}, \bibinfo{author}{L.~O. Hall}, \bibinfo{author}{W.~P. Kegelmeyer},
\newblock \bibinfo{title}{{SMOTE}: synthetic minority over-sampling technique},
\newblock \bibinfo{journal}{Journal of Artificial Intelligence Research} \bibinfo{volume}{16} (\bibinfo{year}{2002}) \bibinfo{pages}{321--357}.
\bibitem[{Rezvani et~al.(2019)Rezvani, Wang, and Pourpanah}]{rezvani2019intuitionistic}
\bibinfo{author}{S.~Rezvani}, \bibinfo{author}{X.~Wang}, \bibinfo{author}{F.~Pourpanah},
\newblock \bibinfo{title}{Intuitionistic fuzzy twin support vector machines},
\newblock \bibinfo{journal}{IEEE Transactions on Fuzzy Systems} \bibinfo{volume}{27} (\bibinfo{year}{2019}) \bibinfo{pages}{2140--2151}.
\bibitem[{Lee(2004)}]{lee2004first}
\bibinfo{author}{K.~H. Lee}, \bibinfo{title}{First course on fuzzy theory and applications}, volume~\bibinfo{volume}{27}, \bibinfo{publisher}{Springer Science \& Business Media}, \bibinfo{year}{2004}.
\bibitem[{Weston et~al.(2006)Weston, Collobert, Sinz, Bottou, and Vapnik}]{weston2006inference}
\bibinfo{author}{J.~Weston}, \bibinfo{author}{R.~Collobert}, \bibinfo{author}{F.~Sinz}, \bibinfo{author}{L.~Bottou}, \bibinfo{author}{V.~Vapnik},
\newblock \bibinfo{title}{Inference with the universum},
\newblock in: \bibinfo{booktitle}{Proceedings of the 23rd International Conference on Machine Learning}, \bibinfo{year}{2006}, pp. \bibinfo{pages}{1009--1016}.
\bibitem[{Akhtar et~al.(2023)Akhtar, Tanveer, and Arshad}]{akhtar2023roboss}
\bibinfo{author}{M.~Akhtar}, \bibinfo{author}{M.~Tanveer}, \bibinfo{author}{M.~Arshad},
\newblock \bibinfo{title}{Roboss: A robust, bounded, sparse, and smooth loss function for supervised learning},
\newblock \bibinfo{journal}{arXiv preprint arXiv:2309.02250}  (\bibinfo{year}{2023}).
\bibitem[{Tian et~al.(2023)Tian, Zhao, and Fu}]{tian2023kernel}
\bibinfo{author}{Y.~Tian}, \bibinfo{author}{X.~Zhao}, \bibinfo{author}{S.~Fu},
\newblock \bibinfo{title}{Kernel methods with asymmetric and robust loss function},
\newblock \bibinfo{journal}{Expert Systems with Applications} \bibinfo{volume}{213} (\bibinfo{year}{2023}) \bibinfo{pages}{119236}.
\bibitem[{Feng et~al.(2016)Feng, Yang, Huang, Mehrkanoon, and Suykens}]{feng2016robust}
\bibinfo{author}{Y.~Feng}, \bibinfo{author}{Y.~Yang}, \bibinfo{author}{X.~Huang}, \bibinfo{author}{S.~Mehrkanoon}, \bibinfo{author}{J.~A. Suykens},
\newblock \bibinfo{title}{Robust support vector machines for classification with nonconvex and smooth losses},
\newblock \bibinfo{journal}{Neural Computation} \bibinfo{volume}{28} (\bibinfo{year}{2016}) \bibinfo{pages}{1217--1247}.
\bibitem[{Das and Suganthan(2010)}]{das2010differential}
\bibinfo{author}{S.~Das}, \bibinfo{author}{P.~N. Suganthan},
\newblock \bibinfo{title}{Differential evolution: A survey of the state-of-the-art},
\newblock \bibinfo{journal}{IEEE Transactions on Evolutionary Computation} \bibinfo{volume}{15} (\bibinfo{year}{2010}) \bibinfo{pages}{4--31}.

\end{thebibliography}
\end{document}